\documentclass[10pt,twocolumn,letterpaper]{article}

\usepackage{iccv}
\usepackage{times}
\usepackage{epsfig}
\usepackage{graphicx}
\usepackage{amsmath}
\usepackage{amssymb}
\usepackage{mathrsfs}
\usepackage{epstopdf}
\usepackage{multirow}
\usepackage{subfigure}
\usepackage{enumitem}
\usepackage[accsupp]{axessibility}

\usepackage[breaklinks=true,bookmarks=false]{hyperref}

\iccvfinalcopy 

\def\iccvPaperID{2701} 

\ificcvfinal\fi

\begin{document}

\title{Complementary Patch for Weakly Supervised Semantic Segmentation}

\author{Fei Zhang, Chaochen Gu\footnotemark[1], Chenyue Zhang\\
Shanghai Jiao Tong University, China\\
{\tt\small \{ferenas, jacygu, lucklypeach\}@sjtu.edu.cn}
\and
Yuchao Dai\\
Northwestern Polytechnical University, China\\
{\tt\small daiyuchao@nwpu.edu.cn}
}

\maketitle

\renewcommand{\thefootnote}{\fnsymbol{footnote}} 
\footnotetext[1]{Corresponding author} 

\ificcvfinal\thispagestyle{empty}\fi

\begin{abstract}
Weakly Supervised Semantic Segmentation (WSSS) based on image-level labels has been greatly advanced by exploiting the outputs of Class Activation Map (CAM) to generate the pseudo labels for semantic segmentation. However, CAM merely discovers seeds from a small number of regions, which may be insufficient to serve as pseudo masks for semantic segmentation.
In this paper, we formulate the expansion of object regions in CAM as an increase in information. From the perspective of information theory, we propose a novel Complementary Patch (CP) Representation and prove that the information of the sum of the CAMs by a pair of input images with complementary hidden (patched) parts, namely CP Pair, is greater than or equal to the information of the baseline CAM.
Therefore, a CAM with more information related to object seeds can be obtained by narrowing down the gap between the sum of CAMs generated by the CP Pair and the original CAM.
We propose a CP Network (CPN) implemented by a triplet network and three regularization functions.
To further improve the quality of the CAMs, we propose a Pixel-Region Correlation Module (PRCM) to augment the contextual information by using object-region relations between the feature maps and the CAMs.
Experimental results on the PASCAL VOC 2012 datasets show that our proposed method achieves a new state-of-the-art in WSSS, validating the effectiveness of our CP Representation and CPN.
\end{abstract}
\section{Introduction}

Thanks to the booming of deep learning methods, recent years have witnessed extraordinary progress in semantic segmentation~\cite{fcn,psp,v2,v3plus}. However, the prerequisite of a successful neural network for semantic segmentation is pixel-level segmentation ground-truth, which requires massive investments in manual annotation. Numerous efforts have been devoted to developing Weakly Supervised Semantic Segmentation (WSSS) to ease the pressure, which aims to train a semantic segmentation network by using weaker supervision, such as image-level classification labels~\cite{sec,subE,affinity,1stage}, bounding boxes~\cite{bbox1,bbox2}, scribbles~\cite{scribble} and points~\cite{point}. Image-level labels, as the most conveniently-acquired annotation format, have been extensively studied in WSSS. In this work, we particularly focus on WSSS using image-level labels.
\begin{figure}
\begin{center}
\includegraphics[height = 1.35in,width=2.8in]{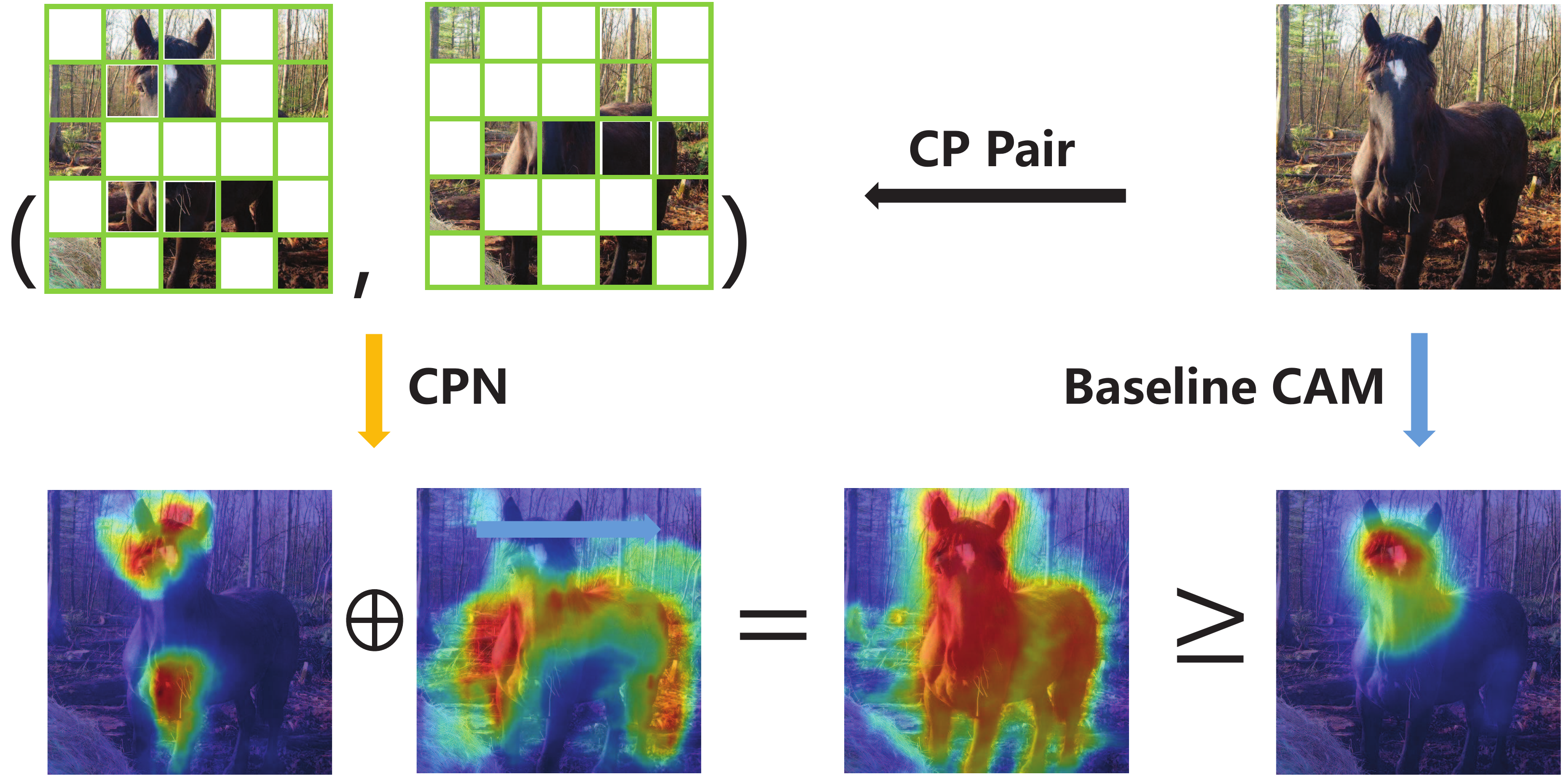} 
\end{center}
\vspace{-3mm}
   \caption{Illustration of our proposed method. The original CAM simply finds object seeds in most discriminative regions. To enlarge the seed areas, our Complementary Patch Network (CPN) uses a pair of images with CP regions (CP Pair) to generate two CAMs, the sum of which are supposed to incorporate more information of the foreground than the original CAM.}

\label{fig:first}
\end{figure}

Most WSSS approaches generating initial seeds through image-level labels heavily rely on an efficient method---Class Activation Map (CAM)~\cite{cam}.
Nevertheless, this architecture appears to be barely sensitive to the most discriminative regions, resulting in many incomplete foreground areas. To address the issue, a promising way is to erase or ignore some high response regions to help CAM 'see' more seeds in an image~\cite{erase1,fickle,has,adl}, \ie, region erasing or mining methods. However, these methods are more or less losing part of the regions of an image in each training epoch due to the randomness of the hiding process. It seems to be effective to intentionally cover the high response areas identified from the CAM in each training epoch, while such iterative operation introduces much computational complexity, and it is difficult to properly determine the number of iterations for each image as well.

In this paper, we show that CAM could explore more high response areas by taking full advantage of the information of an image, specifically including both uncovered and hidden parts. Based on the motivation, we treat the task of expanding object seeds in CAM as an increase in information, and develop a simple yet efficient concept---Complementary Patch (CP) Representation: the self-information of the CAM of an image is less than or equal to the sum of the self-information of the CAMs, which are obtained by CP Pair, namely two images with CP regions. Therefore, an improved CAM could be obtained by adding up the CAMs generated by the CP Pair (shown in Fig.~\ref{fig:first}). In addition, we show that the equality holds under two extreme cases. One is that if the patch size is too large, one of the CP Pair equals the original image, and the other is that two images in the CP Pair are almost the same for the network if the patch size is too small. Under these extreme conditions, the CP Pair is unable to seek out new seed areas compared to the original image. Thus the degree of the increase in information (object seeds) is subject to the patch size for the CP Pair.

Building upon the CP Representation, we propose a CP Network (CPN) to narrow down the gap between the improved CAM mentioned above and the one by the original image. CPN is formed by a triplet network with Triplet CP (TCP) loss and CP Cross Regularization (CPCR) loss, serving as minimizing the above discrepancy. For the generation of the CP Pair, we propose to use grid (Grid Patch) or super-pixel (Super-pixel Patch) as the patch template. Furthermore, CPN introduces a Pixel-Region Correlation Module (PRCM), which aims to capture the relationship between the pixels and regions, and incorporates it with Pixel Correlation Module (PCM)~\cite{seam} to further improve the consistency of the predicted CAM.

Extensive experiments conducted on PASCAL VOC 2012~\cite{voc12} demonstrate the effectiveness of our CPN. As a result, our model yields a new state-of-the-art performance by 67.8\% and 68.5\% on the \textit{val} set and \textit{test} set.
Furthermore, we notice that the performance of our CPN is influenced by the patch size, which is in accordance with our analysis of the CP Representation in extreme cases.

Our main contributions are summarized as three-fold:
\begin{itemize}
\item[$\bullet$] We propose a simple yet effective Complementary Patch (CP) representation to enlarge the seed regions in CAM, which narrows down the gap between the original CAM and the CAMs by summing up the CAMs of the CP pair.
\item[$\bullet$] Building upon the CP representation, we present a triplet network (CPN) with Triplet CP (TCP) loss and CP Cross Regularization (CPCR). Moreover, a Pixel-Region Correlation Module (PRCM) is proposed to further refine the CAM.
\item[$\bullet$] Experimental results on the PASCAL VOC 2012 show that our proposed framework achieves state-of-the-art performance in WSSS.
\end{itemize}
\section{Related Work}
\noindent\textbf{Weakly supervised semantic segmentation}\quad 
As the most economic form in WSSS, image-level supervision has gained increasing attention from academia and industry. Recent advanced methods focus on modifying the seed areas produced by Class Activation Map (CAM)~\cite{cam}. The first category~\cite{wildcat,sec,spx} is dedicated to pooling-based methods to overcome the drawbacks led by Global Max-Pooling (GMP) and Global Average-Pooling (GAP), which are used to aggregate score maps into a classification score. SPN~\cite{spx} proposes to regard super-pixel segmentation of the input image as the pooling module. The second category~\cite{dsrg,affinity,seam,1stage,coatten,irnet,erase2} investigates the inter-pixel or semantic relationship to expand the seed areas or remove the wrong seeds~\cite{conta}. AffinityNet~\cite{affinity} proposes to learn the similarity between pixels and applies Random Walk (RW) to further refine the seed areas. 
The third category concentrates on making efficient use of extra easily-obtained resources, including web images~\cite{video1},videos~\cite{video1,video2} and saliency maps~\cite{stc}. The fourth category turns to region erasing or mining methods, aiming to mark out more object regions in CAM by erasing or mining some high response regions. Adversarial Erasing~\cite{erase1} aims to explore more object seeds by iteratively erasing the discriminative regions detected by the original CAM from the image. However, it is hard to decide the exact number of iterations for each image. Attention-based Dropout Layer~\cite{adl} is a tool that highlights the potential points by thresholding the attention maps obtained from the feature maps. To expand the seed areas, FickleNet~\cite{fickle} calculates the final score maps by randomly selecting the hidden units in the feature maps. As a data augmentation, Hide-and-Seek (HAS)~\cite{has} enlarges the seed areas by randomly hiding grid patches in each image. Nevertheless, these hiding methods above are incapable of using the entire information in an image during each training epoch. To excavate full information in an image as much as possible, we propose Complementary Patch (CP) Representation and design the CPN to support CAM mine out more foreground seeds.
\\
\\
\noindent\textbf{Self-attention model}\quad For the sake of improving the quality of segmentation masks, models based on self-attention~\cite{attention}, refining the feature maps by use of context feature, is widely employed in various segmentation networks. Wang \etal~\cite{nonlocal} proposes non-local block to produce an attention map by taking account of the correlation between each spatial point in the feature maps. To further enrich the contextual information, DANet~\cite{DANet} combines two self-attention modules, namely channel attention and spatial attention. Yuan \etal~\cite{OCR} proposes the object-contextual representations to identify a pixel by using its corresponding object class, reinforcing the object contextual information.
\begin{figure*}
\begin{center}
\includegraphics[width=1.0\linewidth]{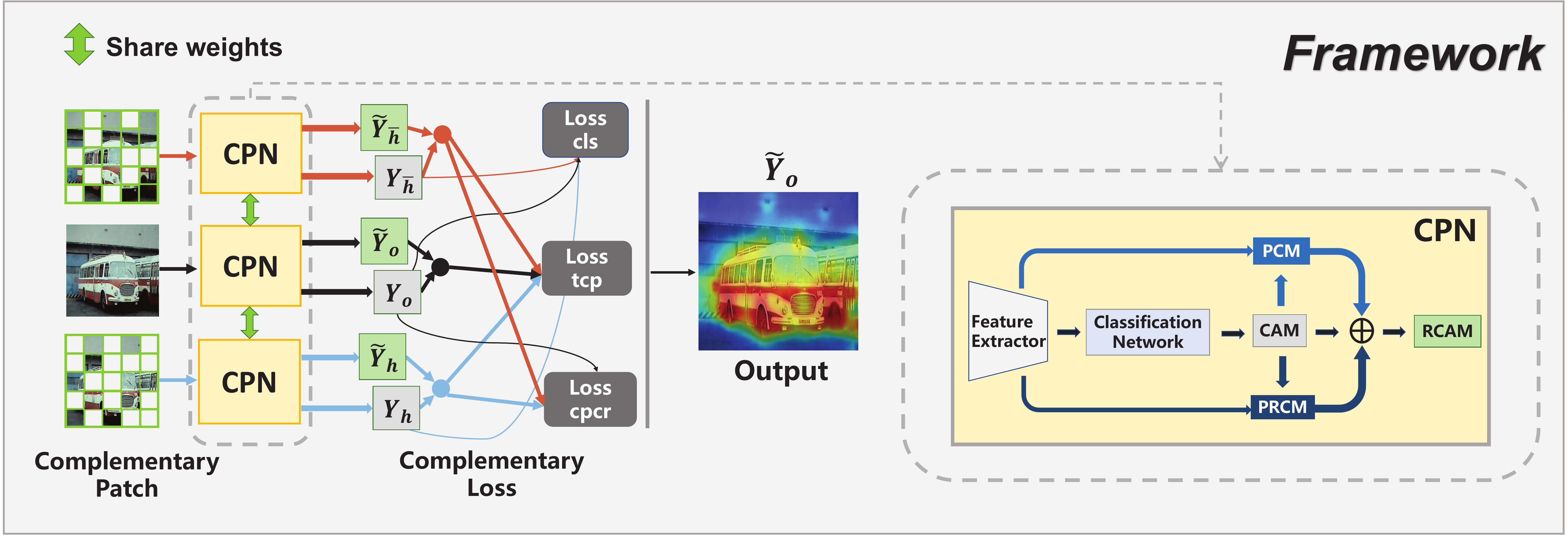}
\end{center}
\vspace{-4mm}
   \caption{The overall framework of our method. The whole structure of CPN is a triplet network with three branches, jointly feeding the original image (the black flow) and the CP Pair (the red and blue flows). PCM and the proposed PRCM collectively improve the quality of the original CAM to the refined CAM (RCAM). Finally, all outputs are constrained with three losses, which are $\mathcal {L}_{cls}, \mathcal {L}_{tcp}$ and $\mathcal {L}_{cpcr}$. The dot (the red, blue, or black one) means that both outputs connected to it are leveraged in the following loss. During inference, the RCAM from the original image ($\widetilde {\boldsymbol{Y}}_{o}$) is used to predict the mask for segmentation. }
\label{fig:datu}
\end{figure*}
\section{Proposed Methods}
%
\subsection{Complementary Patch Representation}\label{CPR}
Let us denote the CAM of an image $\boldsymbol{I}$ of size $3 \times H \times W$ as $\boldsymbol{Y} \in \mathbb{R}^{C \times H \times W}$, where $C$ refers to the number of objects (including background). The generation of $\boldsymbol{Y}$ typically begins with training a multi-label classification network, comprising a feature extractor layer, a Global Average-Pooling (GAP) layer, and a classification layer. Thus $\boldsymbol{Y}$ related to the $c$-th object, denoted as $\boldsymbol{Y}^{c}$, can be gained by:
\begin{equation}
\boldsymbol{Y}^{c}=({\boldsymbol{\theta}^{c}})^\mathsf {T}{\boldsymbol{f}},\label{eq1}
\end{equation}
where $\boldsymbol{f} \in \mathbb{R}^{C_f \times H \times W}$ with $C_f$ channels is the feature maps from the final layer, and $\boldsymbol{\theta}^{c} \in \mathbb{R}^{C_f \times 1}$ is the corresponding classifier weight of $c$-th class in the classification layer. Denote $\boldsymbol{I}$ with some hidden units as $\boldsymbol{I}_h \in \mathbb{R}^{3 \times H \times W}$ and its counterpart with complementary hidden regions as $\boldsymbol{I}_{\overline h} \in \mathbb{R}^{3 \times H \times W}$. Intuitively, we have $\boldsymbol{I_h} + \boldsymbol{I_{\overline h}}  = \boldsymbol{I}$. For convenience, we name $(\boldsymbol{I}_h, \boldsymbol{I}_{\overline h})$ as the Complementary Patch Pair (CP Pair) of $\boldsymbol{I}$.

\begin{figure}
\begin{center}
\includegraphics[height = 1.2in,width=3.3in]{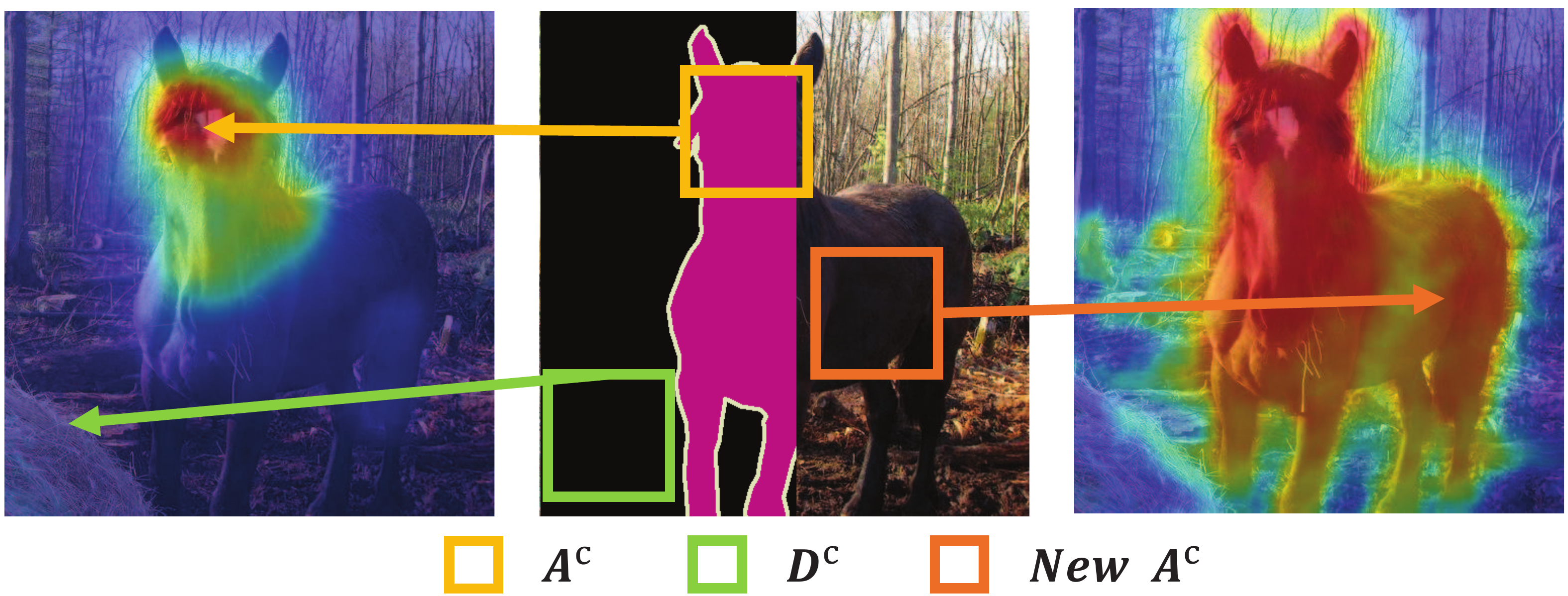}
\end{center}
\vspace{-4mm}
   \caption{Two kinds of patch regions in an image. $\mathcal{A}^c$ refers to the region that covers the seeds related to object $c$, and $\mathcal{D}^c$ contains no seeds of it. The baseline CAM (left) can merely recognize part of $\mathcal{A}^c$ regions, while the new CAM (right) by the CP Pair can find $\boldsymbol{New}\, \mathcal{A}^c$.}
\label{fig:region}
\end{figure}

According to the previous experiments of the region and erasing methods~\cite{has,erase1}, both $\boldsymbol{I}_h$ and $\boldsymbol{I}_{\overline h}$ may individually help CAM dig out more potential areas. However, some parts in $\boldsymbol{I}$ are apparently ignored if simply using $\boldsymbol{I}_h$ or $\boldsymbol{I}_{\overline h}$ in each training epoch since each image could only be used once during one epoch. To make full use of the information in $\boldsymbol{I}$, here we propose to use the CP Pair to find more seeds in $\boldsymbol{Y}^c$.
For a partitioned image in Fig. \ref{fig:region}, suppose $\boldsymbol{I}$ is split into $N$ patches and there are two kinds of patch regions for the $c$-th class, $\Omega=\{\mathcal{A}^{c}_{i}\}_{i=1}^{N_a}$ and $\Gamma=\{\mathcal{D}^{c}_{j}\}_{j=1}^{N_d}$, where $N_a + N_d = N$. $\mathcal{A}^{c}$ represents the patch region that contains the seeds of the object $c$, while $\mathcal{D}^{c}$ covers no seeds related to it. We denote $p_{c}(x), x \in \{\Omega \cup \Gamma\}$ as the probability function of finding $c$-th seeds in $x$. Higher $p_{c}(x)$ refers to more seeds in patch $x$ of $\boldsymbol{I}$. Therefore, we have $\sum\nolimits_{i = 1}^{{N_{a}}} {{p_{c}}(x = {\mathcal{A}^{c}_{i}})}  = 1$ and $p_{c}(x=\mathcal{D}^{c}_{j})  = 0, j \in \{1,2,...,N_d\}$. Under the definition above, the self information of $\boldsymbol{Y}^{c}$ is denoted as $\mathcal{H}(\boldsymbol{Y}^{c})$ and expressed as:
\begin{equation}
\mathcal{H}(\boldsymbol{Y}^{c}) =  - \sum\limits_{x \in {\Omega}_y} \log ({p_{c}(x)}), 
\end{equation}
where ${\Omega}_y \subset {\Omega}$ refers to the set of $\mathcal{A}^{c}$ in $\boldsymbol{Y}^{c}$. Note that the ground truth of the $c$-th class contains all patch $\mathcal{A}^{c}$, so it is said to have the maximum information. Our aim is to increase $\mathcal{H}(\boldsymbol{Y}^{c}) $ by increasing $|{\Omega}_y|$.

Meanwhile, let $\mathcal{H}(\boldsymbol{Y}_{h}^{c})$ and $\mathcal{H}(\boldsymbol{Y}^{c}_{\overline h})$ be the information of the CAMs generated by the CP Pair, which can be jointly calculated as follows:
\begin{align}
\mathcal{H}(\boldsymbol{Y}_{q}^{c}) =   - \sum\limits_{x \in {\Omega}_q} \log ({p_{c}(x))}, \; q \in \{h,\overline h\},
\end{align}
where ${\Omega}_h \, ({\Omega}_{\overline h}) \subset {\Omega}$ is the set of $\mathcal{A}^{c}$ in $\boldsymbol{Y}_{h}^{c}$ ($\boldsymbol{Y}_{\overline h}^{c}$). Randomly covering parts of the object $c$ in $\boldsymbol{I}$ leads to an increase of $\boldsymbol{New} \, \mathcal{A}^{c}$. Unfortunately, it is undecidable to straightly compare $|{\Omega}_y|$ with $|{\Omega}_h|$ or $|{\Omega}_{\overline h}|$, because some discriminative parts in ${\Omega}_y$ are possibly hidden in $\boldsymbol{Y}^{c}_{h}$ or $\boldsymbol{Y}^{c}_{\overline h}$. Due to the complementary attribution (${\Omega}_h  \cap {\Omega}_{\overline h}=\varnothing$), however, the sum of $\mathcal{A}^{c}$ in $\boldsymbol{Y}^{c}_{h}$ and $\boldsymbol{Y}^{c}_{\overline h}$ contains the original high response regions from the baseline CAM, and the $\boldsymbol{New} \, \mathcal{A}_{c}$ regions sought by the CP Pair. Therefore, we have:
 \begin{align}
 {\Omega}_h  \cup {\Omega}_{\overline h} = {\Omega}_y \cup {\Omega}^{\prime},
 \label{eq_n}
\end{align}
where ${\Omega}^{\prime} \subset {\Omega}$ refers to the set of $\boldsymbol{New}  \, \mathcal{A}^c$ and $ {\Omega}^{\prime} \cap {\Omega}_{y} = \varnothing$. Note that ${\Omega}^{\prime} = \varnothing$ if one of the following extreme conditions holds:

1) $\boldsymbol{I}=\boldsymbol{I}_{h}$ or $\boldsymbol{I}=\boldsymbol{I}_{\overline h}$. One of the CP Pair is equal to the original image when the patch size equals the image size;

2) $\boldsymbol{Y}^{c}_{h} \approx \boldsymbol{Y}^{c}_{\overline h} $. It is difficult for the classification net to discriminate the CP Pair if the patch size is too small, resulting in ${\Omega}_h = {\Omega}_{\overline h} = {\Omega}_{y}$.

Based on (\ref{eq_n}), we have that:
\begin{equation}
\begin{aligned}
&\mathcal{H}(\boldsymbol{Y}^{c}_{h})+ \mathcal{H}(\boldsymbol{Y}^{c}_{\overline h}) \\
&= - \sum\limits_{x \in {\Omega}_h} \log ({p_{c}(x)}) - \sum\limits_{x \in {\Omega}_{\overline h}}\log ({p_{c}(x)})\\
&= - \sum\limits_{x \in {\Omega}_y} \log ({p_{c}(x)}) - \sum\limits_{x \in {\Omega}^{\prime}}\log ({p_{c}(x)})
\geq \mathcal{H}(\boldsymbol{Y}^{c}).
\end{aligned}
\label{eg_H}
\end{equation}

According to (\ref{eg_H}), it is concluded that, except for two extreme cases, the sum of CAMs by the CP Pair is able to find more foreground seeds than $\boldsymbol{Y}^{c}$. To achieve an improved CAM, we propose the CP regularization with a pair of chosen parameters ${\lambda} \in [0,1], {\overline \lambda} = 1- {\lambda}$ as follows:
\begin{equation}\label{eq3}
||({\lambda}\boldsymbol{Y}_{h}^{c}+ {\overline \lambda}\boldsymbol{Y}^{c}_{\overline h}) - \boldsymbol{Y}^{c}||_1.
\end{equation}

Denote the number of patches hidden in $\boldsymbol{I}_{h}$ as $N_h$. Then ${\lambda}$ can be obtained by ${\lambda} = 1 - N_h / N$, meaning that the weight is decided by the quantity of uncovered pixels in $\boldsymbol{I}_{h}$.
To corporate (\ref{eq3}) on the original classification net, we turn to a shared-weights triplet network as shown in Fig. \ref{fig:datu}, namely CP Network (CPN). One branch deals with input $\boldsymbol{I}$ and outputs $\boldsymbol{Y}^{c}$, while the other two branches respectively generate $(\boldsymbol{Y}^{c}_{h}, \boldsymbol{Y}^{c}_{\overline h})$ made by the CP Pair. Note that we stop the gradient update for $(\boldsymbol{Y}^{c}_{h}, \boldsymbol{Y}^{c}_{\overline h})$ to push $\boldsymbol{Y}^{c}$ to approximate the better one. In this way, these three outputs are supposed to be regularized by (\ref{eq3}).
\subsection{Complementary Patch Strategies} \label{sec_strategy}
Grid Patch~\cite{has, REDA} is a common method that can be applied to generate a bunch of $\boldsymbol{I}_h$ for $\boldsymbol{I}$. Specifically, a grid patch with a fixed size of $S \times S \times 3$ can partition $\boldsymbol{I}$ into $H \times W /(S \times S)$ patches. Then each patch hidden with a probability $p_h = 0.5$ is fed into the classification net. Following~\cite{has}, $S$ is evenly chosen from a set $K$ of fixed numbers to fit the size of different objects. To guarantee the same data distribution between the training and the testing sets, the value of the hidden pixels is set to equal the mean RGB values of the images among the whole training sets.

On the other hand, the super-pixel region contains rich information about an image. Therefore, we also propose a Super-pixel Patch strategy, which uses the super-pixels generated by SLIC~\cite{slic} as the patches. Here the number of the super-pixels depends on a predefined segment number, denoted as $S_N$. Note that we test these two patch strategies respectively in experiments.
\subsection{Modules in CPN}
We propose the CP Representation to help CAM find more foreground seeds. However, the regularization in (\ref{eq3}) is incapable of sufficiently improving the CAM merely by using the output of the typical network. To further refine the original CAMs, ~\cite{seam} proposes a modified self-attention module named Pixel Correlation Module (PCM), capturing contextual information by taking advantage of pixel relationships in feature maps. Here we take a brief introduction to the self-attention module~\cite{nonlocal}, which can be normally expressed as:
 \begin{equation}\label{n1}
{\boldsymbol{Y}_{out}} = \frac{\mu(\boldsymbol{X_{in}}) \mathcal{J} (\boldsymbol{X_{in}})}{\sum\nolimits_{i = 1}^{HW} {\sum\nolimits_{j = 1}^{HW}\mathcal{J}(\boldsymbol{X_{in}})_{ij}}} + \boldsymbol{X_{in}},
\end{equation}
\begin{equation}\label{n2}
\mathcal{J}(\boldsymbol{X_{in}}) = e^{g(\boldsymbol{X_{in}})^\mathsf{T} \delta(\boldsymbol{X_{in}})},
\end{equation}
where $\boldsymbol{X}_{in}$ and $\boldsymbol{Y}_{out}$ are respectively the input and output feature. $\mathcal{J}$ is used for measuring the relationship between the adjacent pixels, and $\mu$ provides a representation of each pixel in $\boldsymbol{X}_{in}$. Specially, $\mu, \delta$ and $g$ are implemented by is implemented by a $1 \times 1$ convolution layer.

Based on (\ref{n1}) and (\ref{n2}), the PCM refines the CAM $\boldsymbol{Y} \in \mathbb{R}^{C \times HW}$ (flattened into matrix formats) as:
\begin{equation}\label{eq4}
{\boldsymbol{Y}_{pcm}} = \frac{\boldsymbol{Y}\mathcal{J}(\boldsymbol{X})}{\sum\nolimits_{i = 1}^{HW} {\sum\nolimits_{j = 1}^{HW}\mathcal{J}(\boldsymbol{X})_{ij}}},
\end{equation}
\begin{equation}\label{eq5}
\mathcal{J}(\boldsymbol{X}) = {\mathop{\rm ReLU}}(\frac{g(\boldsymbol{X})^\mathsf{T}g(\boldsymbol{X})} {{||g(\boldsymbol{X})||_1}^2}),
\end{equation}
where $\boldsymbol{X} \in \mathbb{R}^{C1 \times HW}$ is the aggregation of some features in the classification net, and $\mathcal{J}:\mathbb{R}^{C_1}\mapsto\mathbb{R}^{HW}$ refers to the cosine distance to measure the inter-pixel feature similarity. Then we can obtain a refined CAM, denoted as ${\boldsymbol{Y}_{pcm}} \in {\mathbb{R}^{C \times H \times W}}$ (reshaped from ${\boldsymbol{Y}_{pcm}} \in {\mathbb{R}^{C \times HW}}$).
\begin{figure}
\begin{center}
\includegraphics[height = 1.8in,width=3.3in]{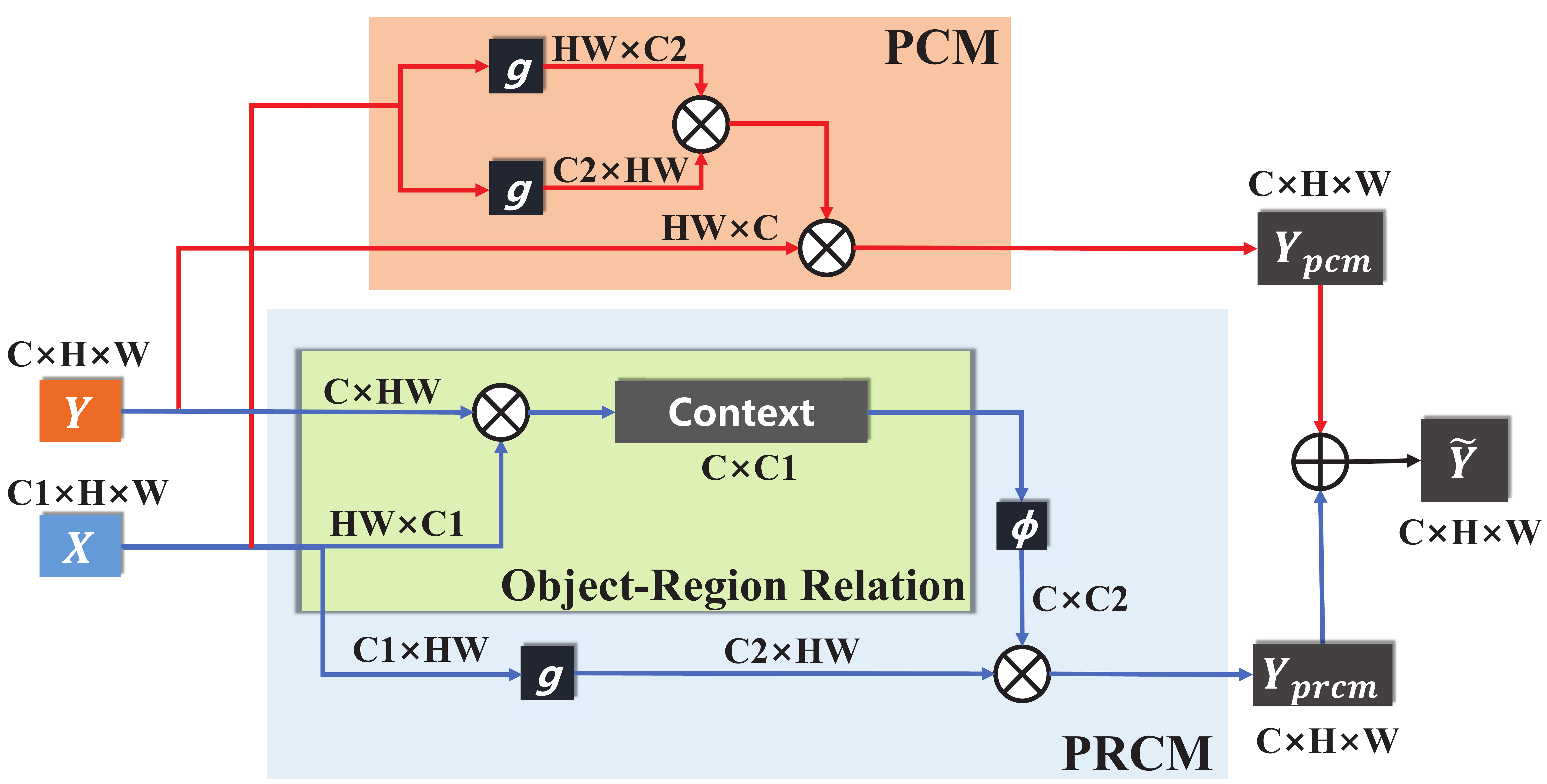}
\end{center}
\vspace{-4mm}
   \caption{The structures of PCM (the red stream) and proposed PRCM (the blue stream). The final refined CAM $\widetilde {\boldsymbol{Y}}$ is the sum of $\boldsymbol{Y}_{pcm}$ and $\boldsymbol{Y}_{prcm}$.}
\label{fig:prcm}
\vspace{-2mm}
\end{figure}

Object Contextual Representation (OCR)~\cite{OCR} is an effective approach to augment the contextual information based on exploring object-pixel relation. Therefore, we propose a Pixel-Region Correlation Module (PRCM) to help further improve the CAMs. Firstly, an Object-Region Relation matrix $\boldsymbol{Z} \in {\mathbb{R}^{C \times C_1}}$ is represented by $\boldsymbol{Z} = \mathop{\rm SoftMax}(\boldsymbol{Y})\boldsymbol{X}^\mathsf{T}$. Here we directly treat $\boldsymbol{Y}$ as soft object regions, which are supposed to be the coarse segmentation maps corresponding to $C$ objects~\cite{OCR}. Then we can obtain a Pixel-Region Relation $ {\boldsymbol{P_{R}}} \in {\mathbb{R}^{C \times H \times W}}$ (reshaped from $ \boldsymbol{P_{R}} \in {\mathbb{R}^{C \times HW}}$) as:
\begin{equation}\label{eq6}
\boldsymbol {P_{R}} = \phi (\boldsymbol{Z})g(\boldsymbol{X}),
\end{equation}
\begin{equation}\label{eq7}
\boldsymbol{Y}_{prcm} = {\boldsymbol{Y}} \circ \mathop{\rm SoftMax}({\boldsymbol{P_{R}}}),
\end{equation}
where $\phi:\mathbb{R}^{C_1}\mapsto\mathbb{R}^{C_2}$ is also an embedding function similar to $g$. Since $\boldsymbol{P_{R}}$ represents the relation between regions in $\boldsymbol{X}$ and pixels in $\boldsymbol{Y}$~\cite{OCR}, we strengthen $\boldsymbol{Y}$ by (\ref{eq7}) to gain the refined CAM, denoted as ${\boldsymbol{Y}_{prcm}} \in {\mathbb{R}^{C \times H \times W}}$.

In order to combine the contextual information collected by PCM and PRCM, the final smoothed CAM, denoted as $\widetilde {\boldsymbol{Y}} \in {\mathbb{R}^{C \times H \times W}}$, is the sum of $\boldsymbol{Y}_{prcm}$ and $\boldsymbol{Y}_{pcm}$. Fig. \ref{fig:prcm} illustrates the structures of PCM and PRCM. Thus, a pair of output CAMs $(\widetilde {\boldsymbol{Y}},\boldsymbol{Y})$ can be readily generated for each branch in CPN. For convenience, we denote the CAMs in branch with $\boldsymbol{I}$ as $(\widetilde {\boldsymbol{Y}}_{o},\boldsymbol {Y}_{o})$, and the CAMs in branch with the CP Pair are respectively denoted as $(\widetilde {\boldsymbol{Y}}_{h},\boldsymbol{Y}_h)$ and $(\widetilde {\boldsymbol{Y}}_{\overline h},\boldsymbol{Y}_{\overline h})$.
\subsection{Loss in CPN}
Following conventions, an additional GAP layer is applied on the CAM, aggregating it into image-level prediction scores $s \in \mathbb{R}^{(C-1) \times 1}$. Note that $s$ simply contains $C-1$ foreground objects since the image-level supervision lacks the background label. Therefore, we can obtain the score maps generated from $\boldsymbol{Y}_o$, $\boldsymbol{Y}_h$ and $\boldsymbol{Y}_{\overline h}$,which are respectively denoted as $s_o$,$s_h$ and $s_{\overline h}$. Then we employ Multi-Label Soft Margin Loss $\boldsymbol{l}_{cls}$ for supervision:
\begin{equation}\label{n_cls}
\mathcal {L}_{cls} =\frac{1} {3}(\boldsymbol{l}_{cls}(s_o) + \boldsymbol{l}_{cls}(s_h) + \boldsymbol{l}_{cls}(s_{\overline h})).
\end{equation}

Meanwhile, the CP Representation is adopted into the CPN for mining out more seeds, which is represented as:
\begin{equation}\label{tcp}
\begin{split}
\mathcal {L}_{tcp} = &||({\lambda}{\boldsymbol{Y}_h} + {\overline \lambda}{\boldsymbol{Y}_{\overline h}}) - {\boldsymbol{Y}_o}||_1 + \\
 &||({\lambda}\widetilde {\boldsymbol{Y}}_{h} + {\overline \lambda}\widetilde {\boldsymbol{Y}}_{{\overline h}}) - \widetilde {\boldsymbol{Y}}_{o}||_1,
 \end{split}
\end{equation}
Here the Triplet CP (TCP) loss, denoted as $\mathcal {L}_{tcp}$, is proposed based on the regularization in (\ref{eq3}). Note that there are six output CAMs in the CPN since each branch posses two of them. Consequently, the TCP loss builds a connection among these six CAMs. Similarly with ~\cite{seam}, to address the problem that $\widetilde  {\boldsymbol{Y}}$ predicts all the pixels as the same class (mostly background), we introduce the CP Cross Regularization (CPCR) loss as:
\begin{equation}\label{cpcr}
\begin{split}
\mathcal {L}_{cpcr} = &||(\boldsymbol{Y}_o - {\lambda}\boldsymbol{Y}_h) - {\overline \lambda}\widetilde {\boldsymbol{Y}}_{{\overline h}}||_1 + \\
 &||(\boldsymbol{Y}_o - {\overline \lambda}\boldsymbol{Y}_{\overline h}) - {\lambda}\widetilde {\boldsymbol{Y}}_{h}||_1,
 \end{split}
\end{equation}
Here we jointly regularize the refined CAMs by the CP Pair to make indirect effects on $\widetilde {\boldsymbol{Y}}_o$. As for the example of regularization of ${\lambda}\widetilde {\boldsymbol{Y}}_{h}$, it might be intuitive to regularize $\widetilde {\boldsymbol{Y}}_{h}$ by ${\boldsymbol{Y}_h}$. However, such direct regularization leads to a further degradation in our early experiments. Therefore, we use the gap between $\boldsymbol{Y}_o$ and ${\lambda}\boldsymbol{Y}_{\overline h}$ to regularize $\widetilde {\boldsymbol{Y}}_{h}$ in light of (\ref{eq3}).

During the training, the background activation map is assessed by:
\begin{equation}\label{bg}
 \boldsymbol{Y}^{c=0}(x,y) = {(1 - \mathop {\max }\limits_{1 \le {\rm{c}} \le C - 1} {\boldsymbol{Y}^{c}}(x,y))^\alpha },
\end{equation}
Here ${ \boldsymbol{Y}^{c}}(x,y)$ is the activation value of category $c$ at the position $(x,y)$ in $ \boldsymbol{Y} \in \{ \boldsymbol{Y}_o,  \boldsymbol{Y}_h,  \boldsymbol{Y}_{\overline h}\})$, and $\alpha$ is a hyper-parameter for adjusting the confidence of background score, which empirically is set to $1$. $ \boldsymbol{Y}$ is firstly normalized by $ \boldsymbol{Y}^c(x,y) =  \boldsymbol{Y}^c(x,y) / max_{x,y}  \boldsymbol{Y}^c(x,y),c \in[1,C-1]$, and all scores irrelevant to ground truth are thresholding to $0$. Finally we concatenate $ \boldsymbol{Y}^{c=0}$ into $ \boldsymbol{Y}^{c}$. During inference, $\widetilde { \boldsymbol{Y}}_{o}$ is used for segmentation, and $\widetilde{ \boldsymbol{Y}}_{o}^{c=0}(x,y)$ is set to a fixed value $\beta$.

In all, the CPN is optimized by the final loss function $\mathcal {L}_{all}$ (\ref{floss}), and empirically we set $w_1=w_2=w_3=1$. Fig. 2 demonstrates the overall CPN framework.
\begin{equation}\label{floss}
\mathcal {L}_{all} =w_1\mathcal {L}_{cls} + w_2\mathcal {L}_{tcp}  + w_3\mathcal {L}_{cpcr}.
\end{equation}
\section{Experiments}
\subsection{Implementation Details}
\noindent\textbf{Dataset and evaluated metric:}\; The proposed approach is evaluated on the PASCAL VOC 2012 segmentation benchmark~\cite{voc12}. There are 20 foreground object categories and 1 background annotated in the dataset. Following conventions, the number of training images is 10,582. The validation dataset contains 1,449 images and the test one has 1,456 samples. During the whole training process, only image-level annotation is provided. To measure the performance of all experiments, the mean Intersection-over-Union (mIoU) is used as the evaluation metric.

\noindent\textbf{Network settings:}\; We adopt the ResNet38~\cite{resnet38}, as one of the prevailing models in most WSSS frameworks, as the backbone of the CPN. The parameters trained on the ImageNet~\cite{imagenet} are used for the initialization of the CPN. Following the previous work, we remove the final GAP layer and fully-connected layer, and replace the last three convolution layers with the atrous convolutions with the adapted dilation rates, so that the output stride of the net is 8. According to~\cite{seam}, for the aggregated features $\boldsymbol{X}$ in PCM and PRCM, we firstly extract feature maps from stage 3 and 4, and then jointly decrease their channels into 64 and 128 by use of $1\times1$ convolution layers. Lastly, we concatenate these features and input images to form $\boldsymbol{X}$.

\noindent\textbf{Training settings:}\; Typical data augmentation on the training set: randomly scaling, color jittering, randomly cropping the images by $448 \times 448$, and horizontal flip. The whole model implemented by Pytorch is trained on 1 RTX 3090 GPU with 24 GB memory. We take a mini-batch size of 4 images to train the CPN for 8 epochs. The initial learning rate is 0.01 and decreases by the poly policy with a decay power of  0.9. We leverage SGD Optimizer using weight decay 0.0005 with momentum 0.9. For each mini-batch, we sort the losses in $\mathcal {L}_{cpcr}$ in descending order, and select the top {\textbf{20\%}} losses as the hard examples for training (Online Hard Example Mining (OHEM)) to further improve the performance. Similarly with the settings of~\cite{seam}, we block the gradients backpropagation stream from the PCM and PRCM to the network, to avoid the mutual interference of CAMs and the refined CAMs.
\subsection{Ablation Studies}
In this section, we aim to certify the effectiveness of CPN. All experimental results are generated from VOC 2012 \textit{train} set. For a fair comparison, the background score $\beta$ is the value that results in the best mIoU of the pseudo labels. Note that the patch strategy in Tab. \ref{tab:scale}-\ref{tab:ablation on ph} is Super-pixel Patch with $S_N = $200.

\noindent\textbf{Improvements on CAM:}\; To improve the performance of the final masks, it is a common way to aggregate prediction maps with different scales. Tab. \ref{tab:scale} shows the mIoU of the segments using the baseline CAM, SEAM~\cite{seam} and our CPN under single- and multi-scale cases. The results show that our CPN presents superior mining ability than the baseline in all different scaling cases. In the multi-scale test, the CPN improves the mIoU over the baseline by nearly \textbf{10\%}. For SEAM, which is implemented by a siamese network with equivariant regularization, we adopt the hyperparameters that achieve the best performance in~\cite{seam}. By adding the PRCM, the new SEAM* outperforms the original one in all scale tests. Compared to SEAM, our framework achieves higher performance (\textbf{57.43\%}) in the multi-scale test.

\begin{figure*}[htbp]
\centering
\begin{minipage}[t]{0.02\linewidth}
        \centering
        {}
        \vspace{-0.8cm}
        {(a)}
        \\
        \vspace{1.1cm}
        {(b)}
        \\
        \vspace{1.1cm}
        {(c)}
        \\
        \vspace{1.0cm}  
        {(d)}
        \end{minipage}%
\subfigure{
    \begin{minipage}[t]{0.123\linewidth}
        \centering
        \includegraphics[width=0.8in,height=0.56in]{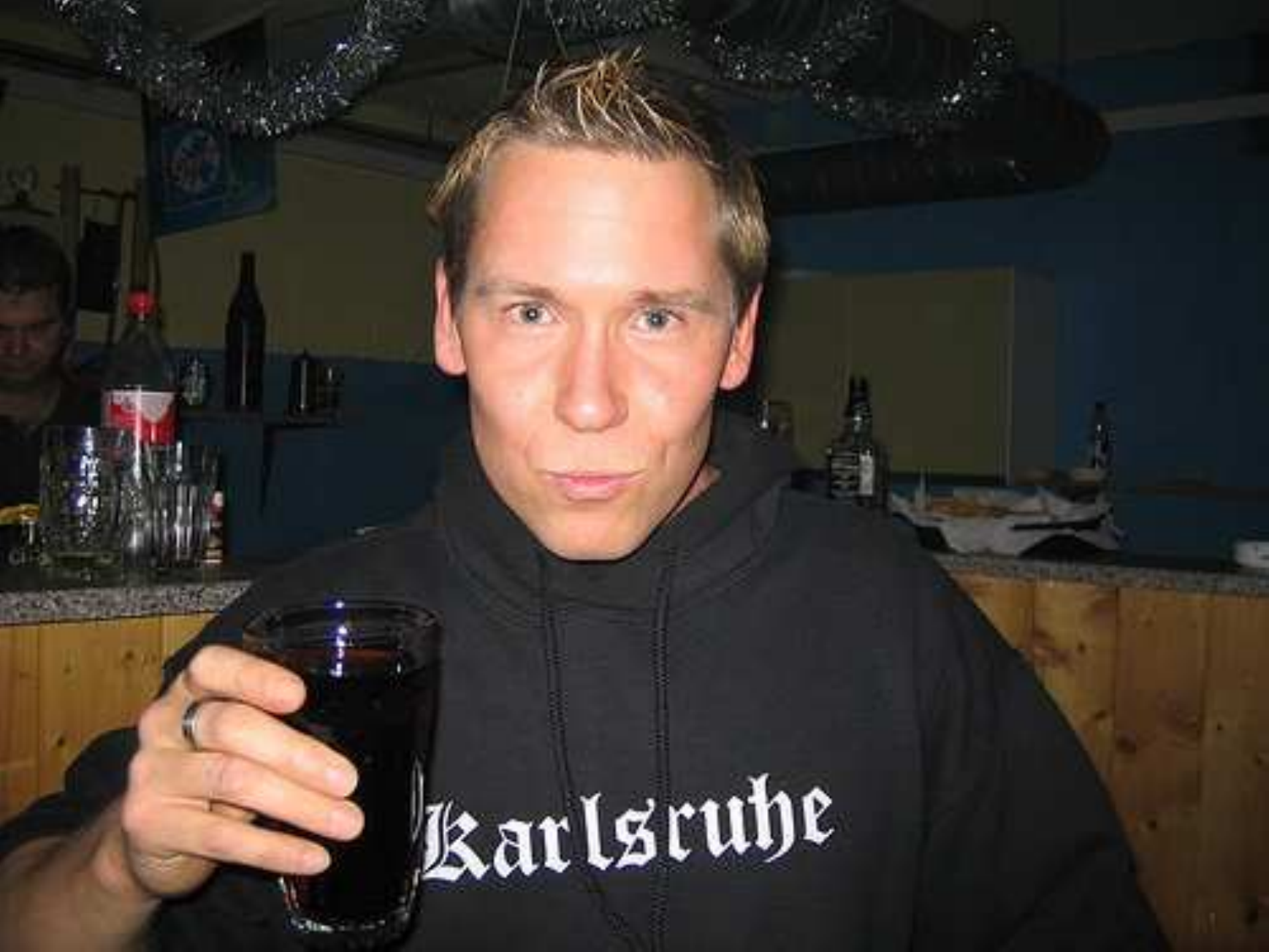}\\   
        \vspace{0.02cm}
        \includegraphics[width=0.8in,height=0.56in]{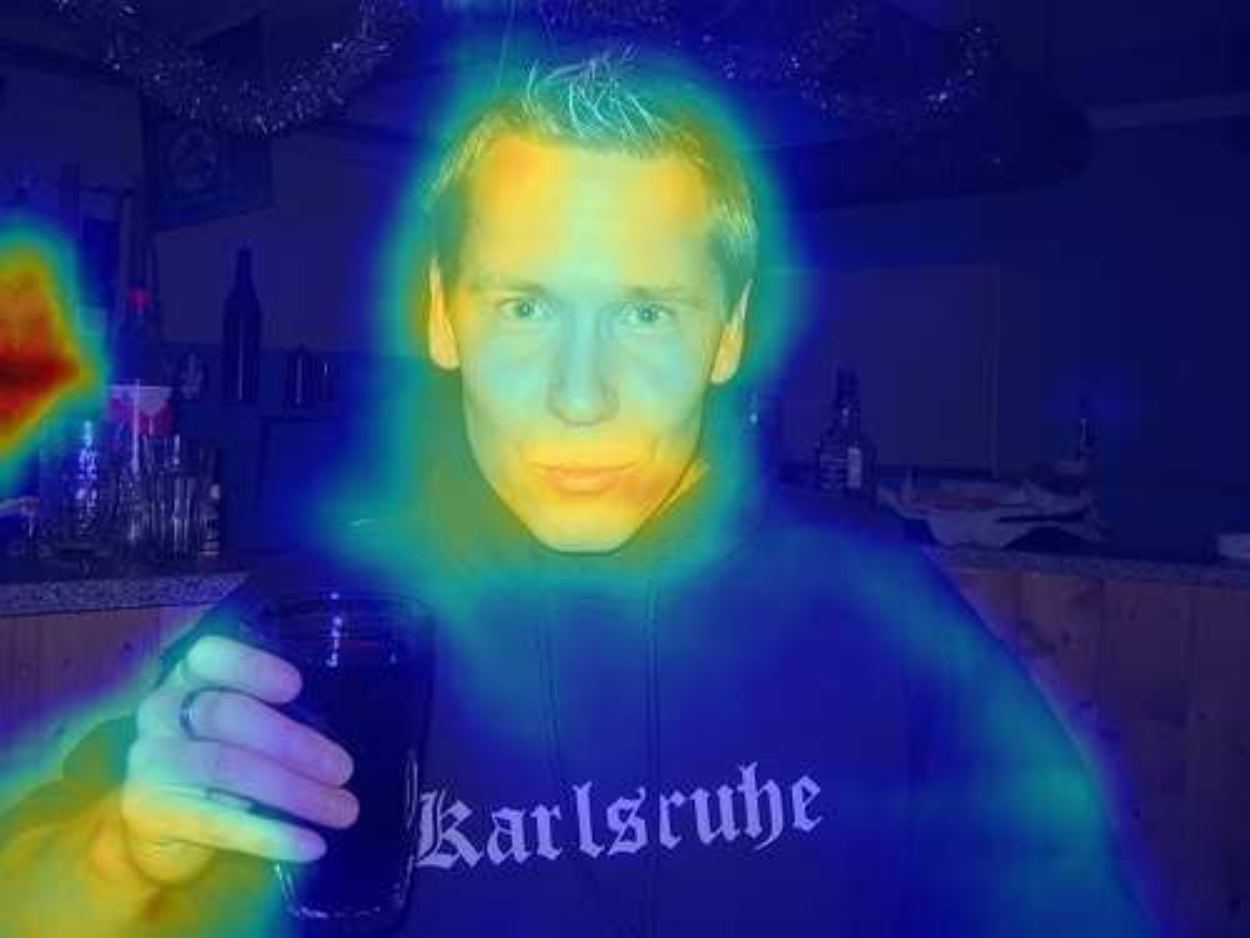}\\
        \vspace{0.02cm}
        \includegraphics[width=0.8in,height=0.56in]{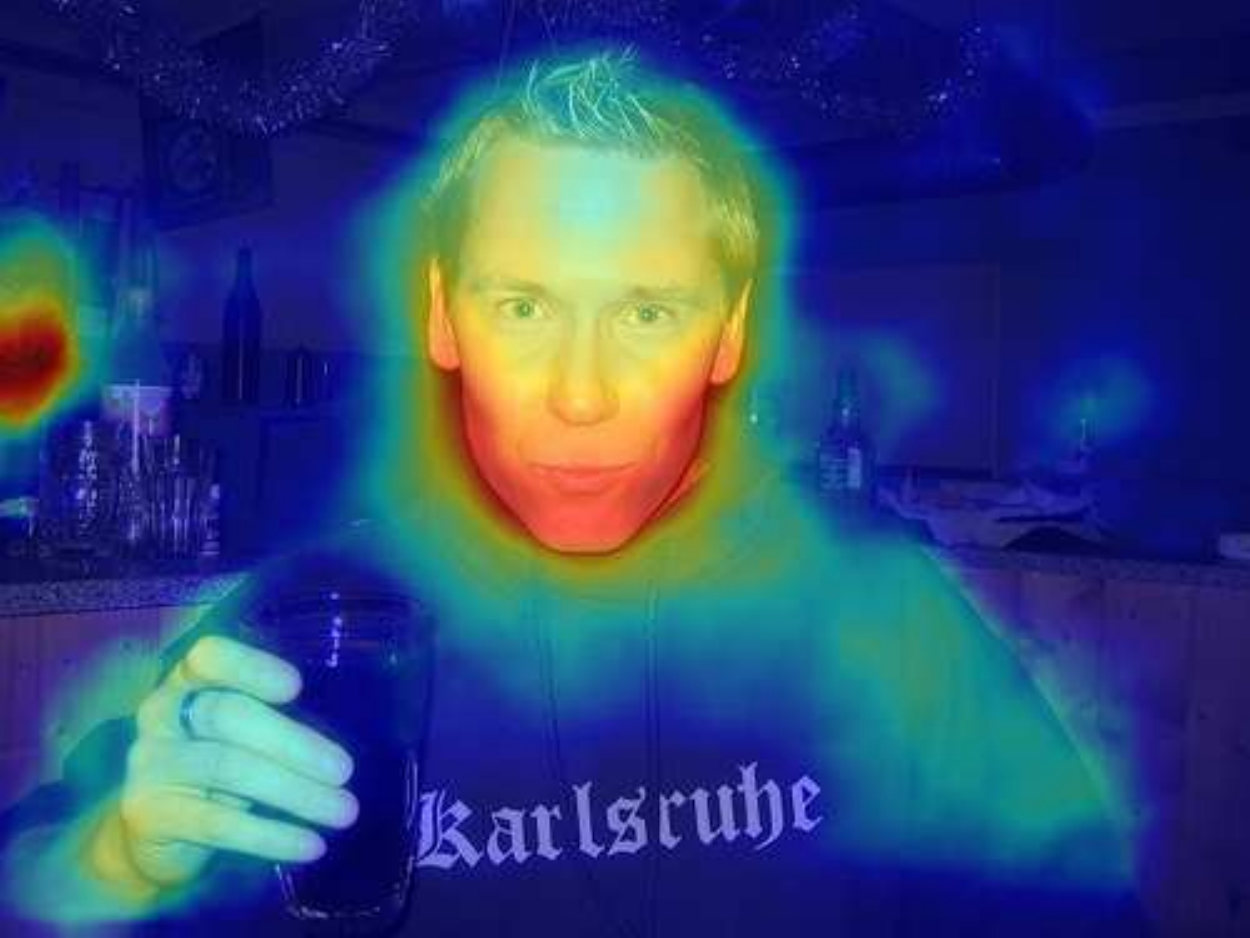}\\
        \vspace{0.02cm}
        \includegraphics[width=0.8in,height=0.56in]{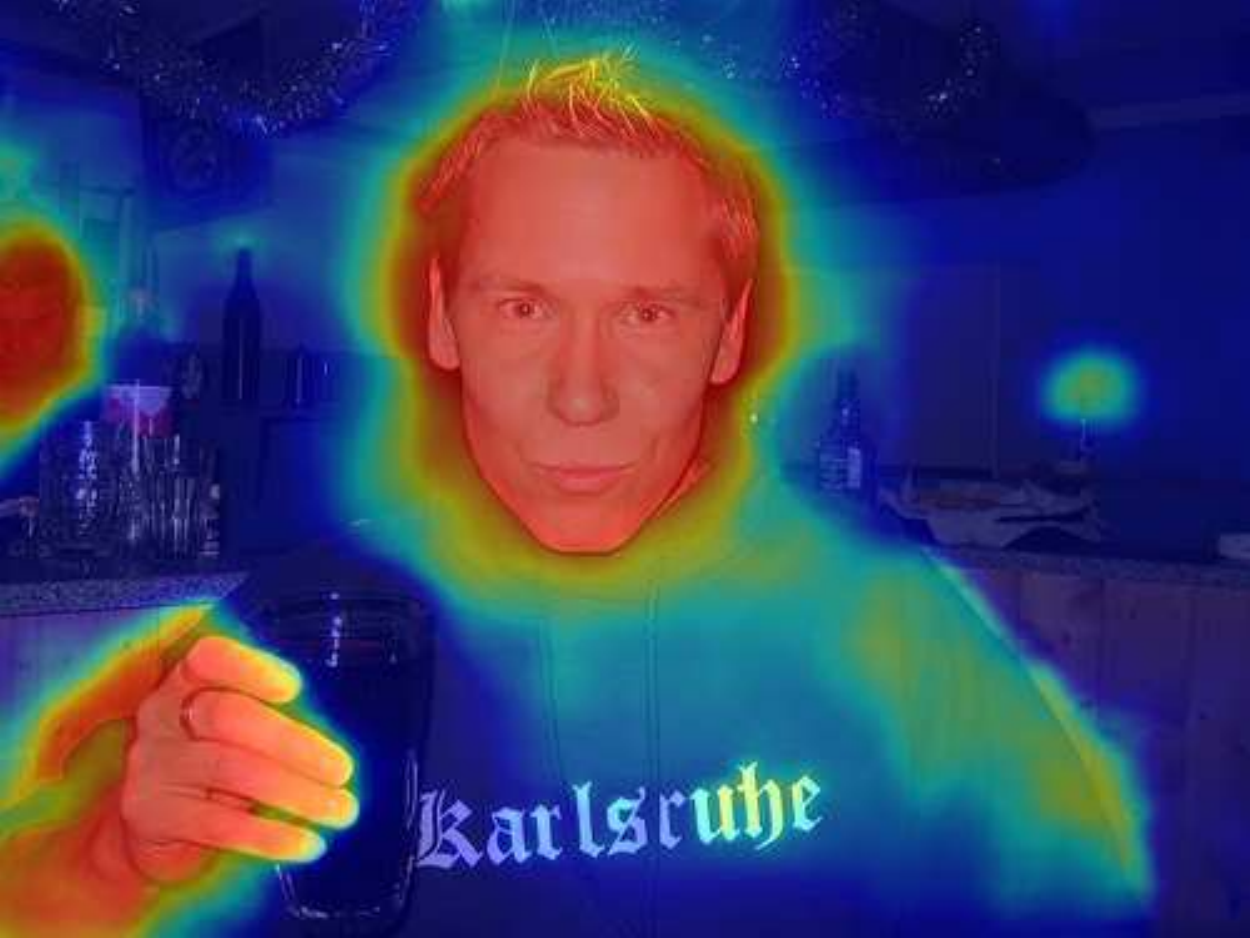}\\
        \vspace{0.02cm}
    \end{minipage}%
}
\hspace{-3mm}
\subfigure{
    \begin{minipage}[t]{0.123\linewidth}
        \centering
        \includegraphics[width=0.8in,height=0.56in]{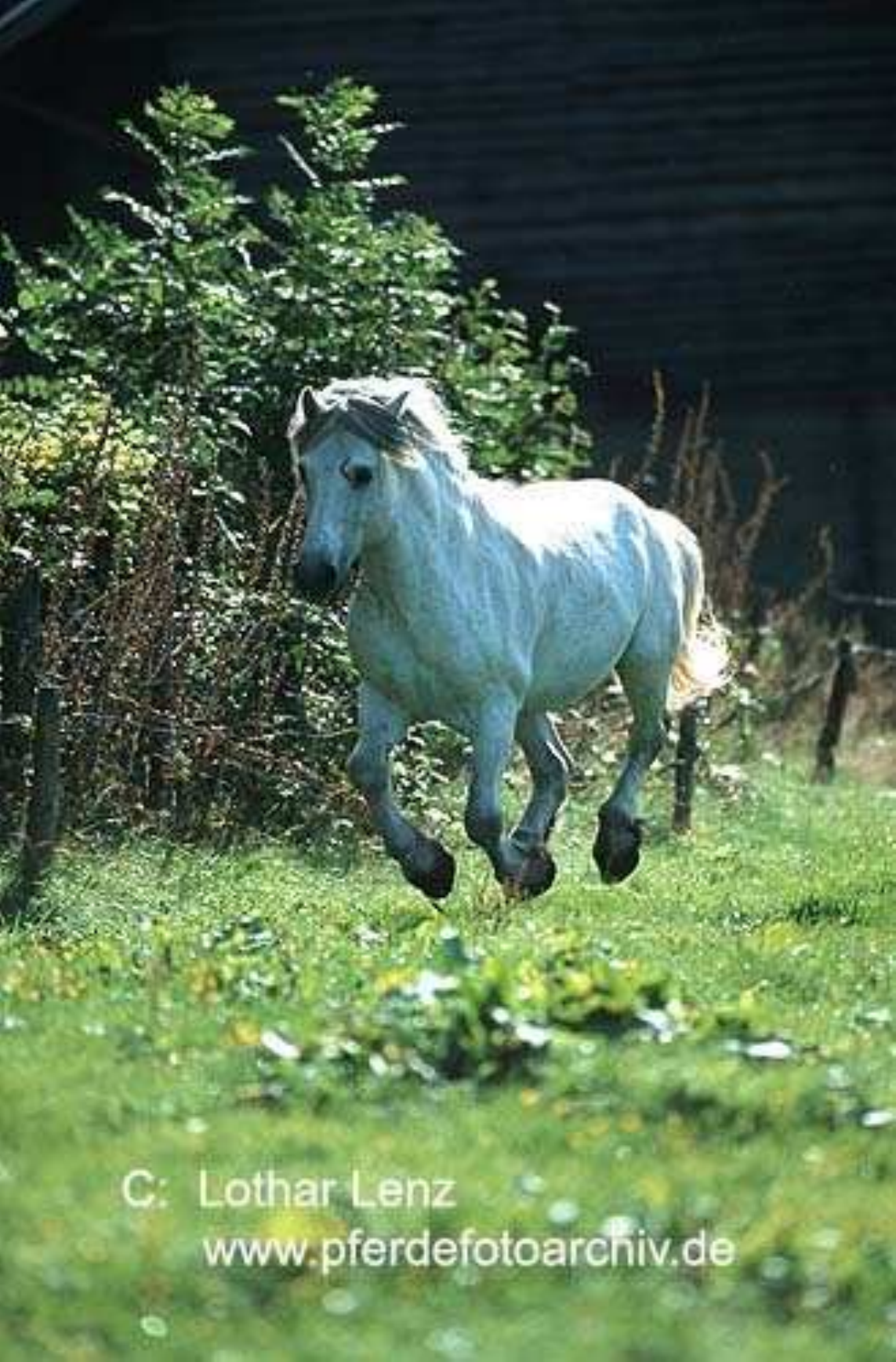}\\
        \vspace{0.02cm}
        \includegraphics[width=0.8in,height=0.56in]{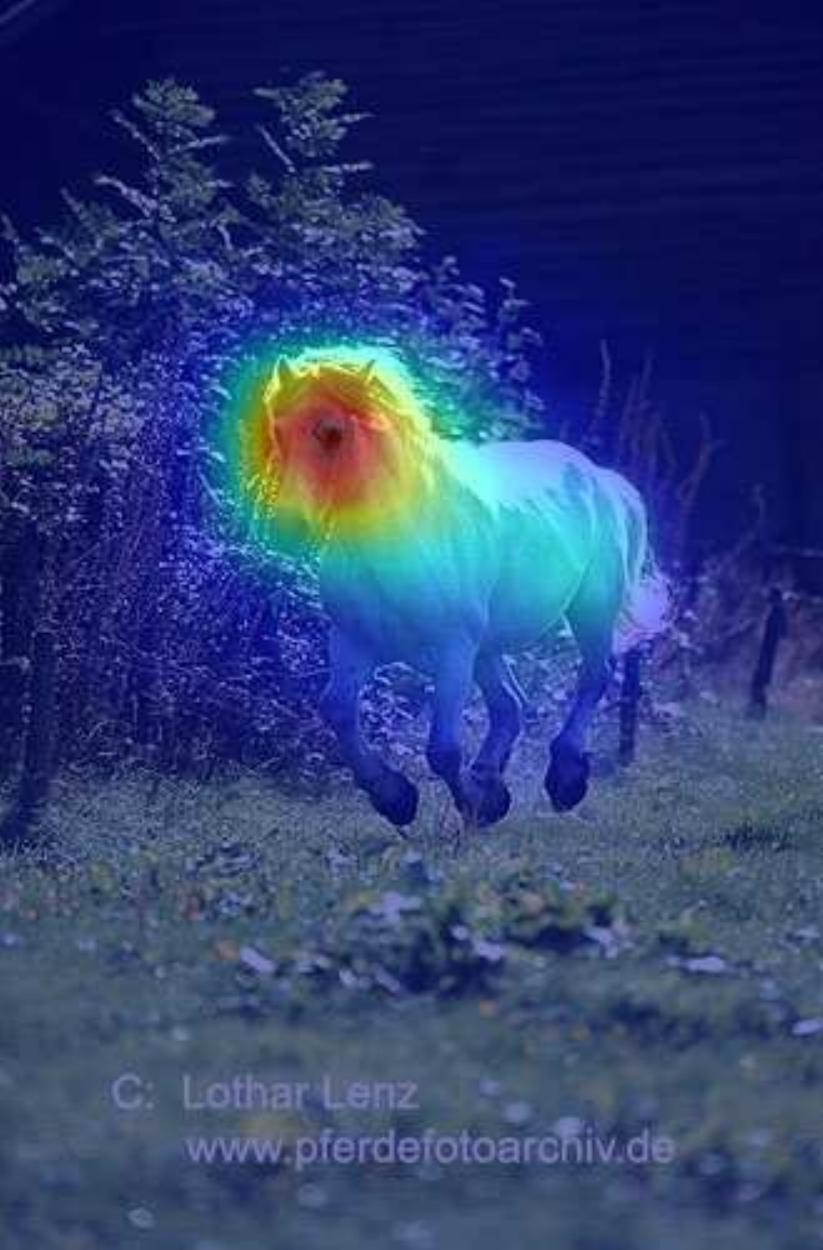}\\
        \vspace{0.02cm}
        \includegraphics[width=0.8in,height=0.56in]{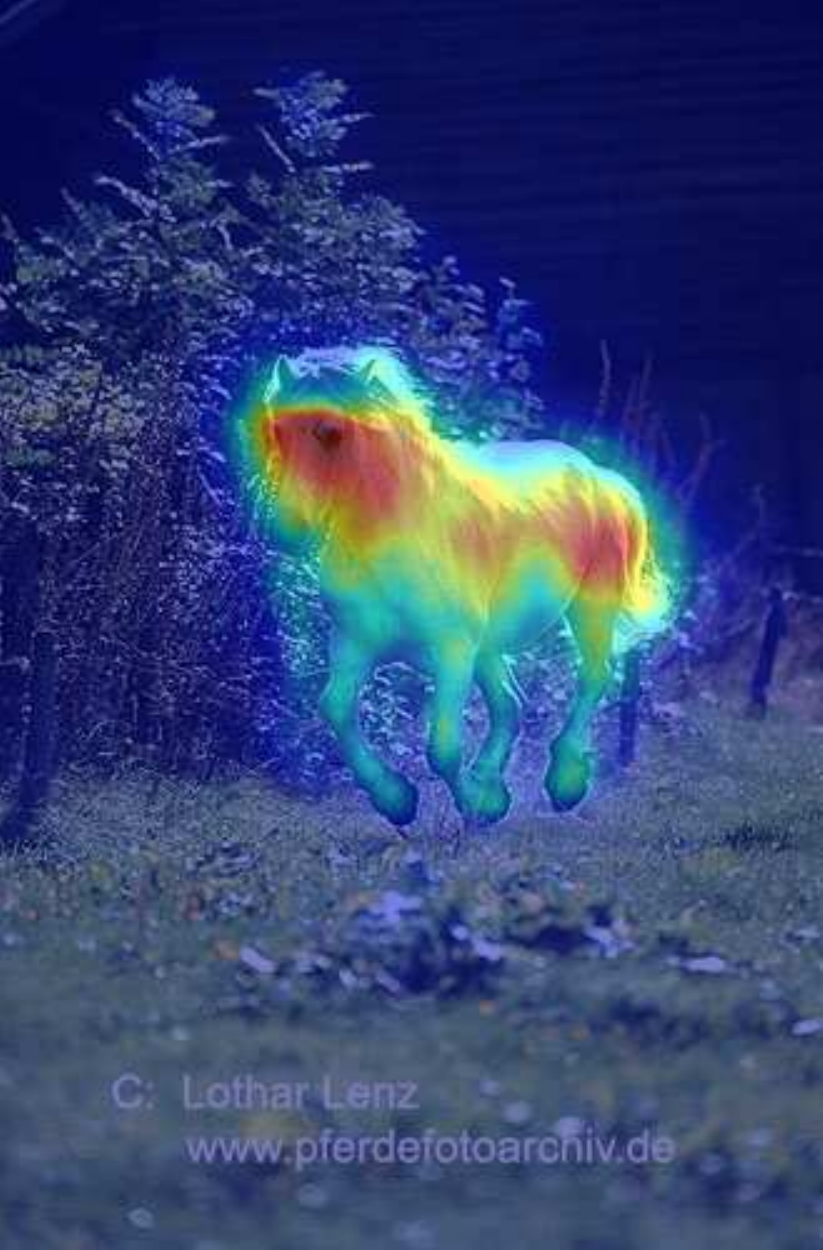}\\
        \vspace{0.02cm}
        \includegraphics[width=0.8in,height=0.56in]{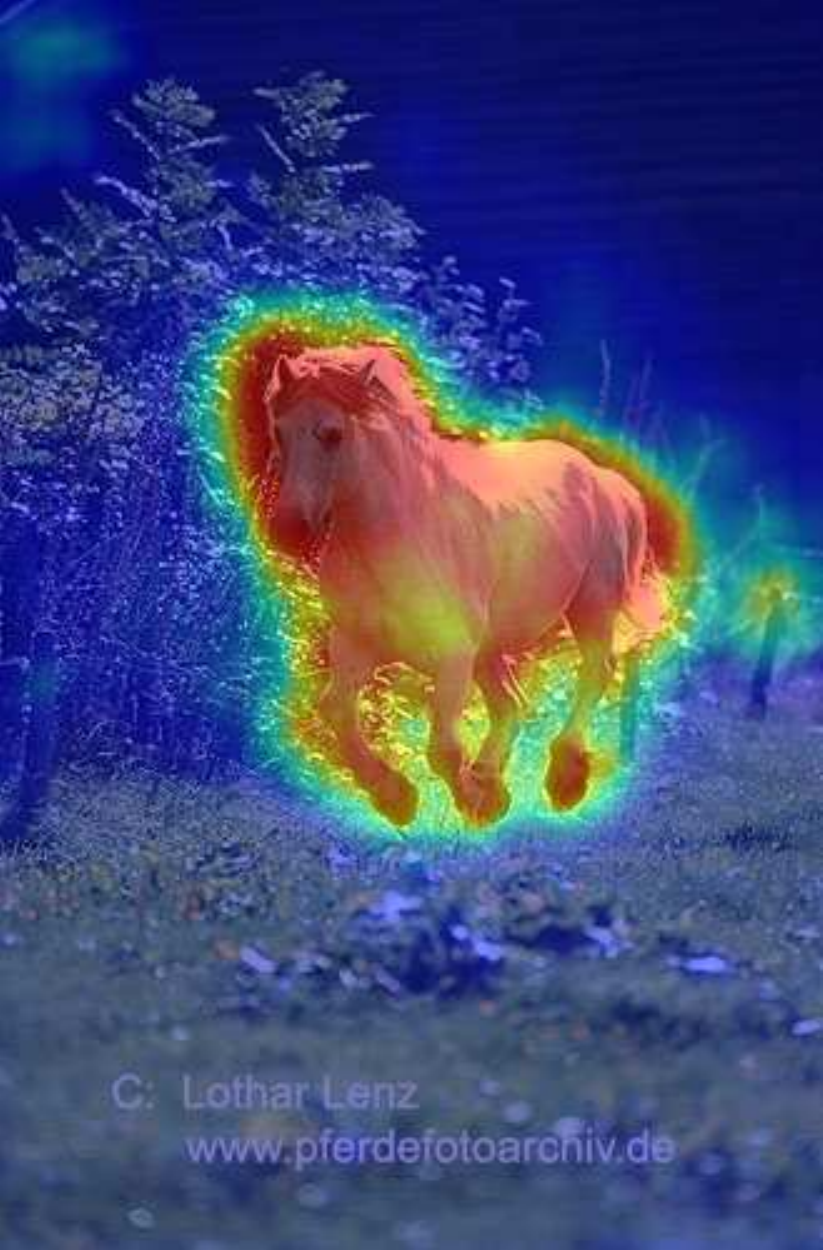}\\
        \vspace{0.02cm}
    \end{minipage}%
}%
\hspace{-2mm}
\subfigure{
    \begin{minipage}[t]{0.123\linewidth}
        \centering
        \includegraphics[width=0.8in,height=0.56in]{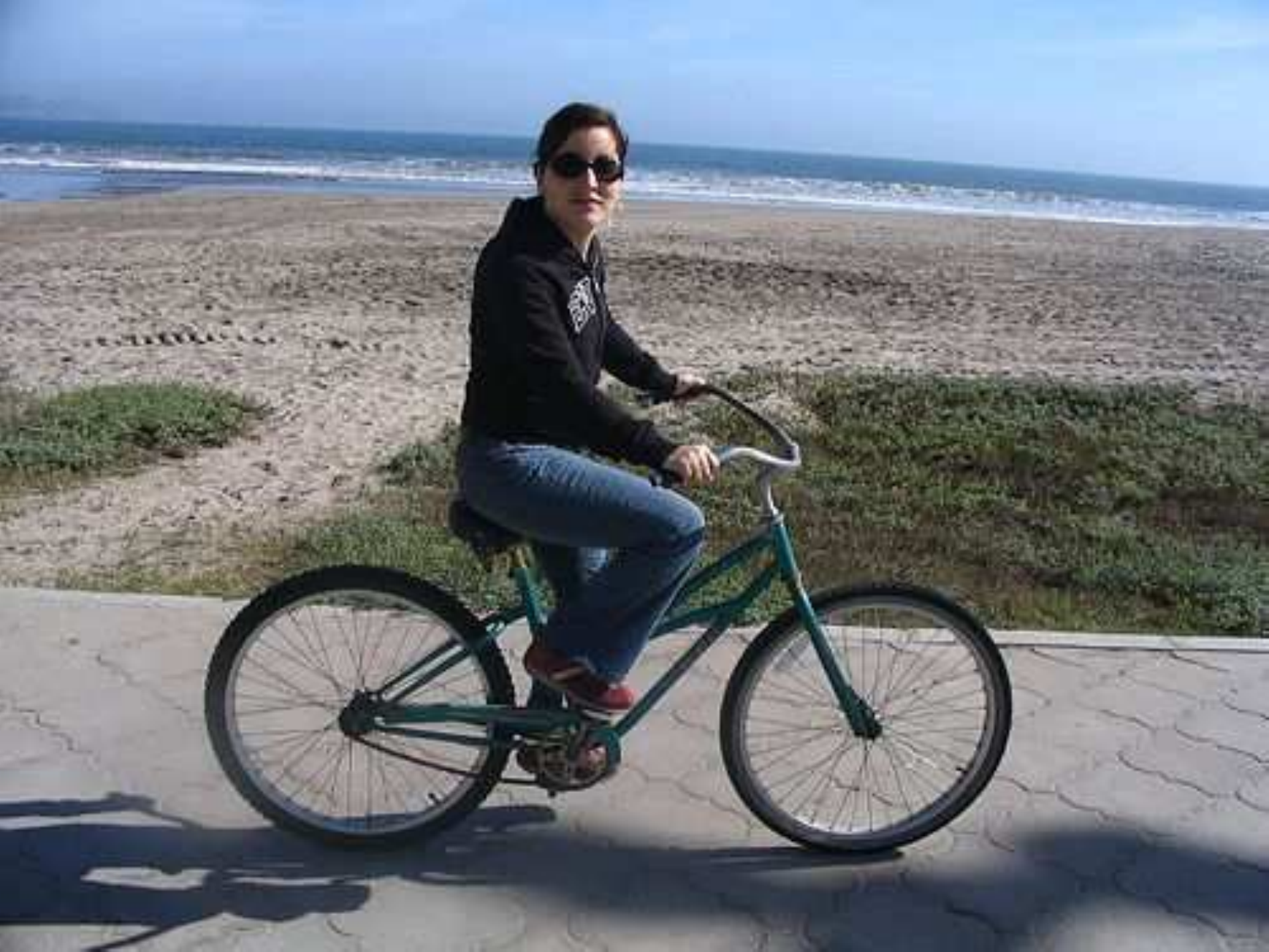}\\
        \vspace{0.02cm}
        \includegraphics[width=0.8in,height=0.56in]{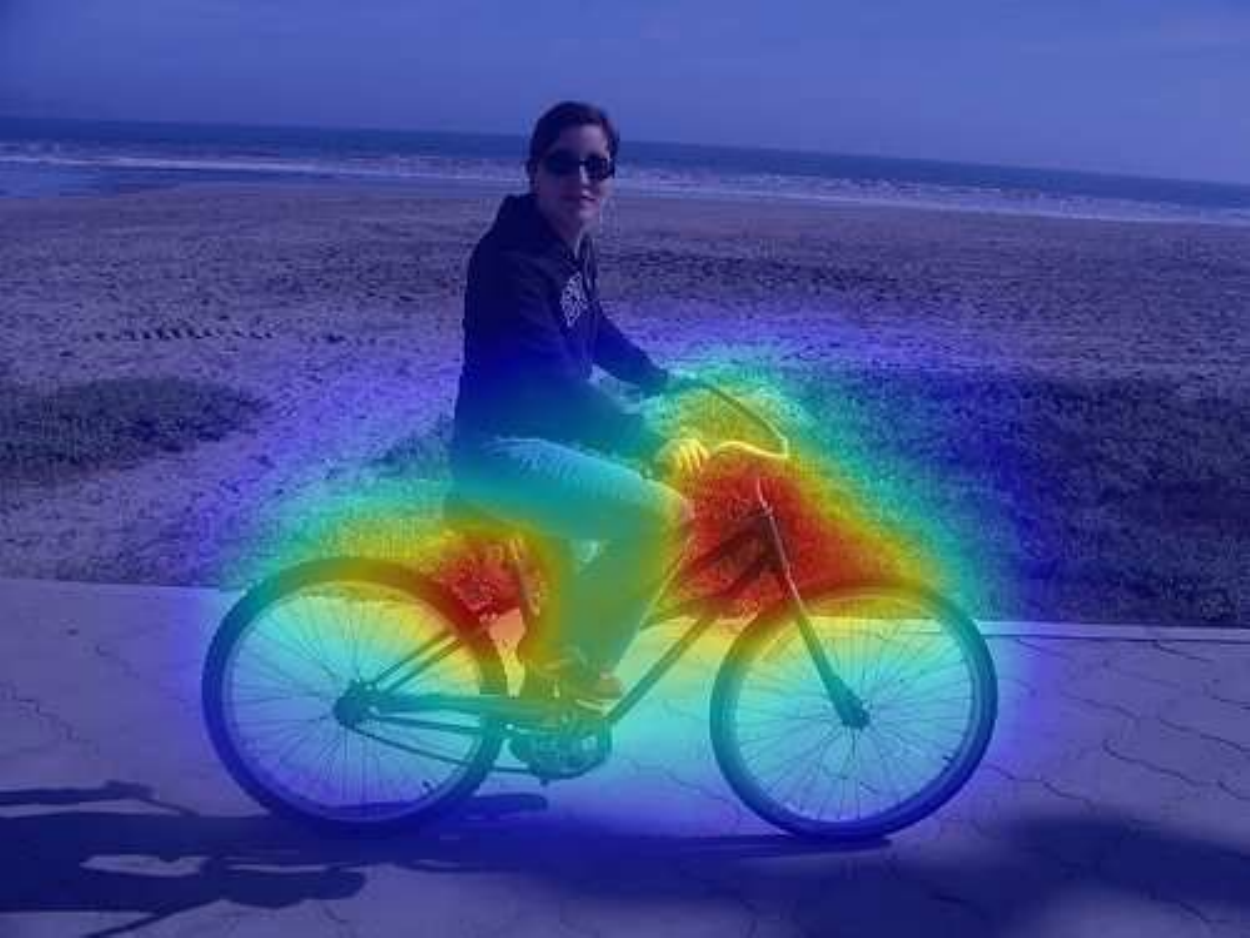}\\
        \vspace{0.02cm}
        \includegraphics[width=0.8in,height=0.56in]{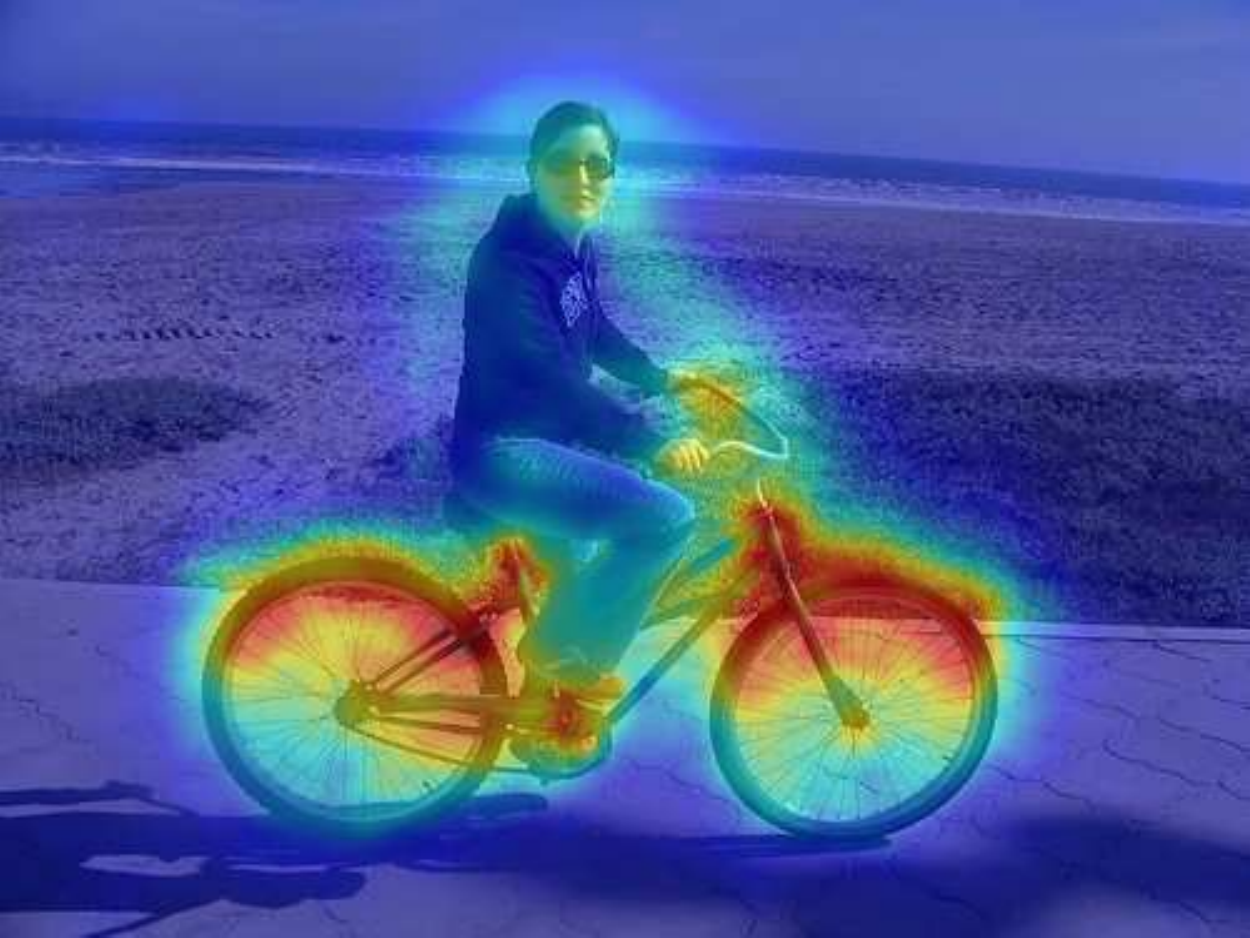}\\
        \vspace{0.02cm}
        \includegraphics[width=0.8in,height=0.56in]{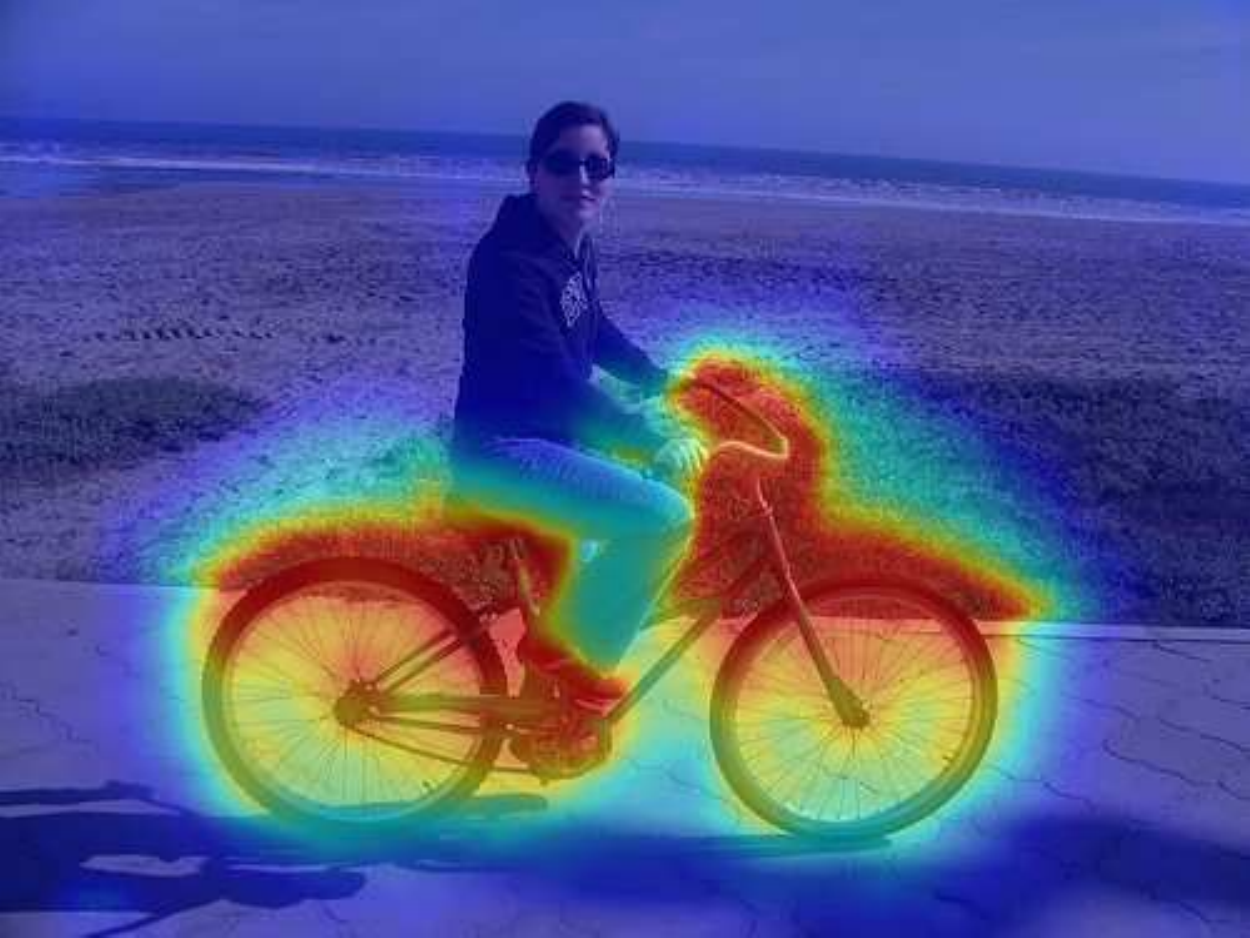}\\
        \vspace{0.02cm}
    \end{minipage}%
}%
\hspace{-2mm}
\subfigure{
    \begin{minipage}[t]{0.123\linewidth}
        \centering
        \includegraphics[width=0.8in,height=0.56in]{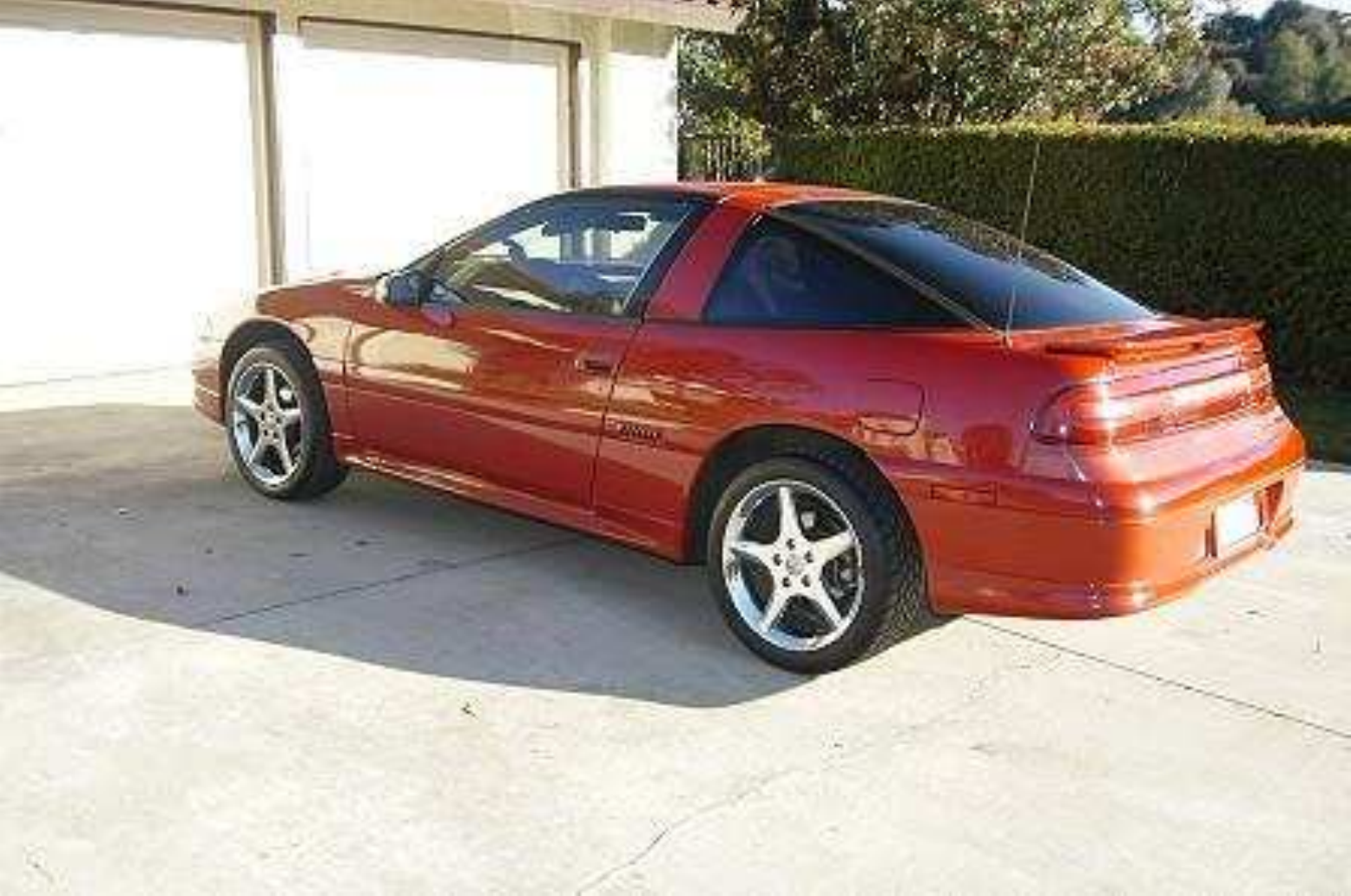}\\
        \vspace{0.02cm}
        \includegraphics[width=0.8in,height=0.56in]{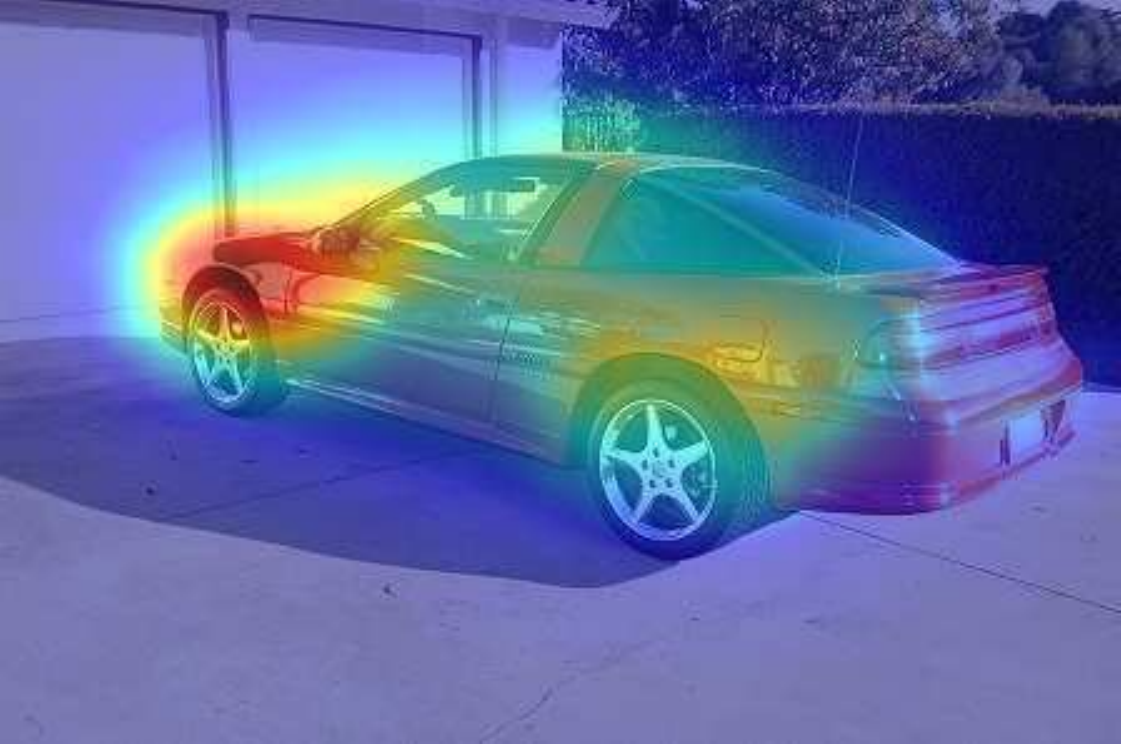}\\
        \vspace{0.02cm}
        \includegraphics[width=0.8in,height=0.56in]{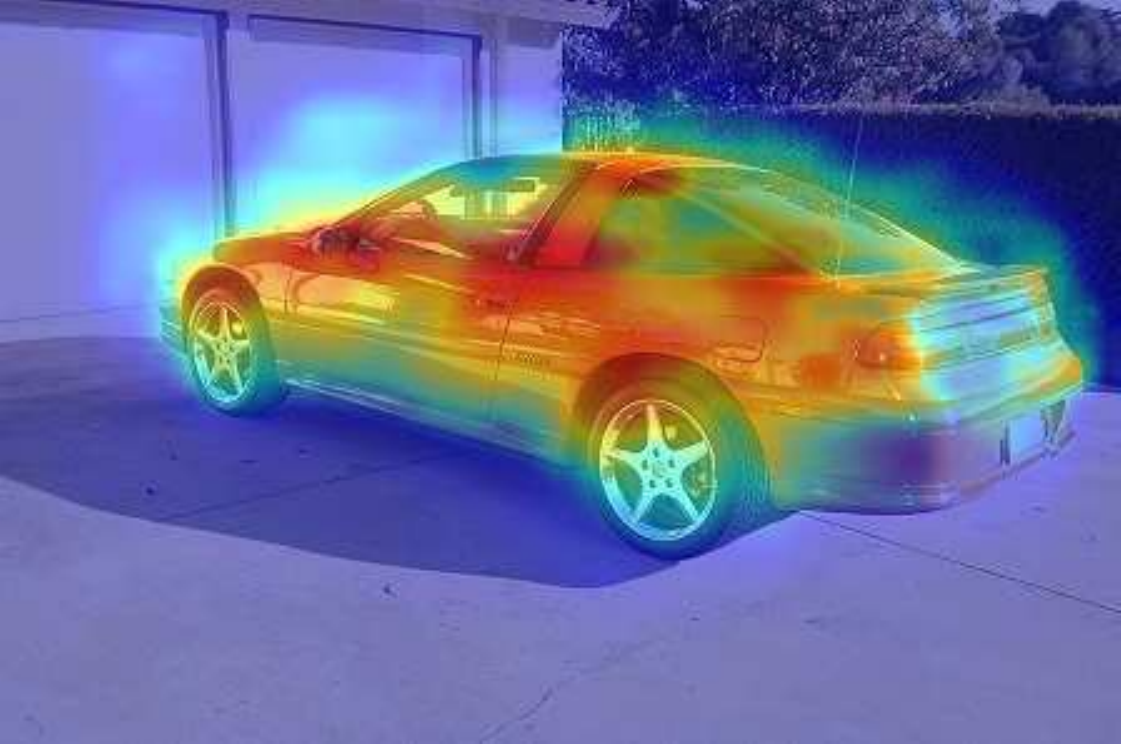}\\
        \vspace{0.02cm}
        \includegraphics[width=0.8in,height=0.56in]{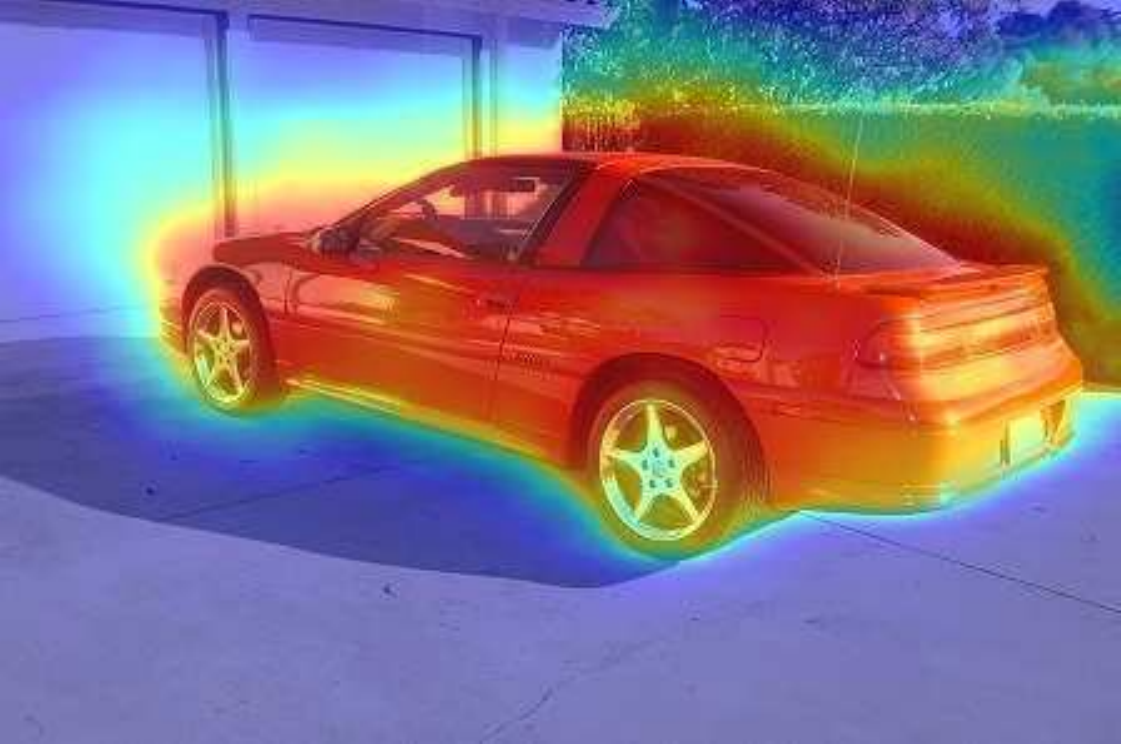}\\
        \vspace{0.02cm}
    \end{minipage}%
}%
\hspace{-2mm}
\subfigure{
    \begin{minipage}[t]{0.123\linewidth}
        \centering
        \includegraphics[width=0.8in,height=0.56in]{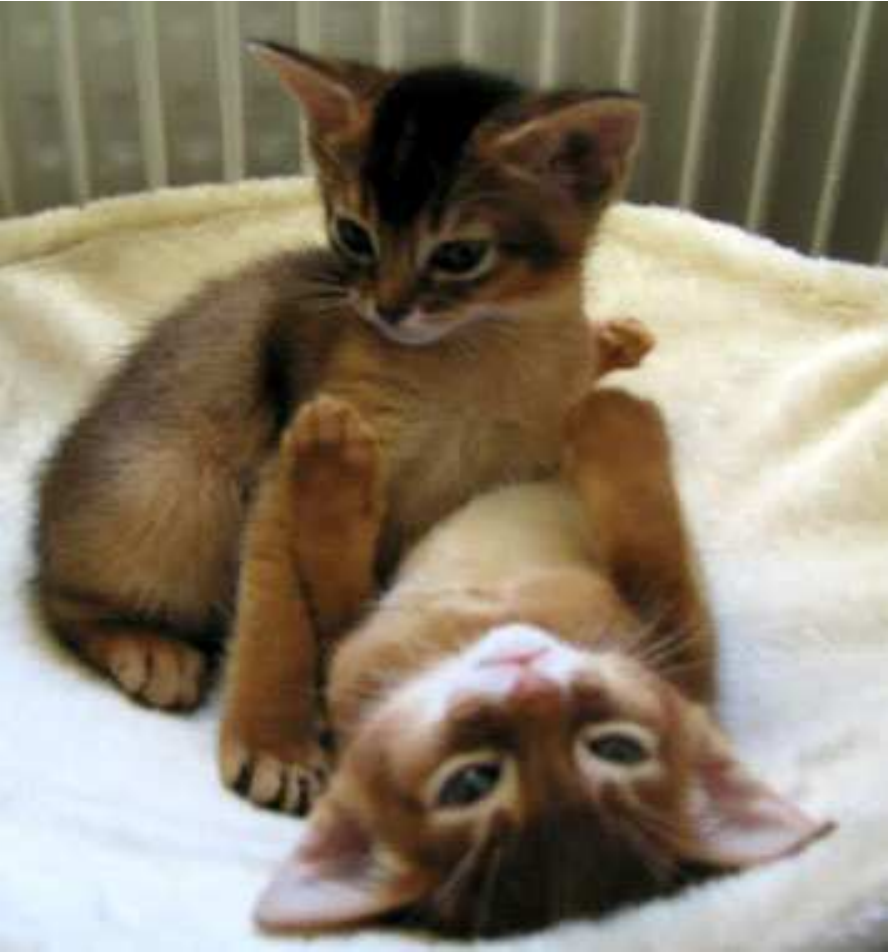}\\
        \vspace{0.02cm}
        \includegraphics[width=0.8in,height=0.56in]{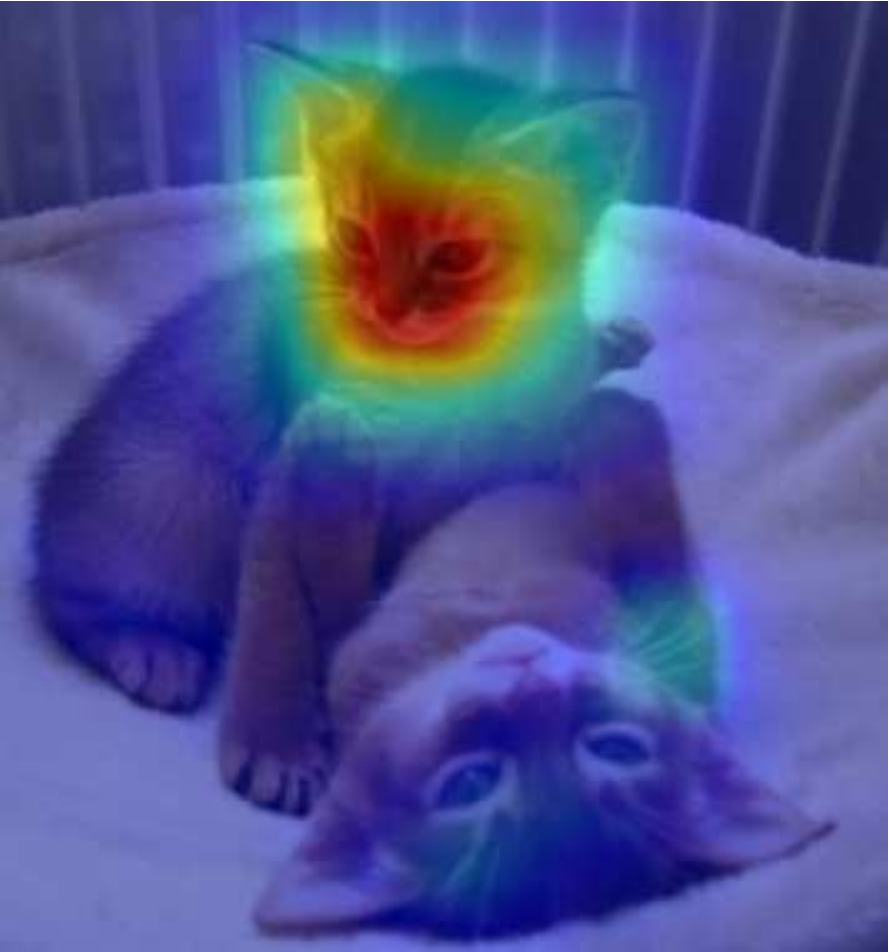}\\
        \vspace{0.02cm}
        \includegraphics[width=0.8in,height=0.56in]{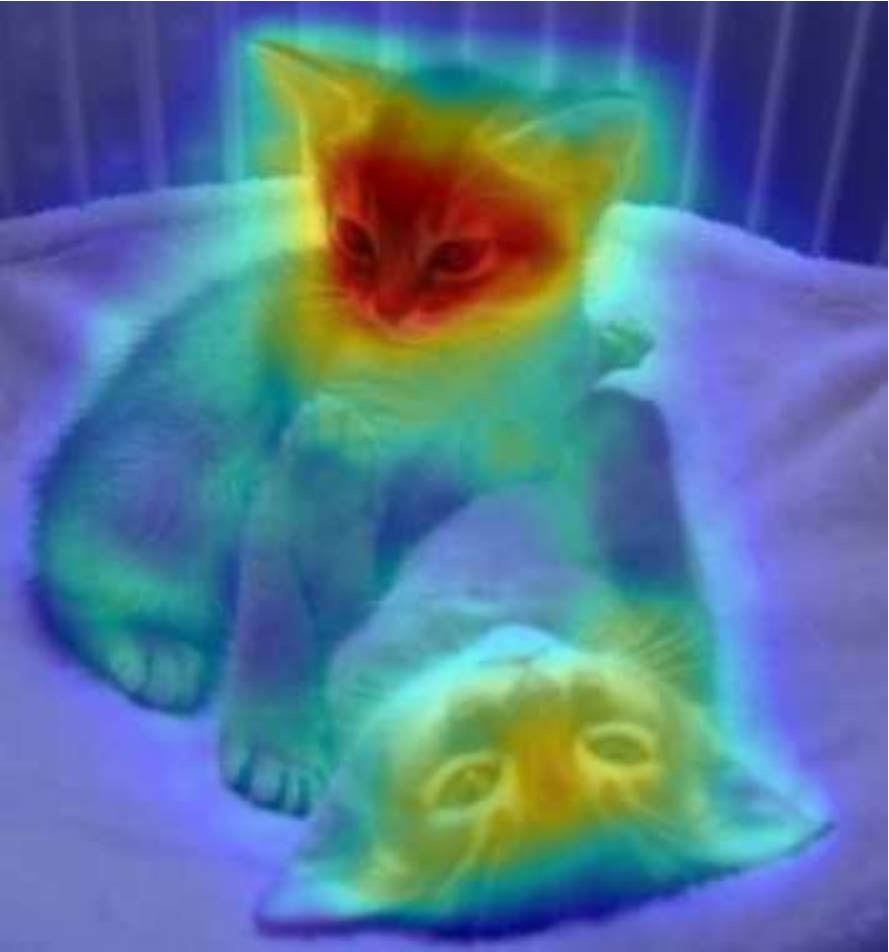}\\
        \vspace{0.02cm}
        \includegraphics[width=0.8in,height=0.56in]{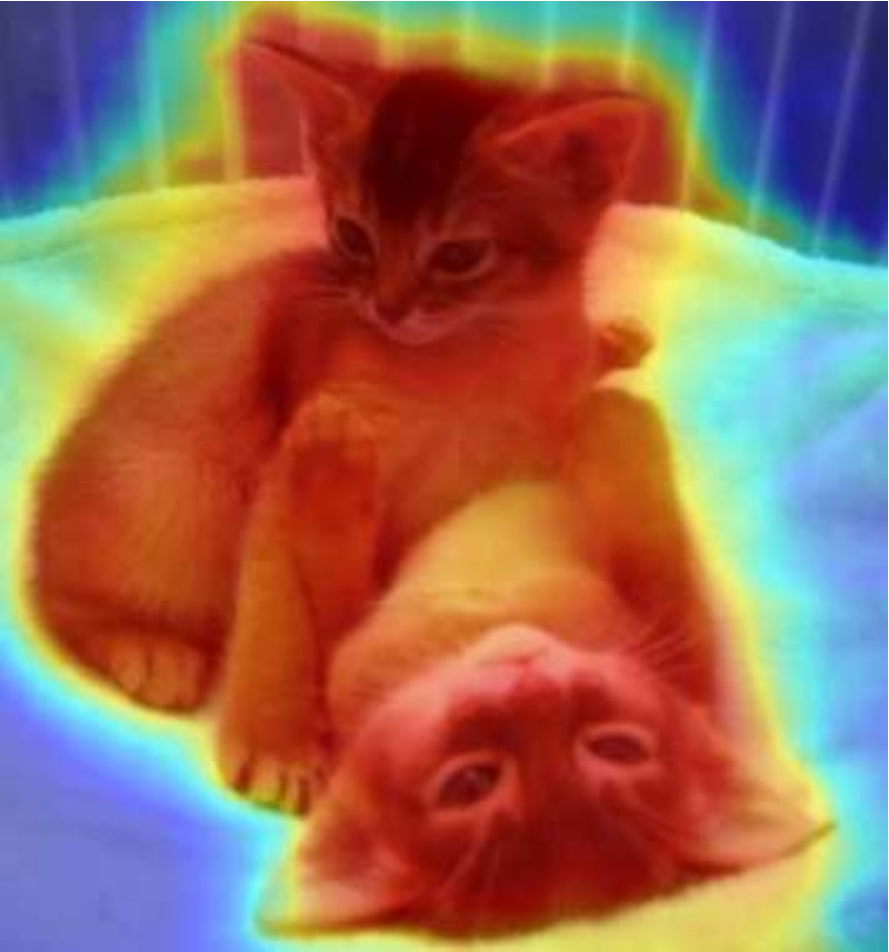}\\
        \vspace{0.02cm}
    \end{minipage}%
}%
\hspace{-2mm}
\subfigure{
    \begin{minipage}[t]{0.123\linewidth}
        \centering
        \includegraphics[width=0.8in,height=0.56in]{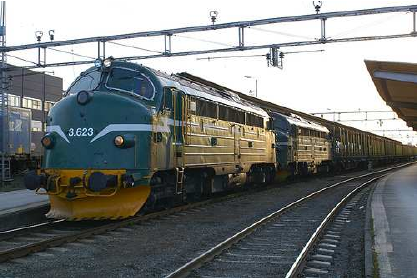}\\
        \vspace{0.02cm}
        \includegraphics[width=0.8in,height=0.56in]{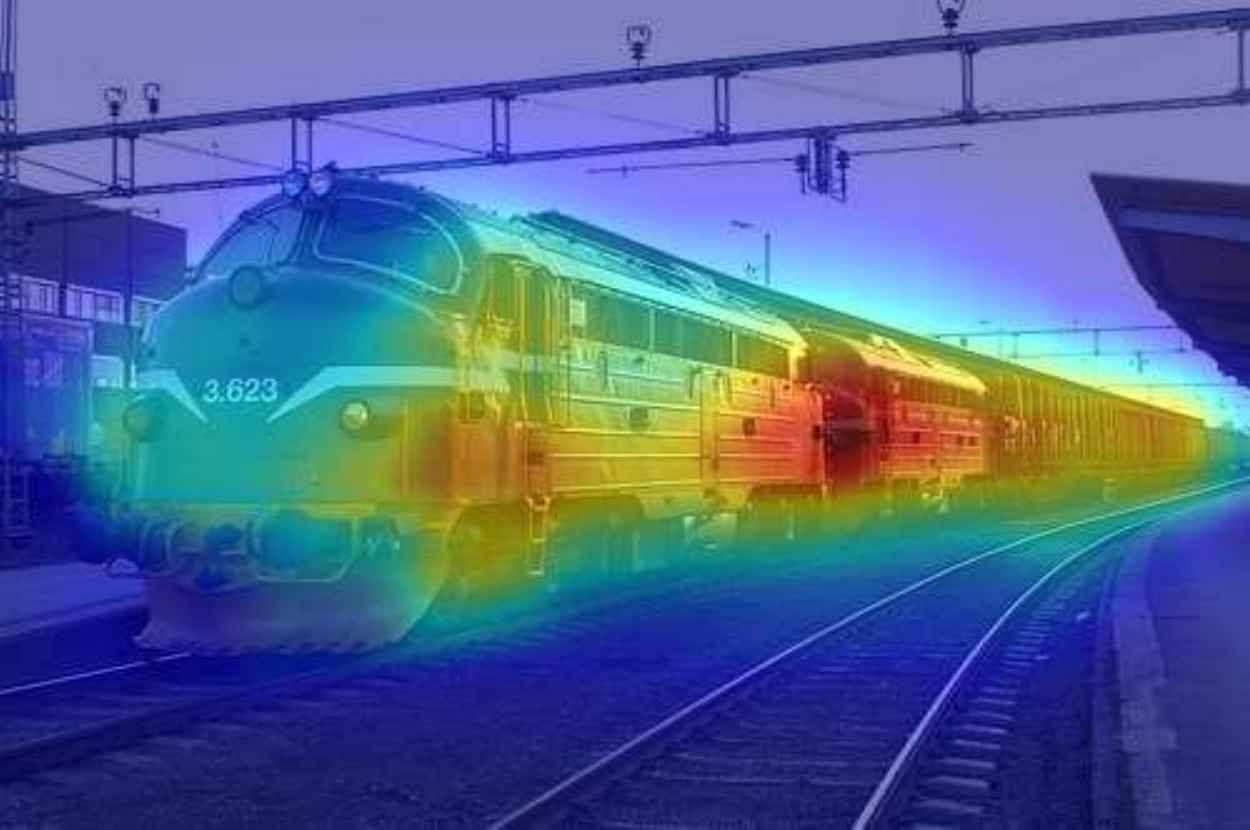}\\
        \vspace{0.02cm}
        \includegraphics[width=0.8in,height=0.56in]{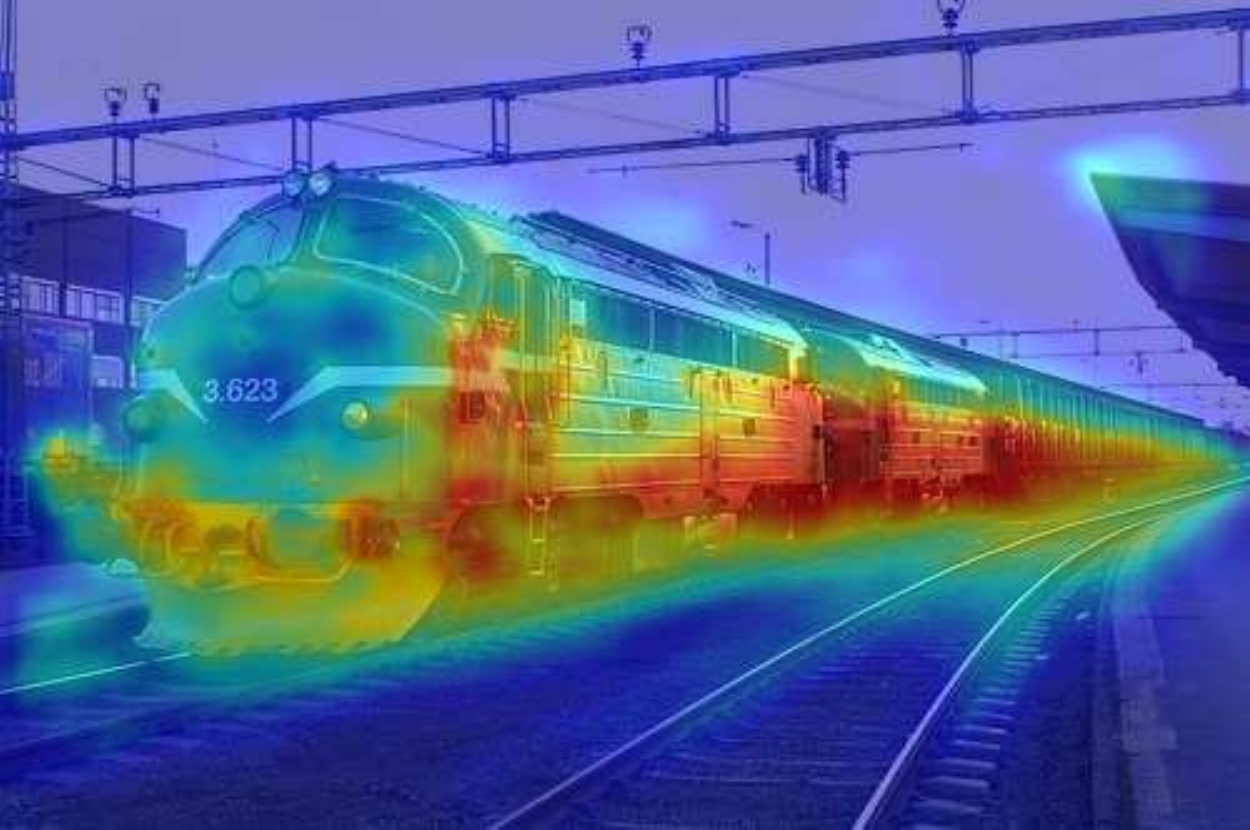}\\
        \vspace{0.02cm}
        \includegraphics[width=0.8in,height=0.56in]{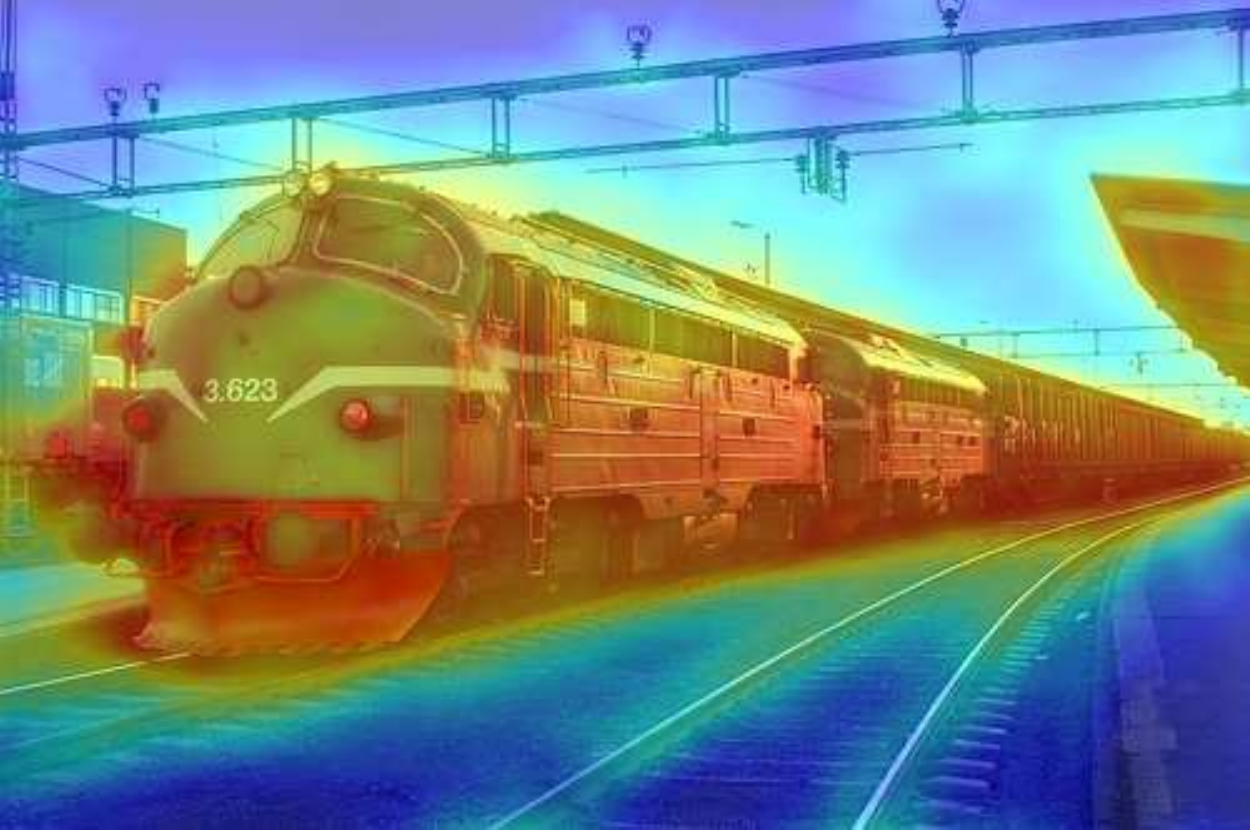}\\
        \vspace{0.02cm}
    \end{minipage}%
}%
\hspace{-2mm}
\subfigure{
    \begin{minipage}[t]{0.123\linewidth}
        \centering
        \includegraphics[width=0.8in,height=0.56in]{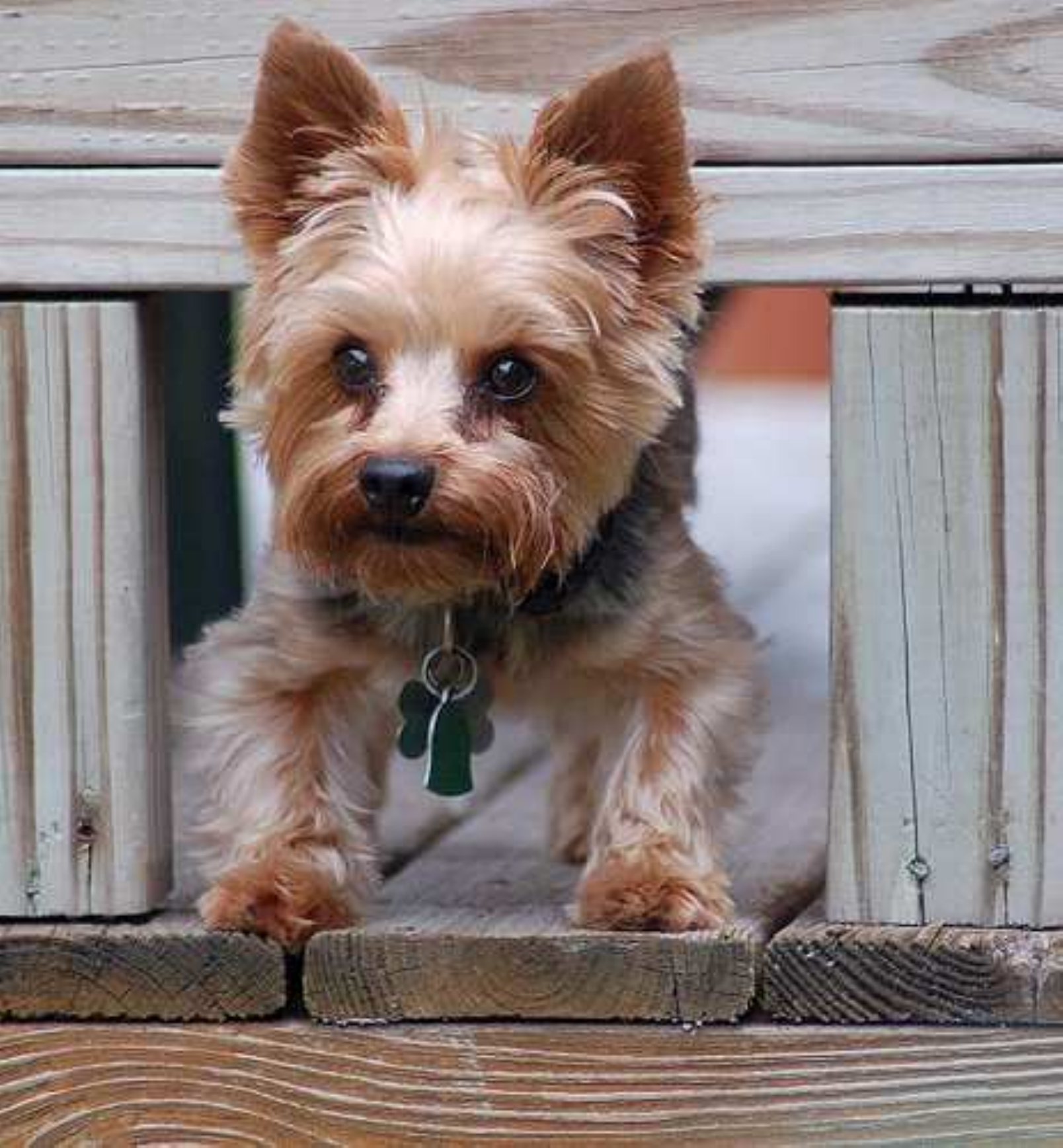}\\
        \vspace{0.02cm}
        \includegraphics[width=0.8in,height=0.56in]{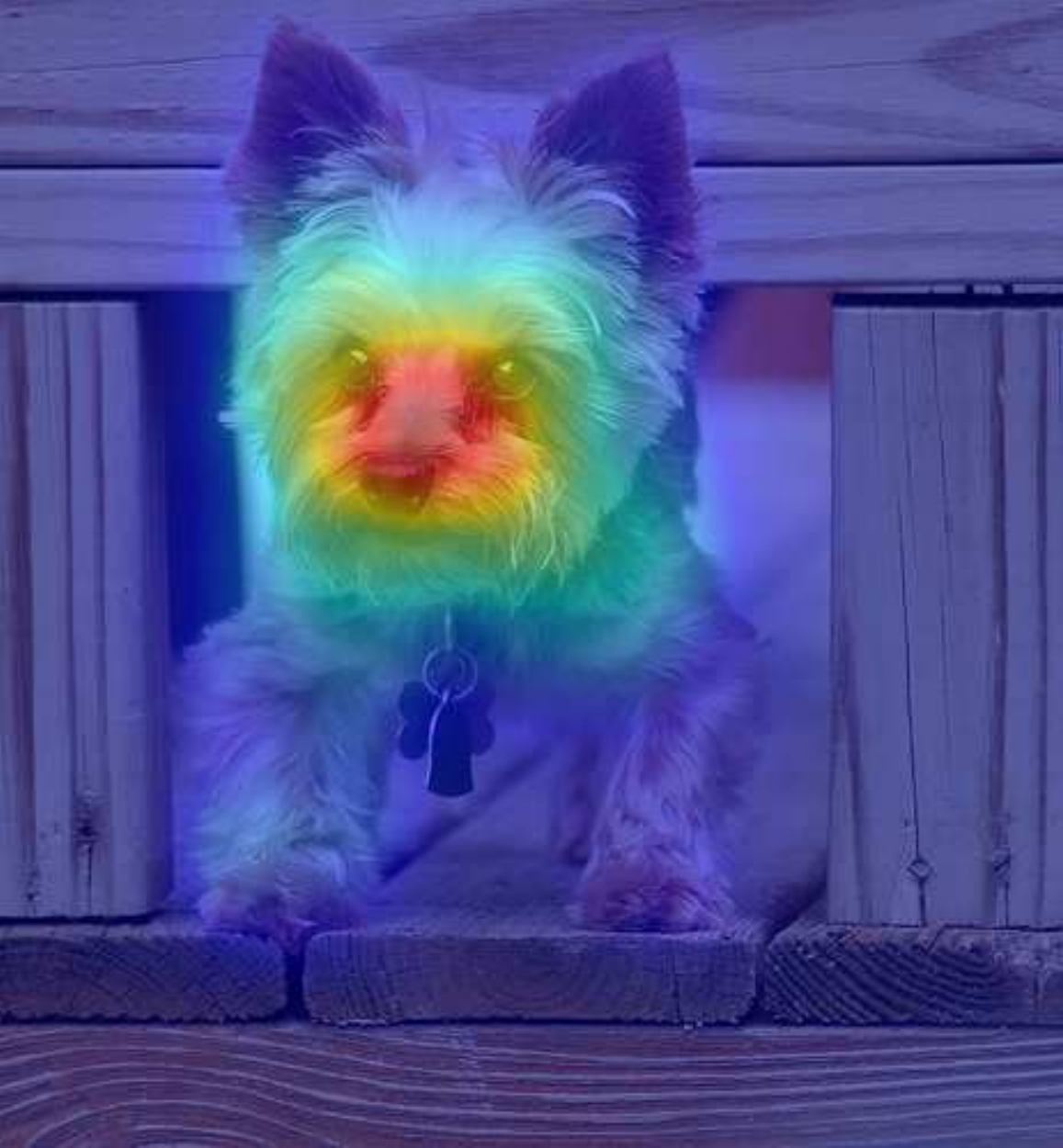}\\
        \vspace{0.02cm}
        \includegraphics[width=0.8in,height=0.56in]{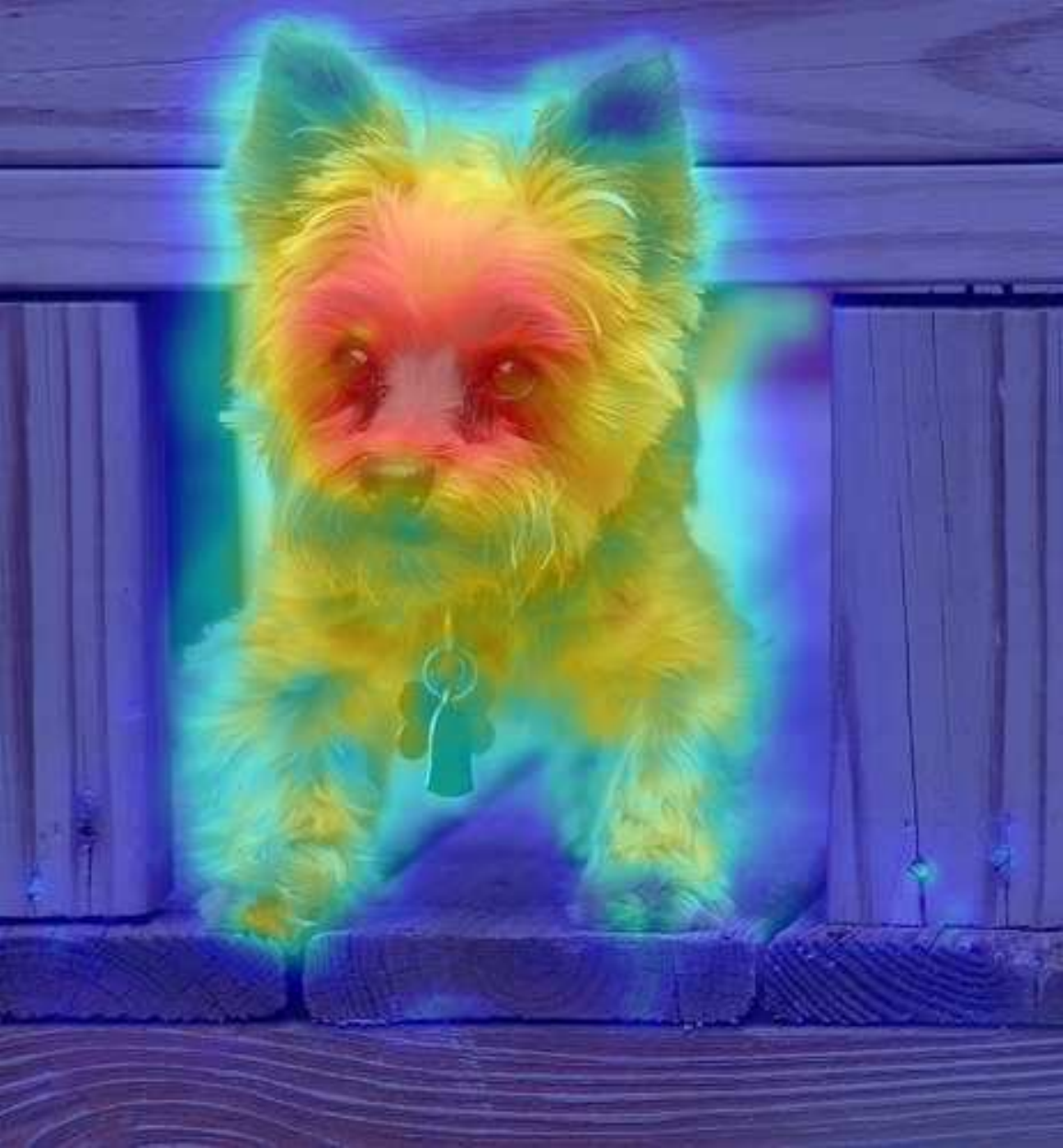}\\
        \vspace{0.02cm}
        \includegraphics[width=0.8in,height=0.56in]{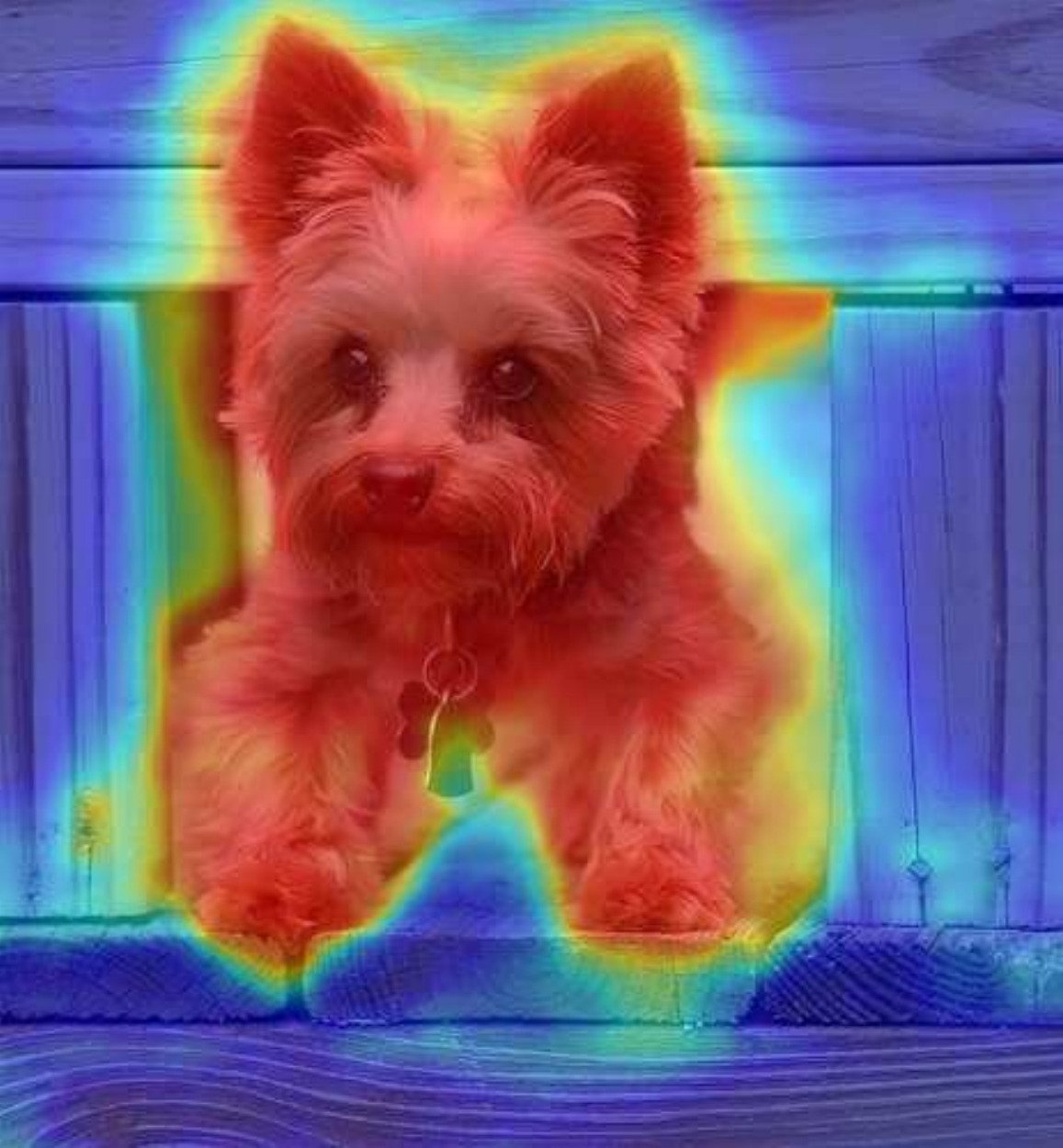}\\
        \vspace{0.02cm}
    \end{minipage}%
}%
\hspace{-2mm}
\subfigure{
    \begin{minipage}[t]{0.123\linewidth}
        \centering
        \includegraphics[width=0.8in,height=0.56in]{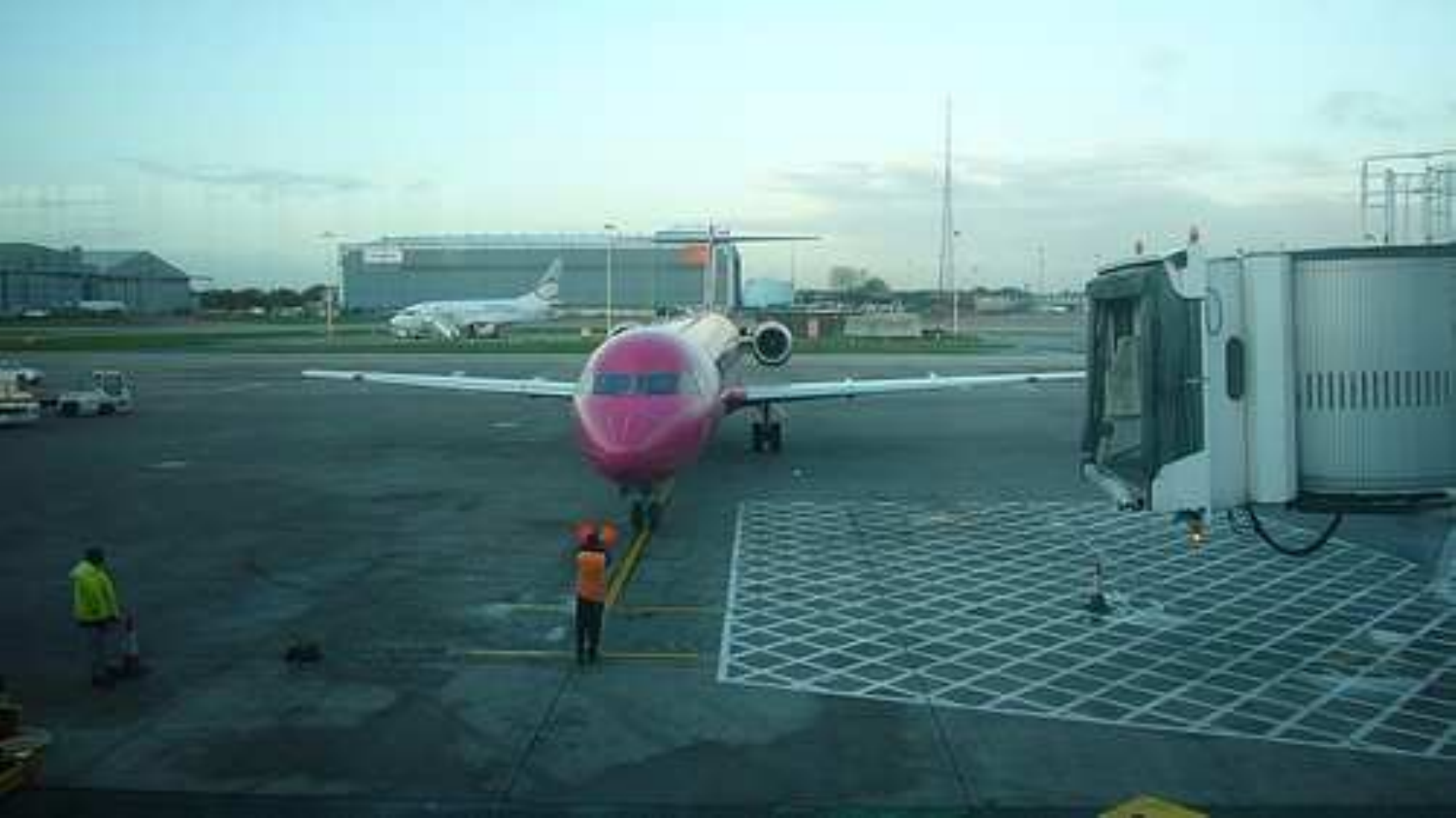}\\
        \vspace{0.02cm}
        \includegraphics[width=0.8in,height=0.56in]{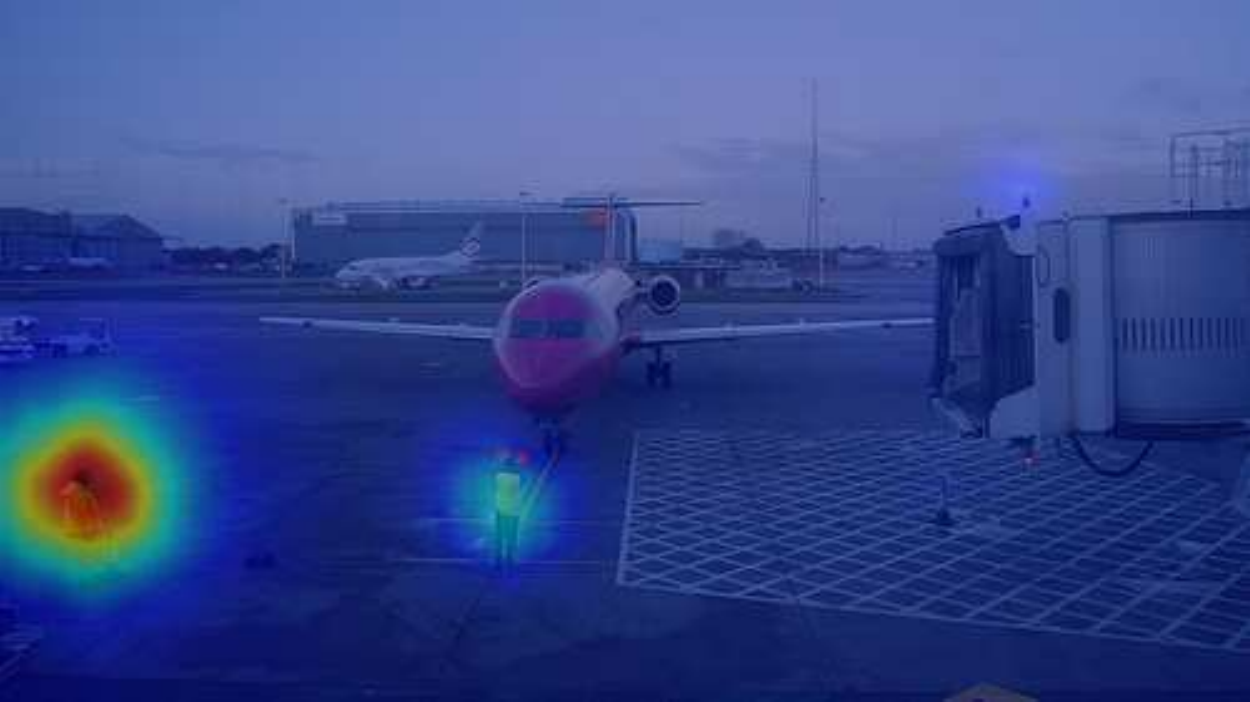}\\
        \vspace{0.02cm}
        \includegraphics[width=0.8in,height=0.56in]{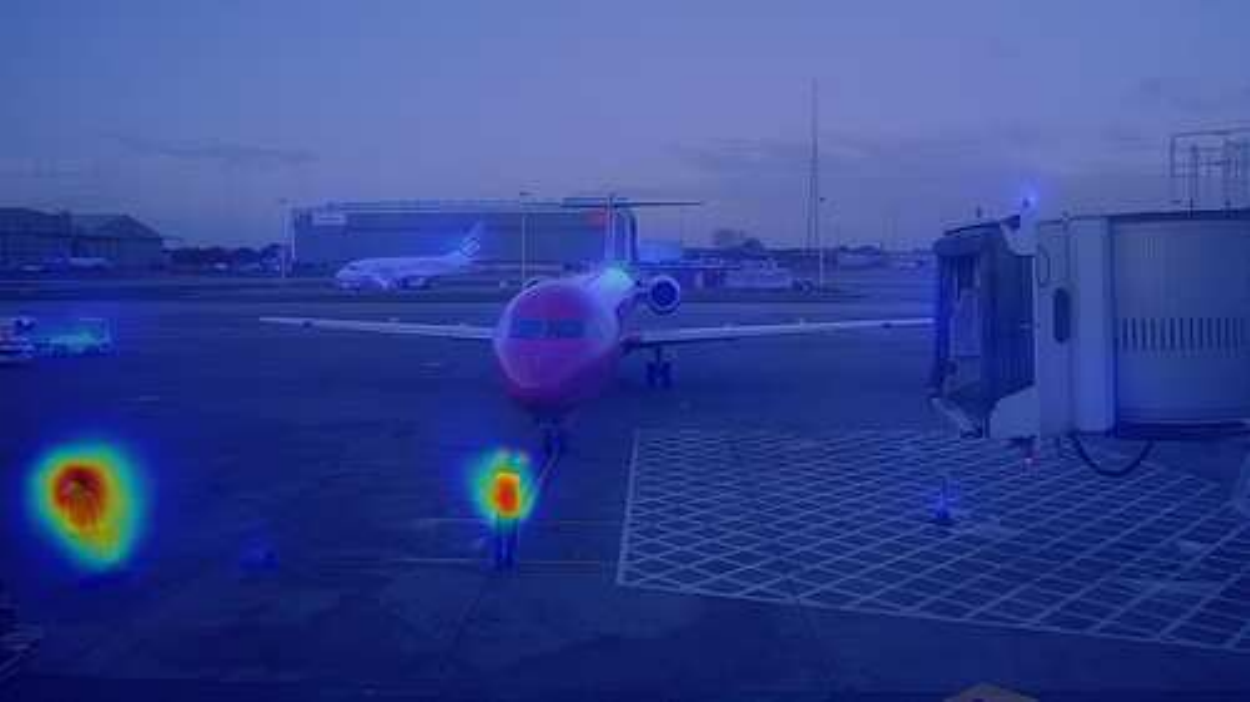}\\
        \vspace{0.02cm}
        \includegraphics[width=0.8in,height=0.56in]{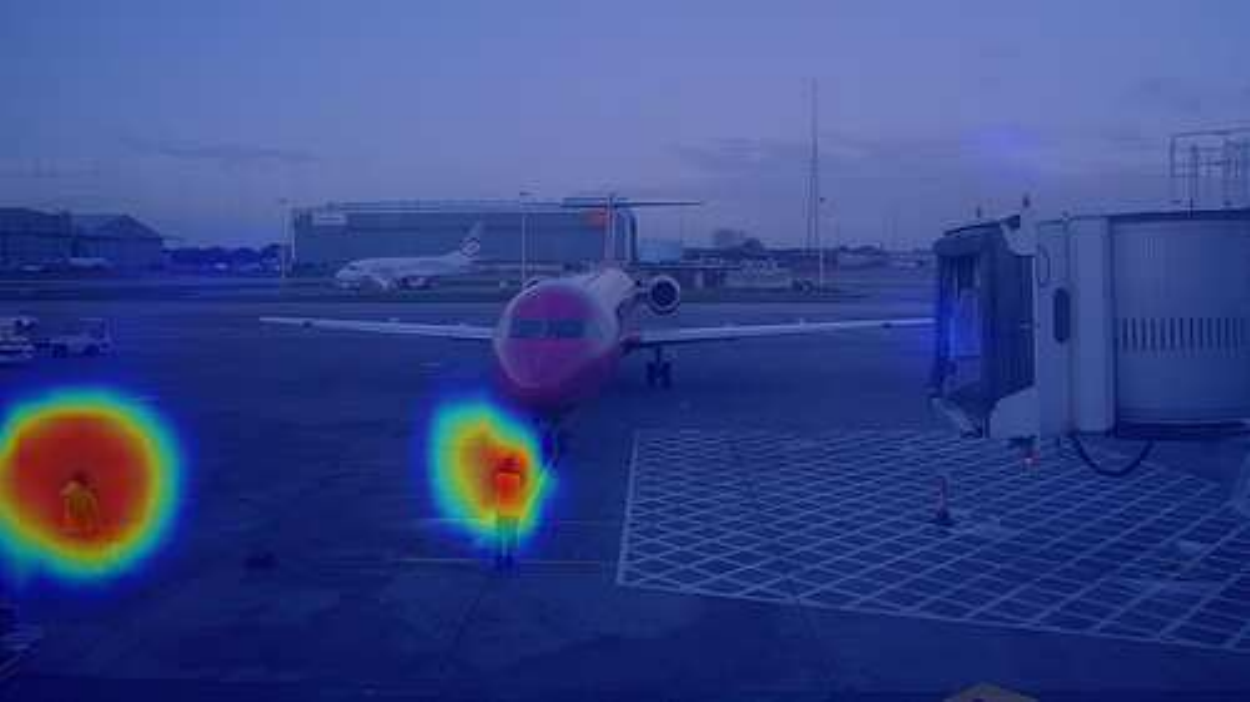}\\
        \vspace{0.02cm}
    \end{minipage}%
}%
\centering
\caption{Sample original images (a) and the corresponding visual results by different methods, which are the baseline (b), SEAM (c) and CPN (d). Our method can find more seeds that both the baseline and the SEAM are unable to dig out.}
\label{fig:compare_fig}
\end{figure*}

Fig.~\ref{fig:compare_fig} shows several samples of the visualized CAM made by the baseline, SEAM and CPN. Compared to the baseline and SEAM, our CPN can help CAM seek more seeds in low response areas to generate a complete CAM for the foreground.  However, for small objects (the last column in Fig. \ref{fig:compare_fig}), it can be seen that the foreground seeds by CPN are over-segmented since it is indeed difficult to mine out accurate seeds for small objects without boundary.

\noindent\textbf{Effectiveness of regularization and PRCM:}\; Tab. \ref{tab:ablation on cpn} illustrates the effect of every single module in our approach. Note that for the baseline method only $\boldsymbol{l}_{cls}(s_o)$ is included in $\mathcal {L}_{cls}$ since it lacks the CP Pair. Compared with the baseline, the $\mathcal {L}_{tcp}$ and PCM improves the mIoU up to 51.08\%. Benefit by $\mathcal {L}_{cpcr}$, the model further achieves a 4.63\% improvement. By further applying OHEM to the $\mathcal {L}_{cpcr}$, the results achieves 56.58\% mIoU on the VOC12 \textit{train} set. Finally, the model achieves a 0.85\% improvement after adopting the PRCM.

\begin{table}[!htbp]\small
\centering\
\begin{tabular}{c|c|c|c|c|c}
\hline
\multirow{2}{*}{Method} &
\multicolumn{5}{|c}{Image scale}\\
\cline{2-6}
  & 0.5 & 1.0 & 1.5 & 2.0 & all\\
\hline
baseline &41.15&48.29&49.51&47.44&47.84\\
SEAM~\cite{seam} & 49.35 & 51.57 & 52.25 & 49.79 & 55.41\\
SEAM* & \textbf{49.64}& 52.15& 53.14 & 50.55 & 55.71\\
CPN &48.91& \textbf{54.51} &\textbf{ 55.44} & \textbf{54.01} & \textbf{57.43}\\
\hline
\end{tabular}
\caption{Experiments with single- and multi-scale tests on several methods. * marks the method uses our PRCM module. It is shown that the CPN boosts the overall performance of CAM in various scales, and achieves better CAM than SEAM and the baseline. In addition, our PRCM is effective for improving the results on different scales.}
\label{tab:scale}
\end{table}

\begin{table}[!htbp]\small
\centering\
\begin{tabular}{lc}
\hline
model& mIoU (\%)\\
\hline
baseline ($ \mathcal {L}_{cls}$)   &     47.84\\
\hline
baseline + $ \mathcal {L}_{tcp}$ + PCM & 51.08\\
baseline+$ \mathcal {L}_{tcp}$+ $\mathcal {L}_{cpcr}$+PCM & 55.71\\
baseline+$ \mathcal {L}_{tcp}$+ $\mathcal {L^{*}}_{cpcr}$+PCM & 56.58\\
baseline+$ \mathcal {L}_{tcp}$+ $\mathcal {L^{*}}_{cpcr}$ + PCM+PRCM & 57.43\\
\hline
\end{tabular}
\caption{Ablation studies on every part of our method. $\mathcal {L^{*}}_{cpcr}$ refers to $\mathcal {L}_{cpcr}$ with Online Hard Example Mining (OHEM) }
\label{tab:ablation on cpn}
\end{table}

\noindent\textbf{Improvements on foreground localization:}\; The CPN aims to improve the CAM by capturing more seeds related to the foreground objects. To verify the idea, we collect the mIoU of background and 20 foreground objects from the baseline, SEAM, and our CPN in Tab. \ref{tab:scale}. As shown in Tab. \ref{tab:ablation on fb}, compared to the baseline, SEAM achieves compelling mIoU elevation by 5.14\% in the background and 7.68\% in the foreground. In addition, the CPN improves the mIoU of the foreground up to 56.13\%, which outperforms the baseline by 9.75\% and the SEAM by 2.07\%. The result validates that our CPN can find more foreground object regions than the baseline and the SEAM.
\begin{table}[!htbp]\small
\centering\
\begin{tabular}{c|c|c|c}
\hline
Method & baseline & SEAM & CPN \\
\hline
 \textit{bg.} & 77.19 & 82.33 & \textbf{83.44} \\
 \textit{f.} & 46.38 & 54.06 & \textbf{56.13} \\
\hline
\end{tabular}
\caption{The mIoU (\%) of the foreground objects (\textit{f.}) and background (\textit{bg.}) in different methods. }
\label{tab:ablation on fb}
\vspace{-2mm}
\end{table}

\noindent\textbf{Patch size:}\; Recall from that two patch strategies, namely Grid Patch and Super-pixel Patch, can both be applied on our CPN (Sec. \ref{sec_strategy}). Note that both patch strategies are closely related to the patch size, and the total number of patches increases with the reduction of it.

\begin{figure}
\begin{center}
\includegraphics[height = 1.2in,width=3.3in]{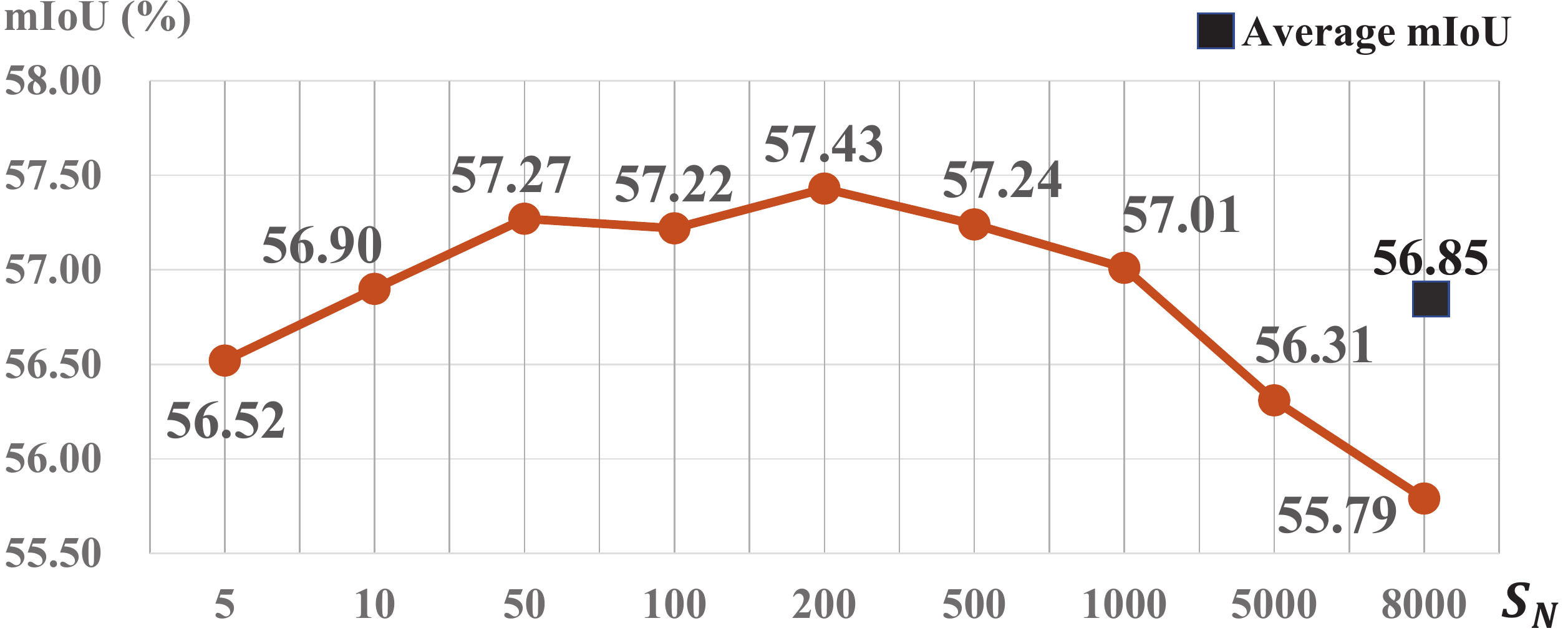}
\end{center}
\vspace{-4mm}
   \caption{The performance of the Super-pixel Patch strategy with different $S_N$.}
\label{tab:ablation on ss}
\end{figure}

For Super-pixel Patch, we explore the effect by simply changing $S_N$, ranging from 5 to 8000. Fig. \ref{tab:ablation on ss} reports the mIoU of the results by the CPN with different $S_N$. It shows that the mIoU firstly shows a rough increasing trend with the increase of $S_N$, and reaches the peak point (\textbf{57.43\%}) with $S_N=$ 200. Then the quality of CAM declines with higher $S_N$, and arrives at the lowest point (55.79\%) with $S_N=8000$. Note that lower $S_N$ causes the larger patch size, thus both too high and low patch size in Super-pixel Patch suppress the improvements on the CAM.

\begin{figure}
\begin{center}
\includegraphics[height = 1.2in,width=3.3in]{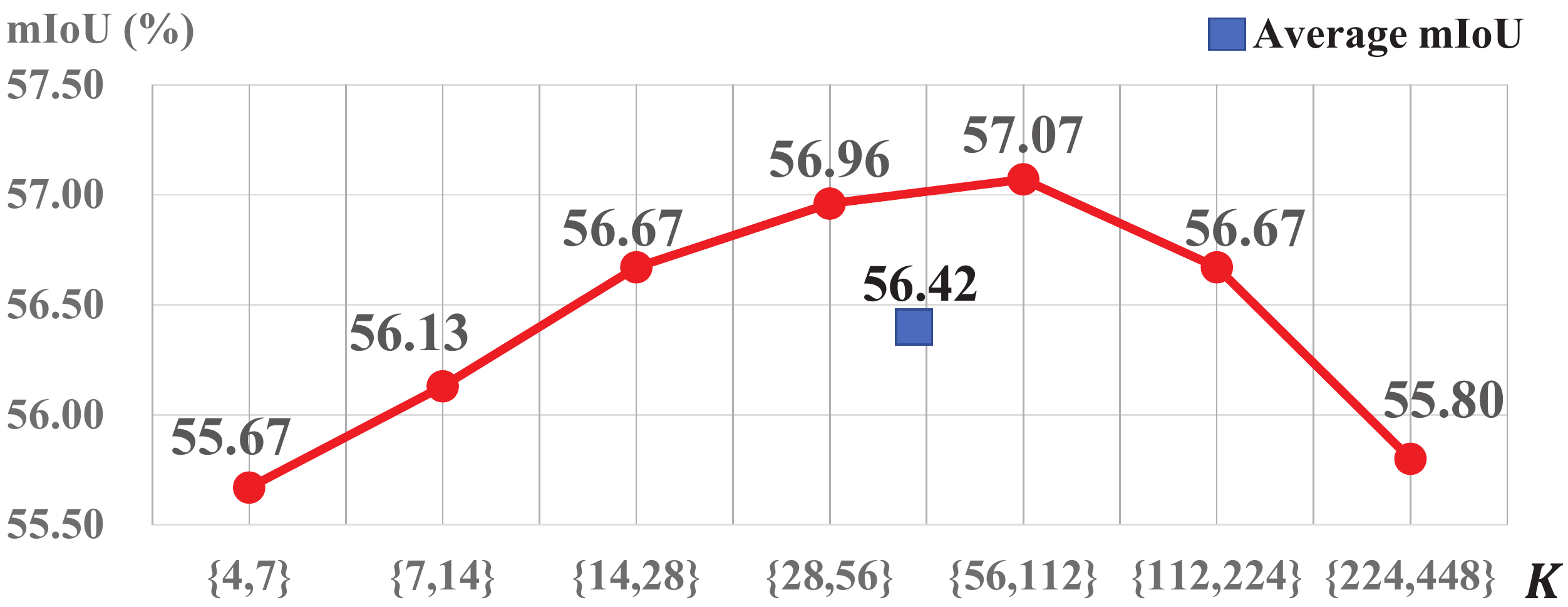}
\end{center}
   \caption{The performance of the Grid Patch strategy with various patch sets $K$.}
\label{tab:ablation on sp}
\vspace{-2mm}
\end{figure}

For Grid Patch, patch size $S$ is evenly chosen from the set $K$ of some fixed numbers. Thus we implement the experiment by changing the elements of $K$. We keep $|K| = 2$ and gradually increase the smaller element. Note that the input size of the image is 448. Fig. \ref{tab:ablation on sp} summaries the results. We notice that with the increase of the $S$, the mIoU follows a similar tendency to the one in the Super-pixel Patch. The result achieves the best performance (\textbf{57.07\%}) with $K = \{56,112\}$, and reaches the bottom with $K = \{4,7\}$ and $K = \{224,448\}$, respectively achieving 55.67\% and 55.80\% mIoU.

Recall that two extreme conditions hold the equality of the CP representation (Sec. \ref{CPR}). The results in large and small patch size are exactly corresponding to the condition 1) and 2). Therefore, it is concluded that proper hidden patch size is crucial for significantly improving the performance of the CAM.

\noindent\textbf{Grid Patch vs Super-pixel Patch:}\;To contrast the performance between the two strategies, we respectively calculate the average mIoU of the results in Fig. \ref{tab:ablation on sp} and Fig. \ref{tab:ablation on ss}. It is observed that CPN with the Super-pixel Patch strategy globally achieves better results (+\textbf{0.43\%}) than the Grid Patch strategy. It is reasonable since the former one benefits from the pre-classification by the super-pixel. Besides, we test the average time consumption. The Super-pixel Patch (1.45 sec/per image) apparently consumes much more time than the Grid Patch (\textbf{0.006 sec/per image}), so the latter one could better meet some real-time operations.

\begin{figure*}[htbp]
\centering
\begin{minipage}[t]{0.03\linewidth}
        \centering
        {}
        \vspace{-0.8cm}
        {(a)}
        \\
        \vspace{1.1cm}
        {(b)}
        \\
        \vspace{1.0cm}
        {(c)}
        \end{minipage}%
\subfigure{
    \begin{minipage}[t]{0.125\linewidth}
        \centering
        \includegraphics[width=0.8in,height=0.551in]{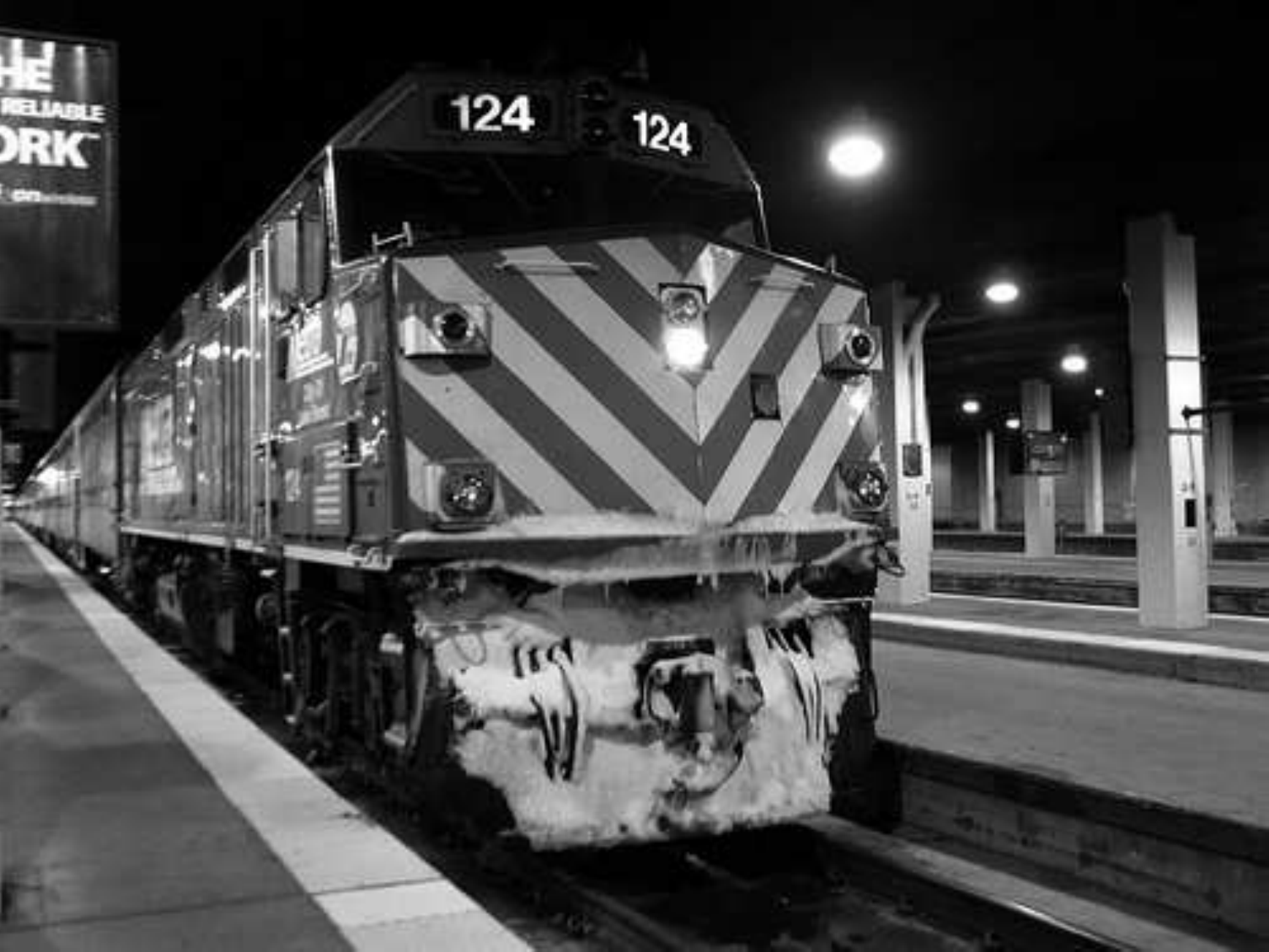}\\
        \vspace{0.02cm}
        \includegraphics[width=0.8in,height=0.551in]{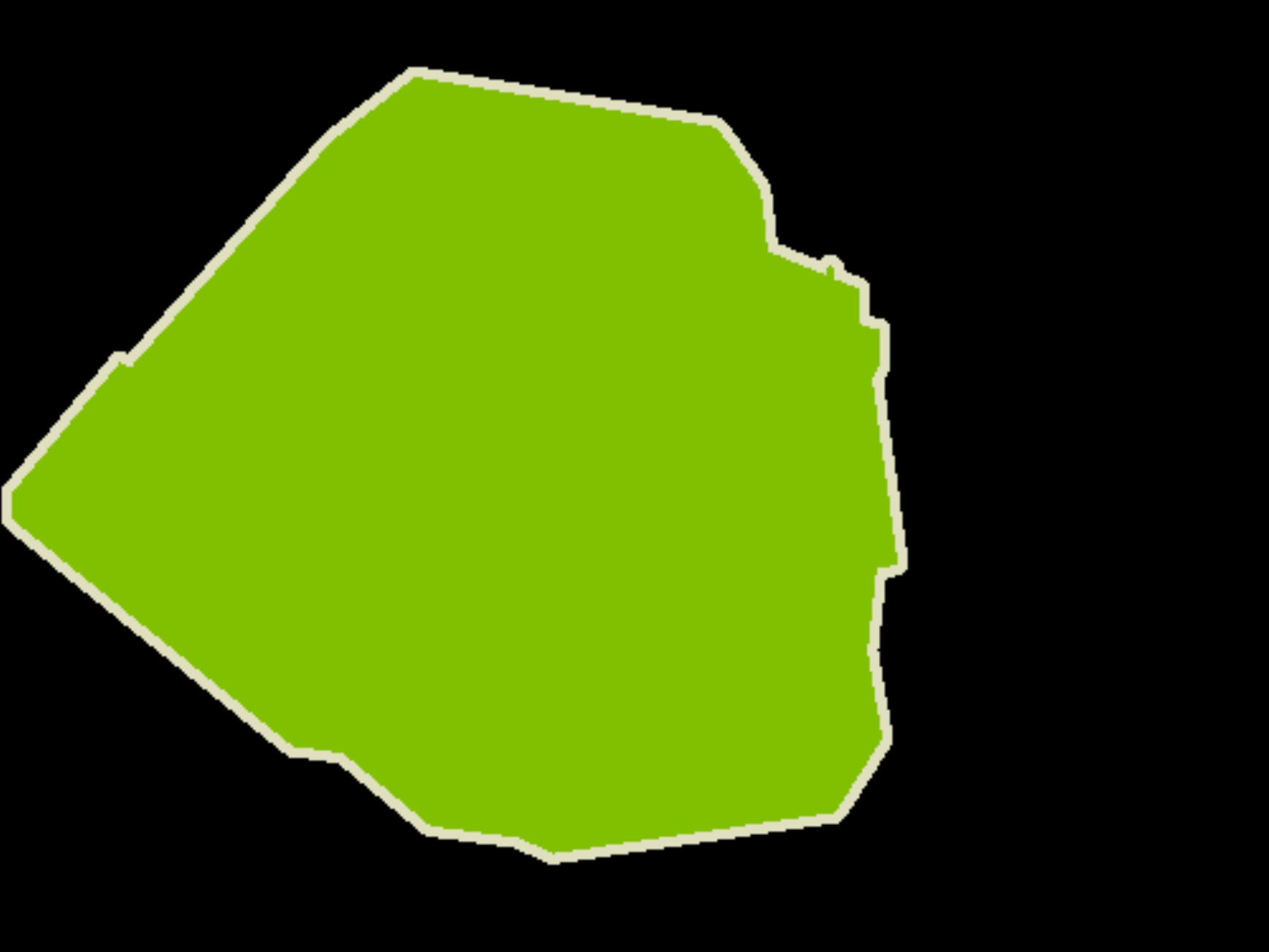}\\
        \vspace{0.02cm}
        \includegraphics[width=0.8in,height=0.551in]{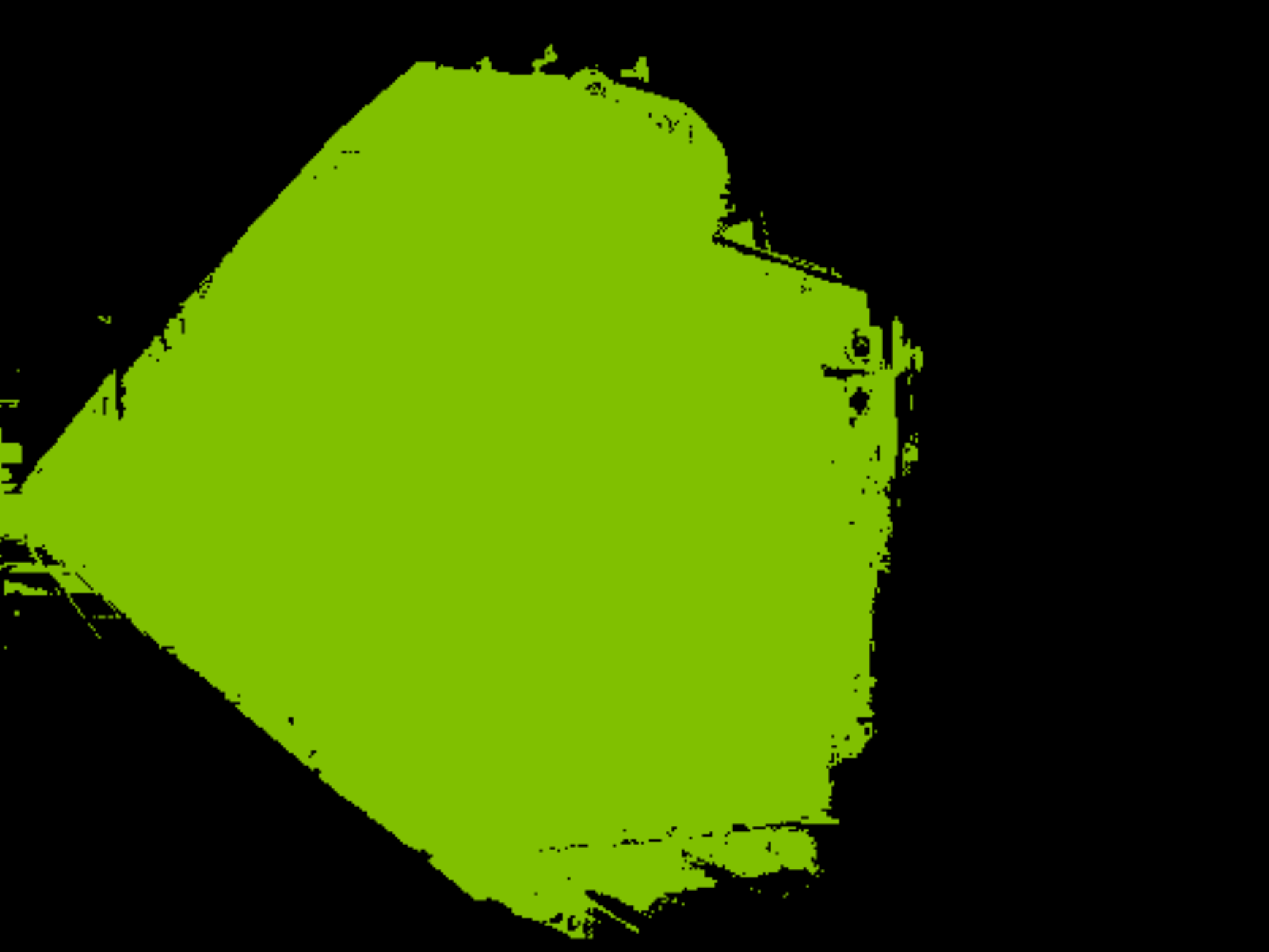}\\
        \vspace{0.02cm}
    \end{minipage}%
}
\hspace{-4.8mm}
\subfigure{
    \begin{minipage}[t]{0.100\linewidth}
        \centering
        \includegraphics[width=0.55in,height=0.551in]{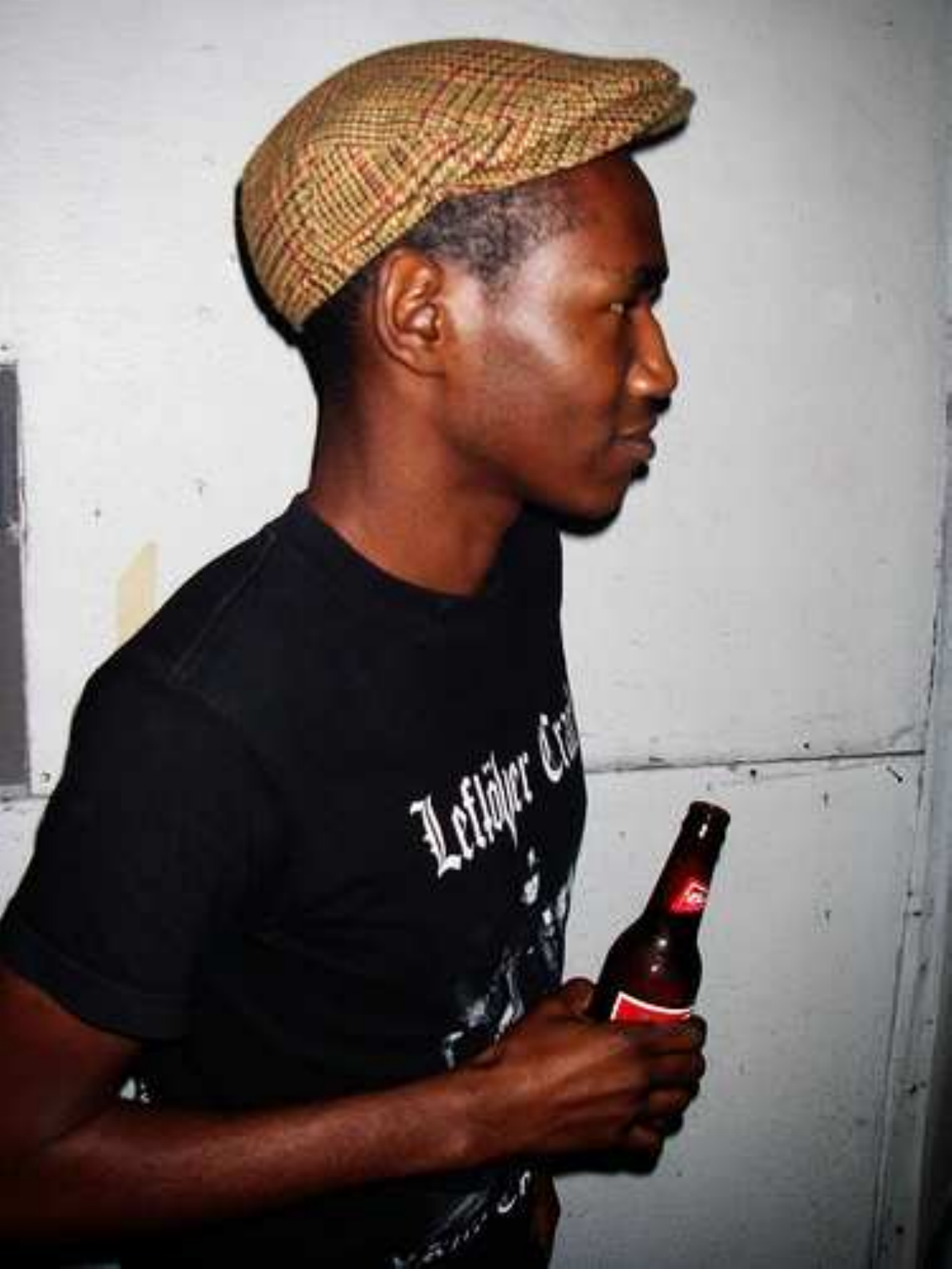}\\
        \vspace{0.02cm}
        \includegraphics[width=0.55in,height=0.551in]{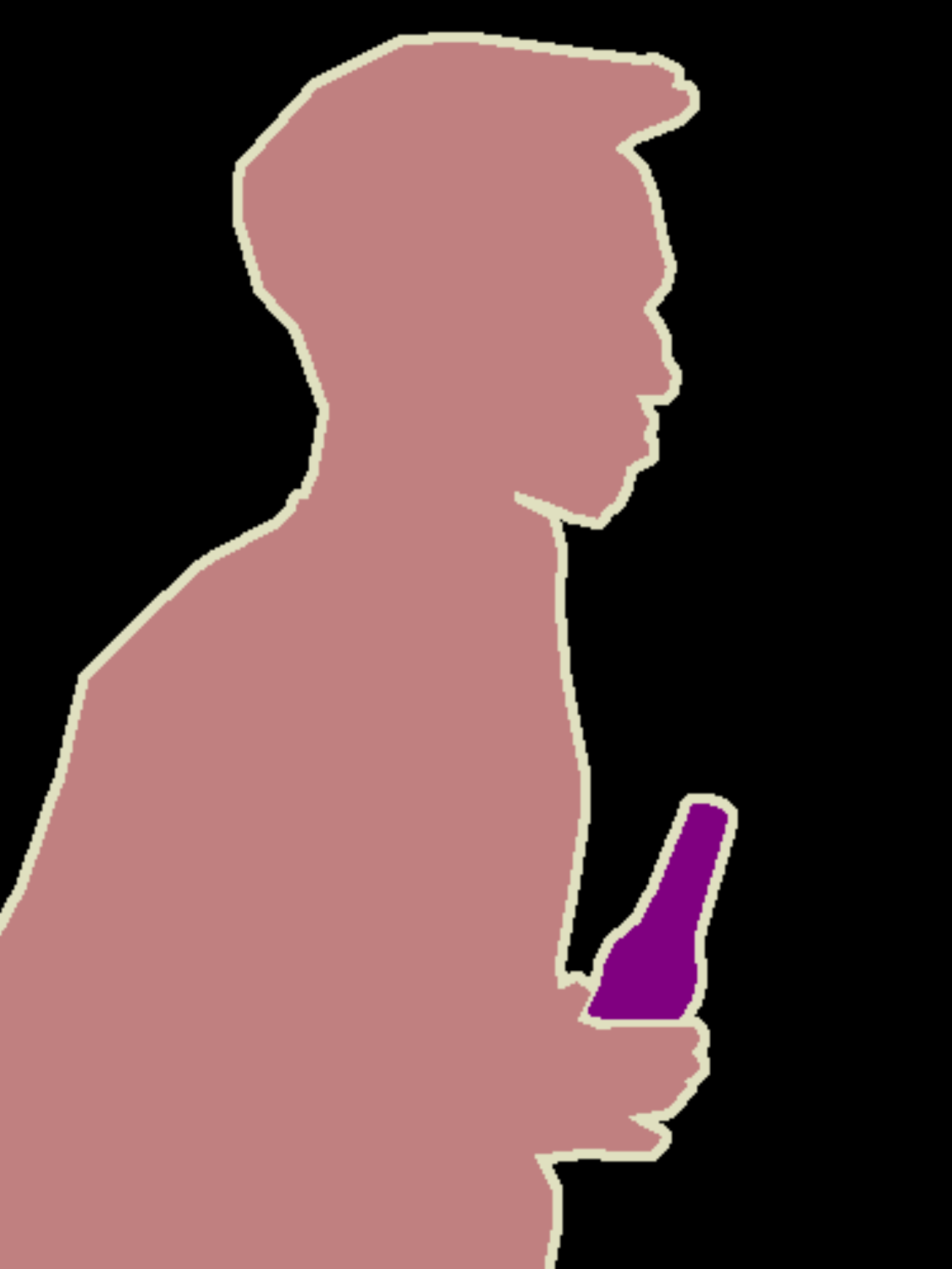}\\
        \vspace{0.02cm}
        \includegraphics[width=0.55in,height=0.551in]{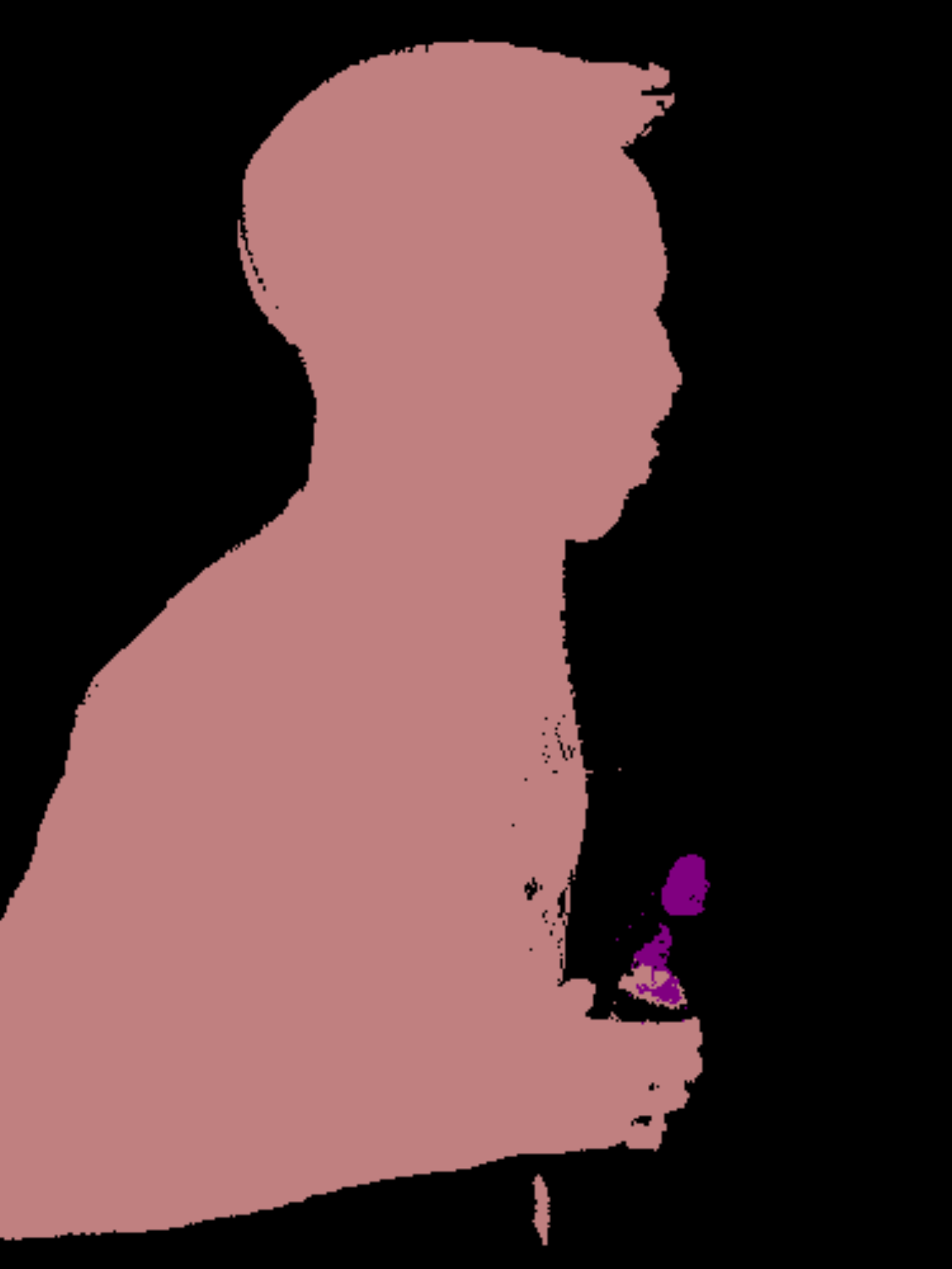}\\
        \vspace{0.02cm}
    \end{minipage}%
}%
\hspace{-4mm}
\subfigure{
    \begin{minipage}[t]{0.125\linewidth}
        \centering
        \includegraphics[width=0.8in,height=0.551in]{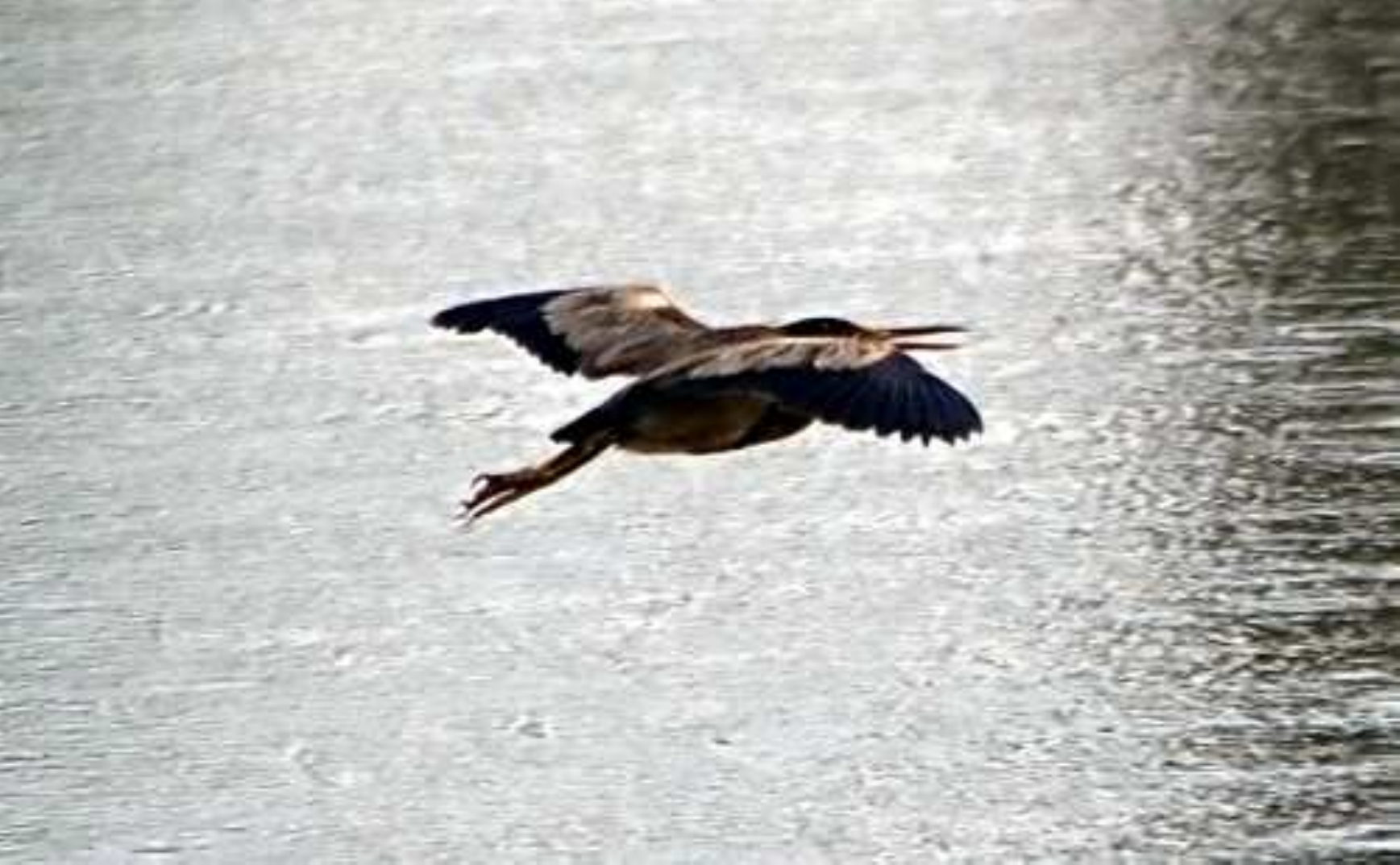}\\
        \vspace{0.02cm}
        \includegraphics[width=0.8in,height=0.551in]{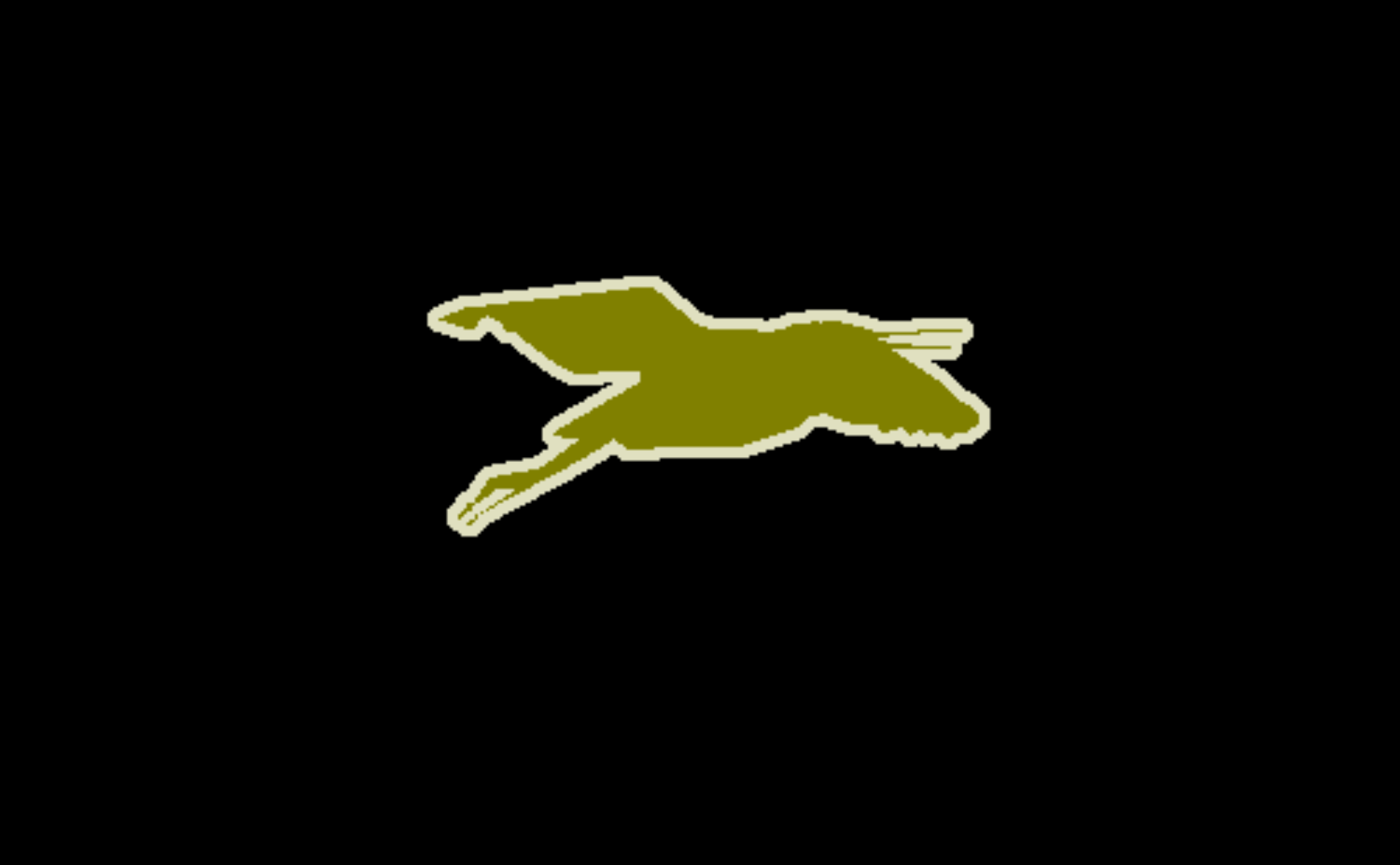}\\
        \vspace{0.02cm}
        \includegraphics[width=0.8in,height=0.551in]{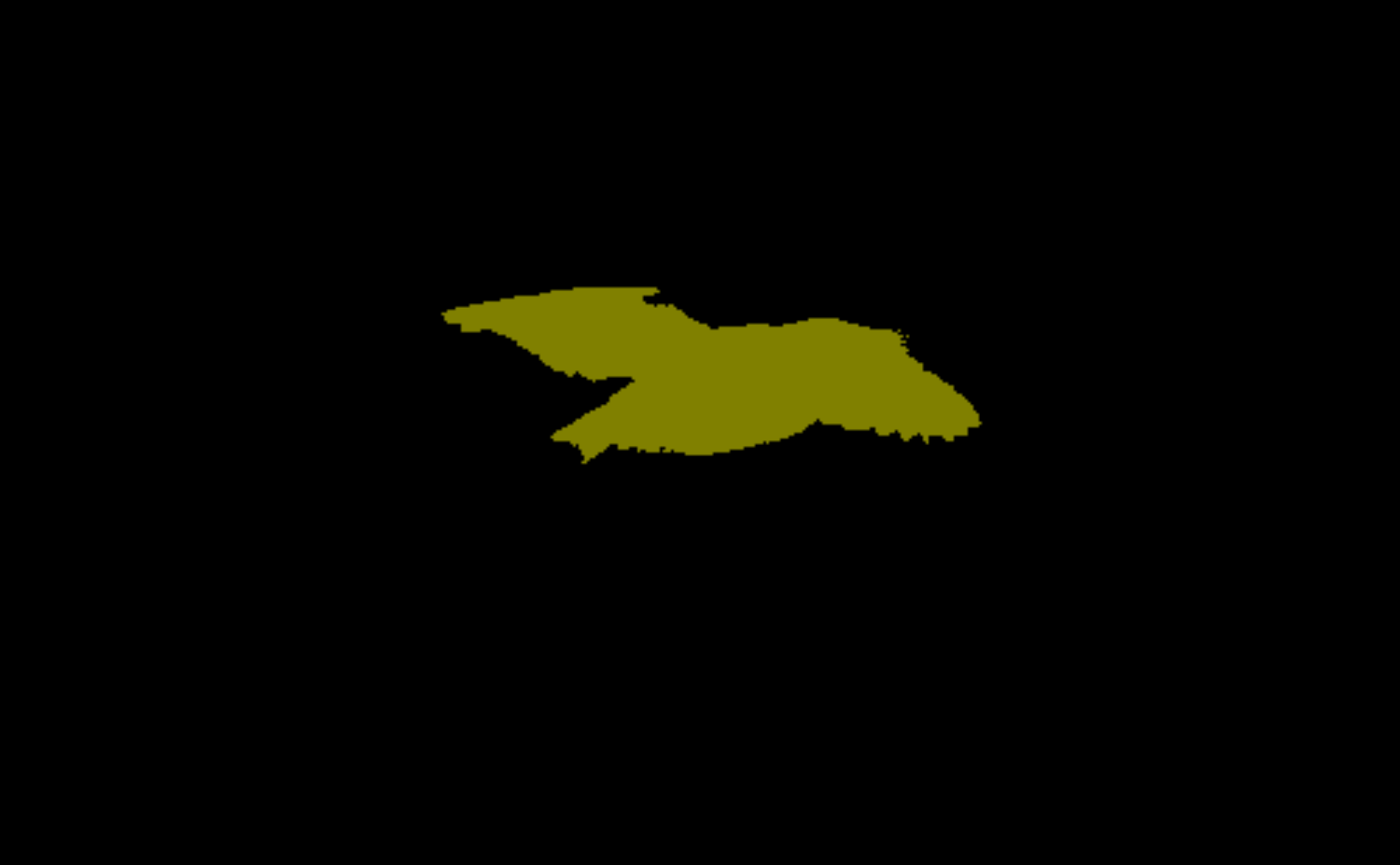}\\
        \vspace{0.02cm}
    \end{minipage}%
}%
\hspace{-3mm}
\subfigure{
    \begin{minipage}[t]{0.125\linewidth}
        \centering
        \includegraphics[width=0.8in,height=0.551in]{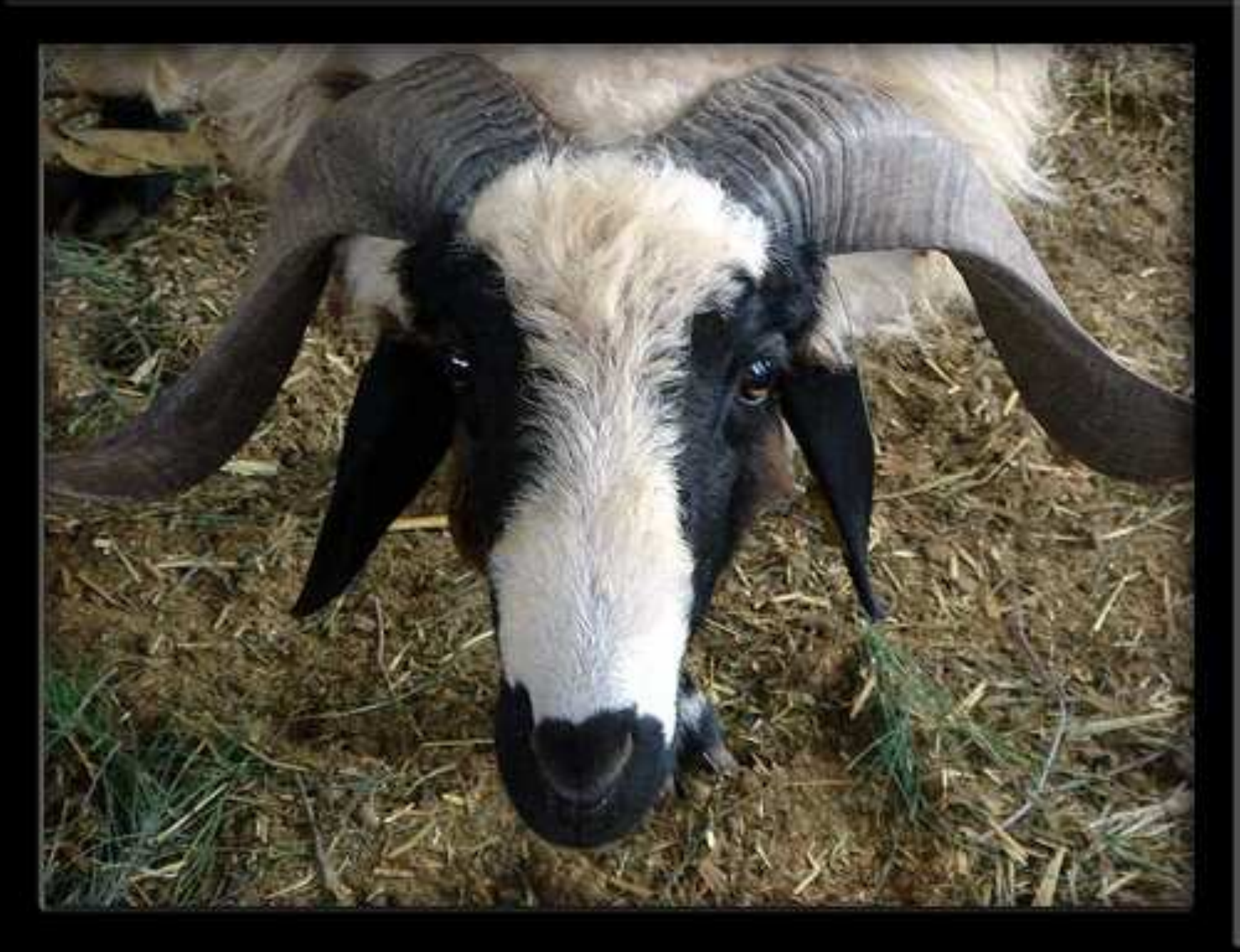}\\
        \vspace{0.02cm}
        \includegraphics[width=0.8in,height=0.551in]{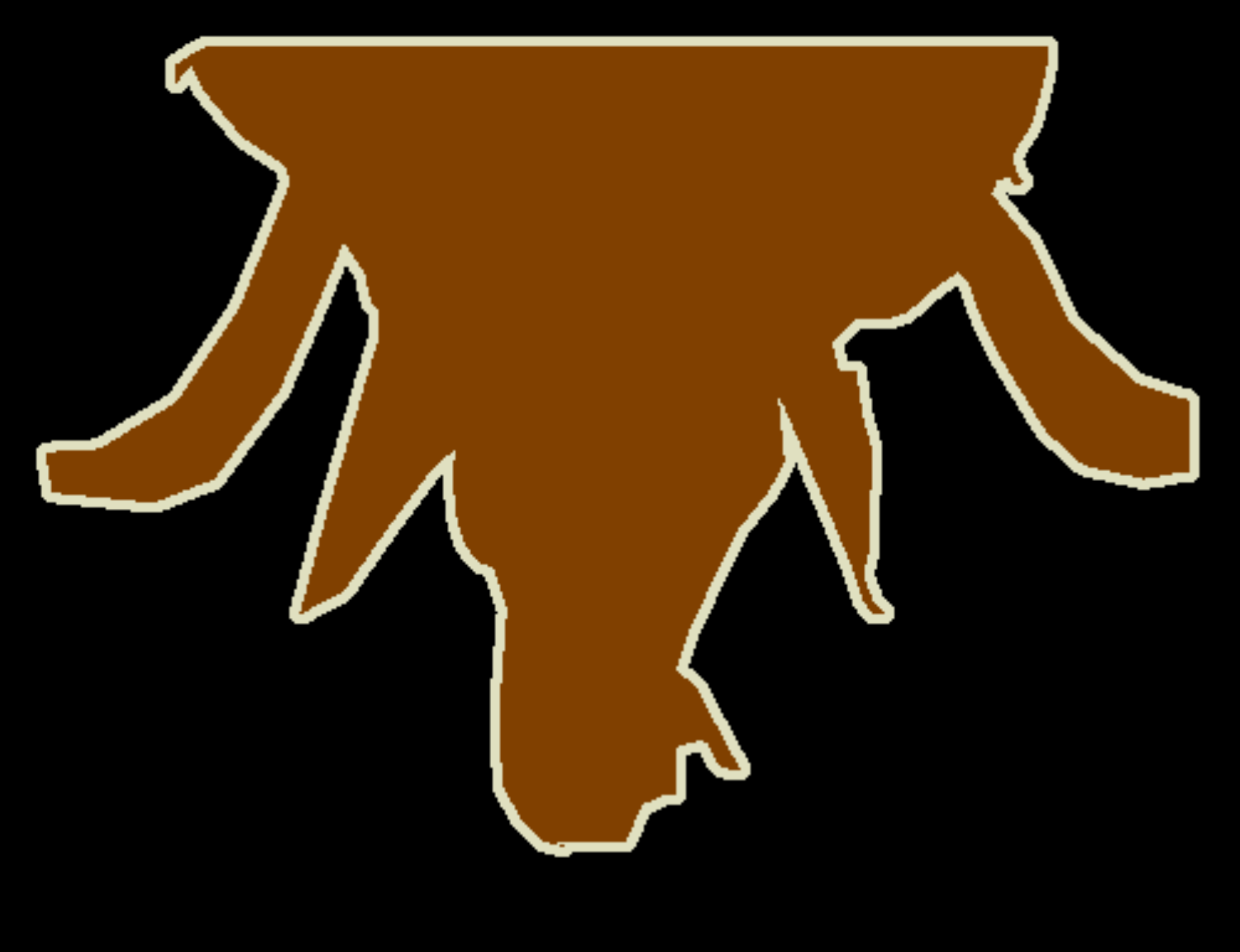}\\
        \vspace{0.02cm}
        \includegraphics[width=0.8in,height=0.551in]{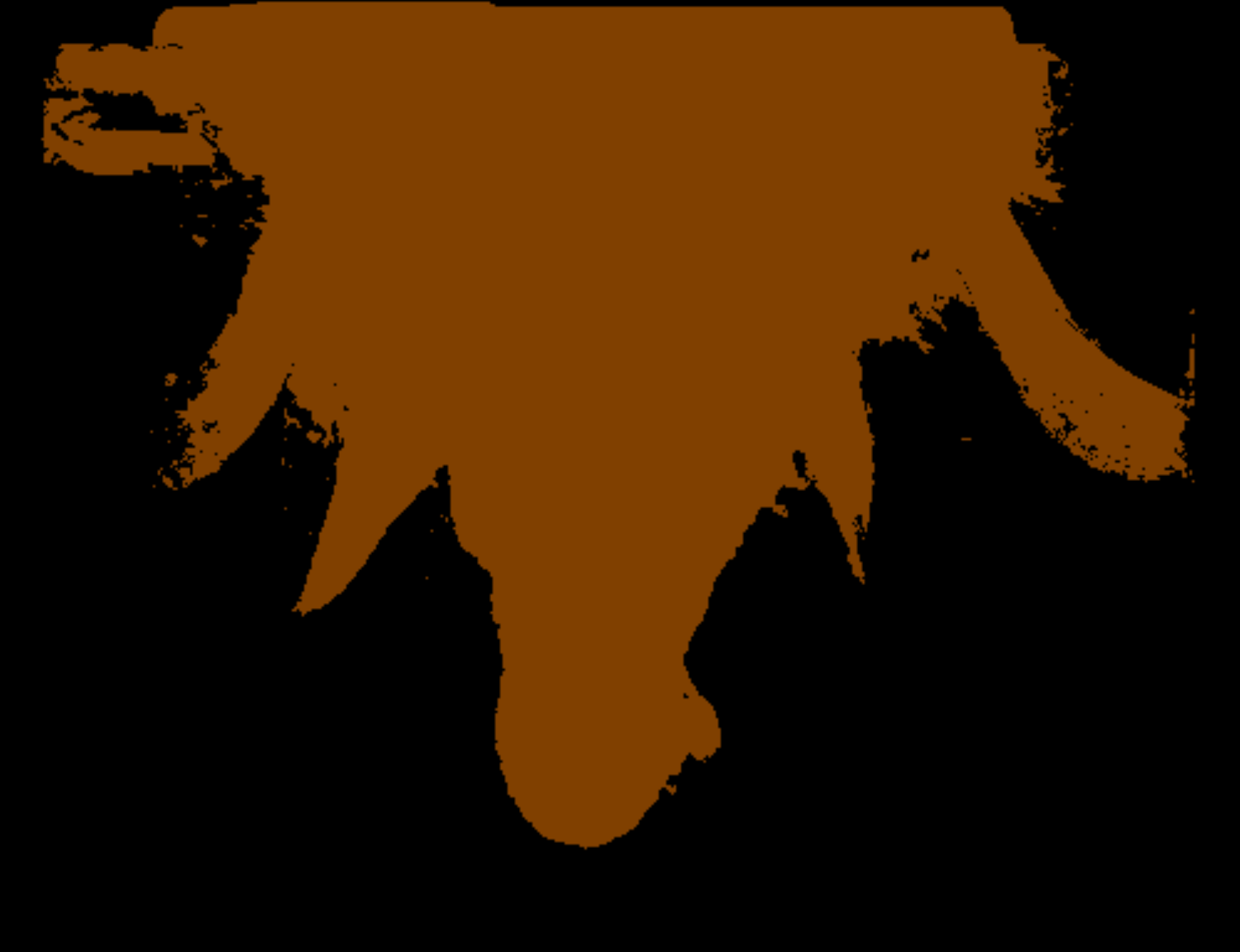}\\
        \vspace{0.02cm}
    \end{minipage}%
}%
\hspace{-3mm}
\subfigure{
    \begin{minipage}[t]{0.125\linewidth}
        \centering
        \includegraphics[width=0.8in,height=0.551in]{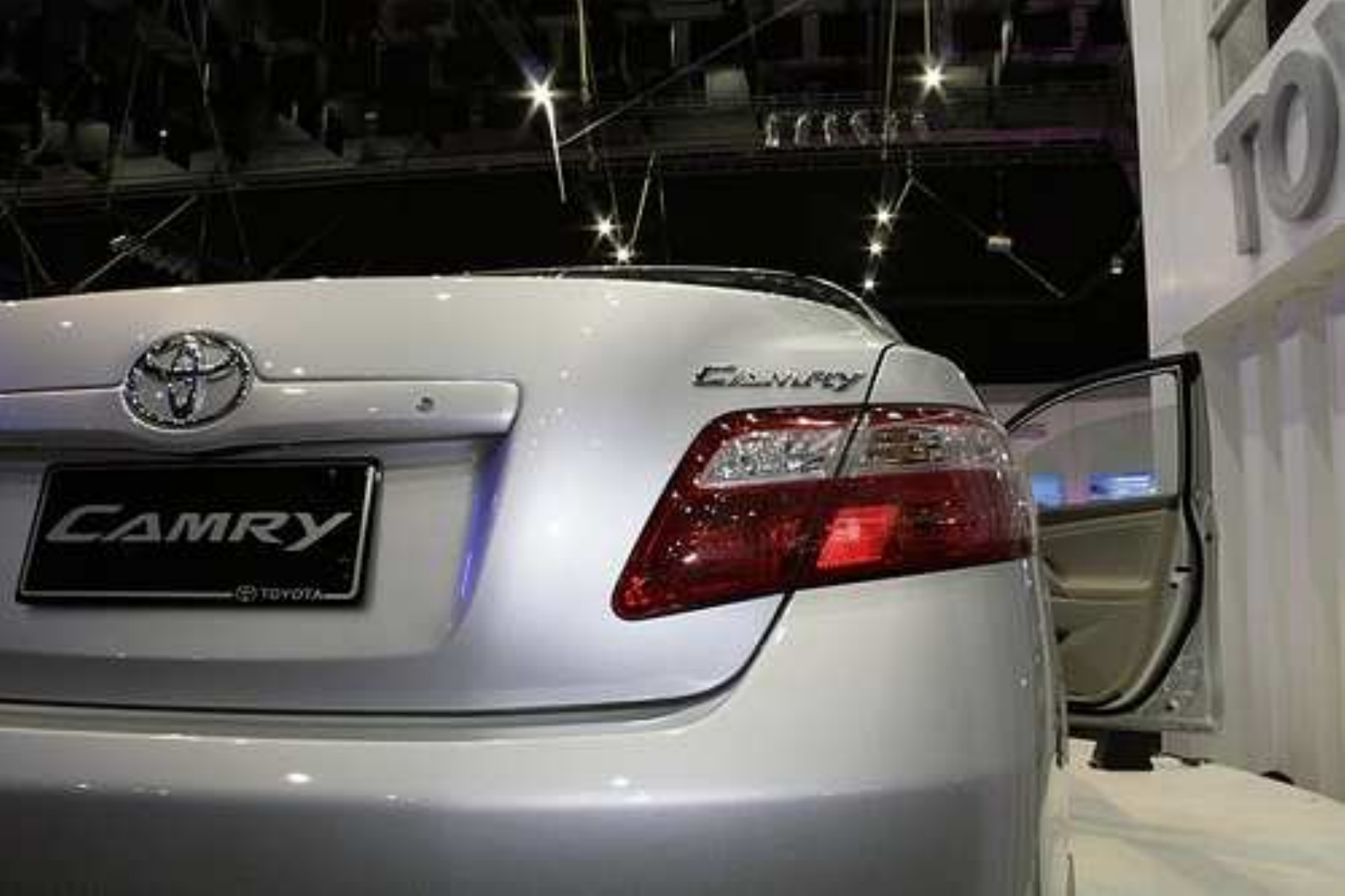}\\
        \vspace{0.02cm}
        \includegraphics[width=0.8in,height=0.551in]{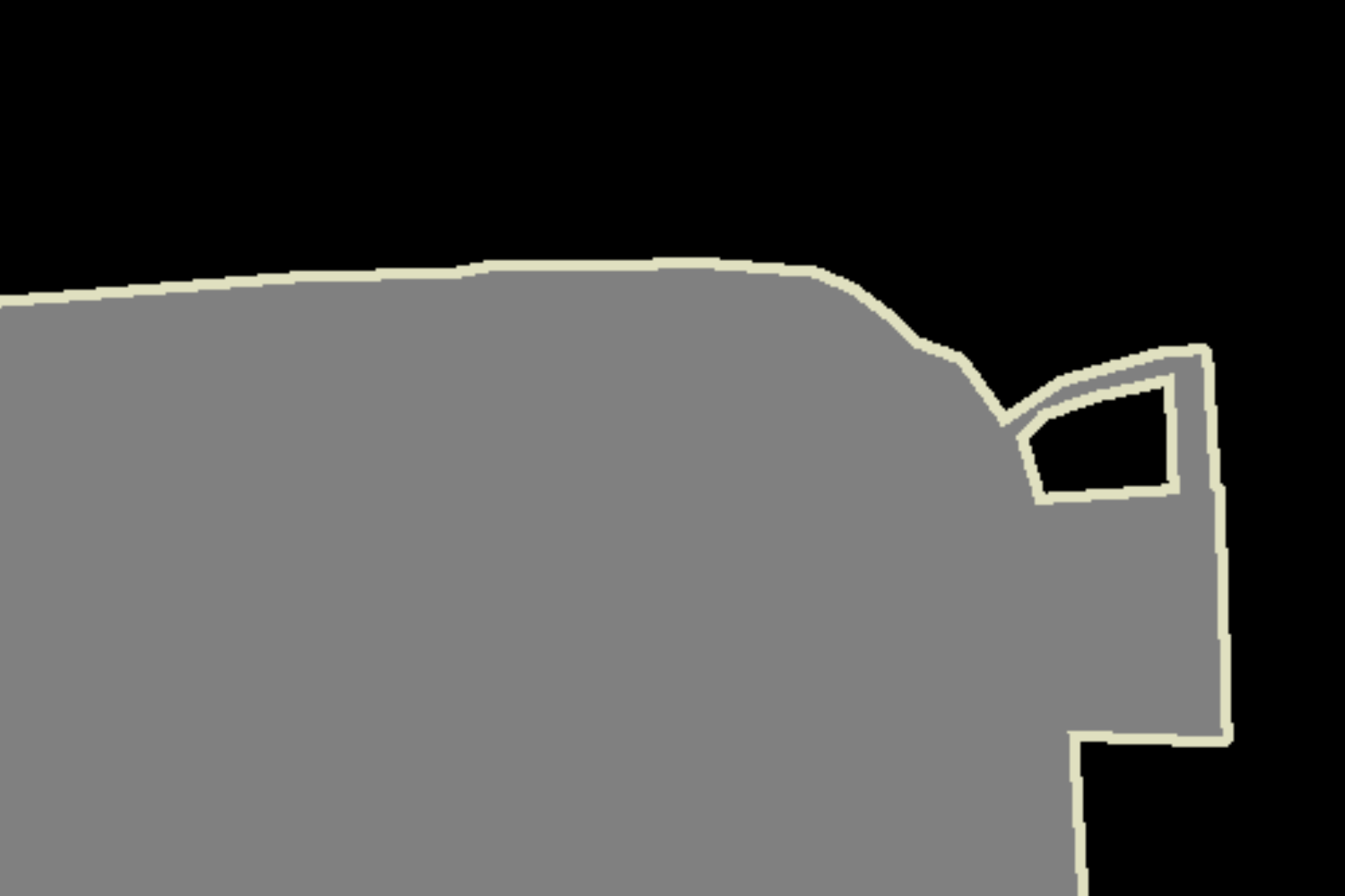}\\
        \vspace{0.02cm}
        \includegraphics[width=0.8in,height=0.551in]{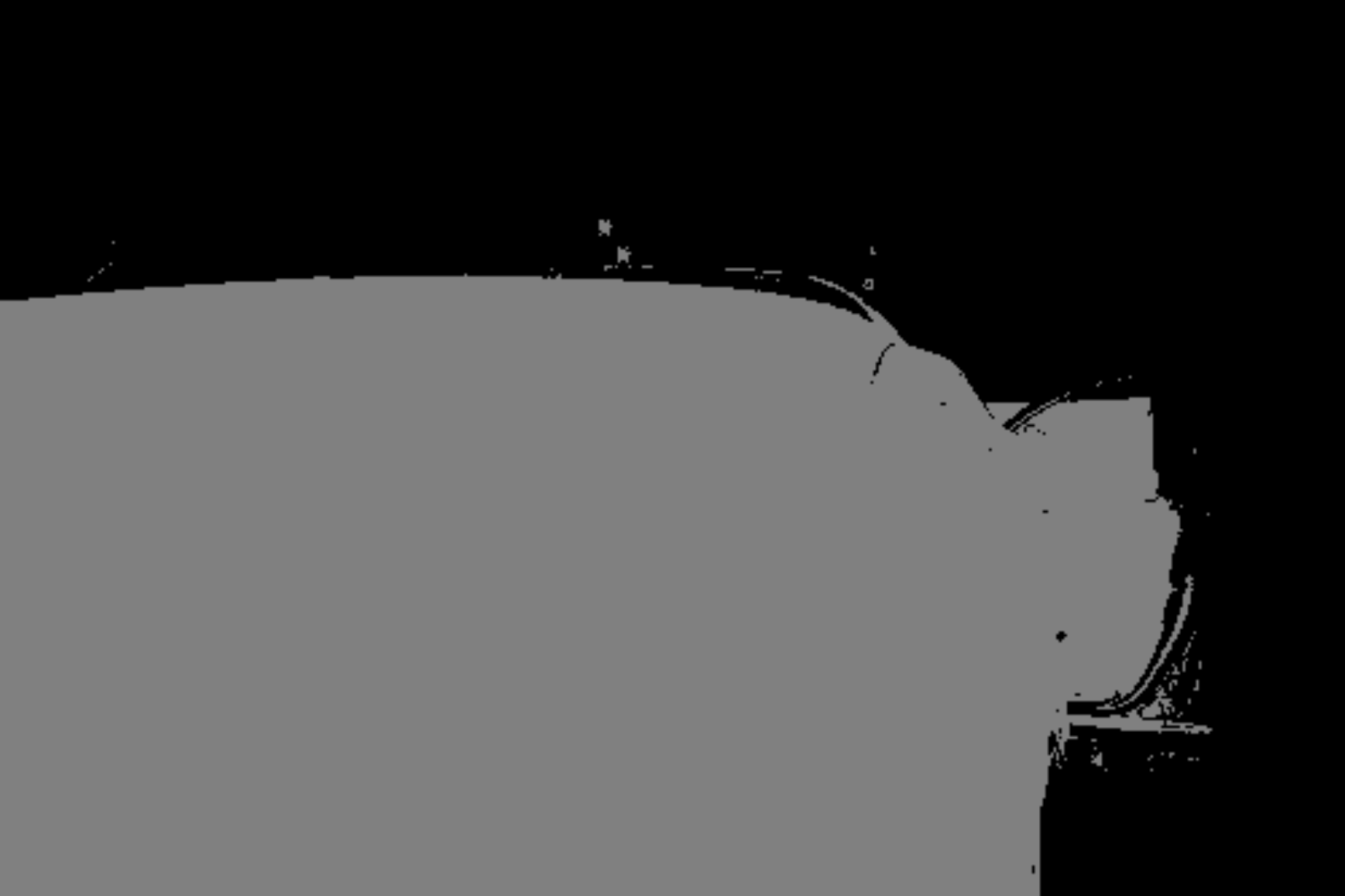}\\
        \vspace{0.02cm}
    \end{minipage}%
}%
\hspace{-3mm}
\subfigure{
    \begin{minipage}[t]{0.125\linewidth}
        \centering
        \includegraphics[width=0.8in,height=0.551in]{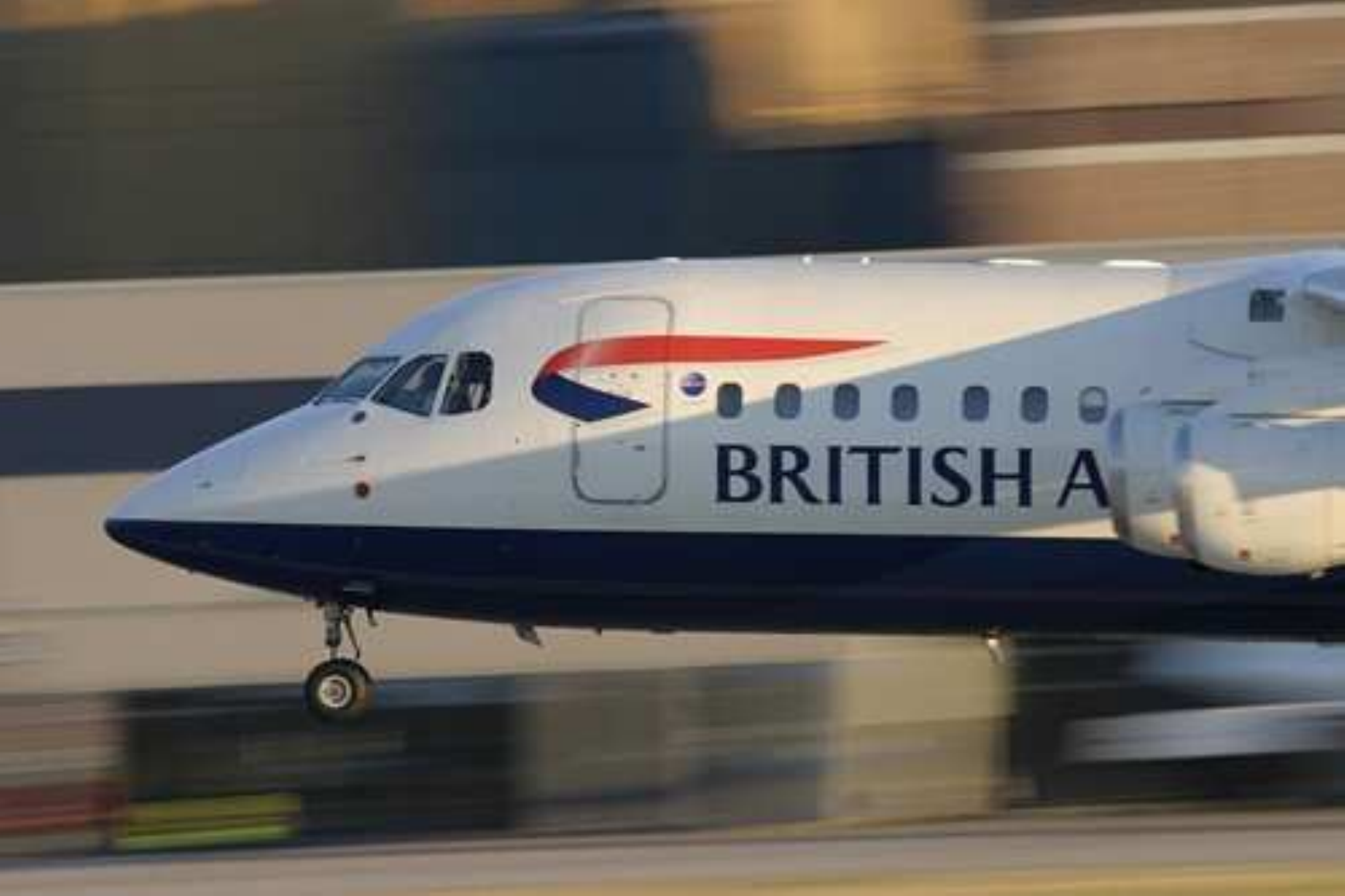}\\
        \vspace{0.02cm}
        \includegraphics[width=0.8in,height=0.551in]{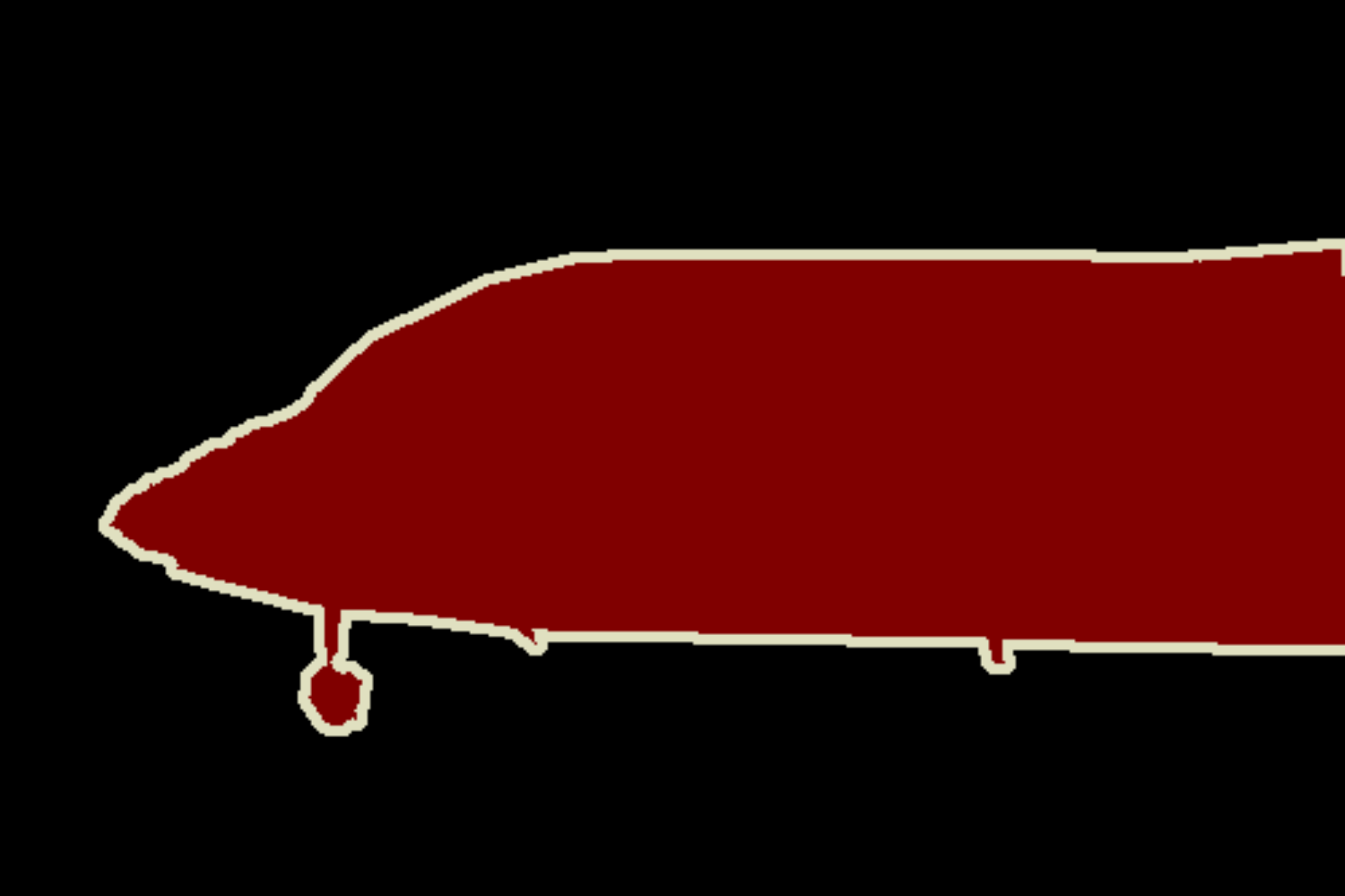}\\
        \vspace{0.02cm}
        \includegraphics[width=0.8in,height=0.551in]{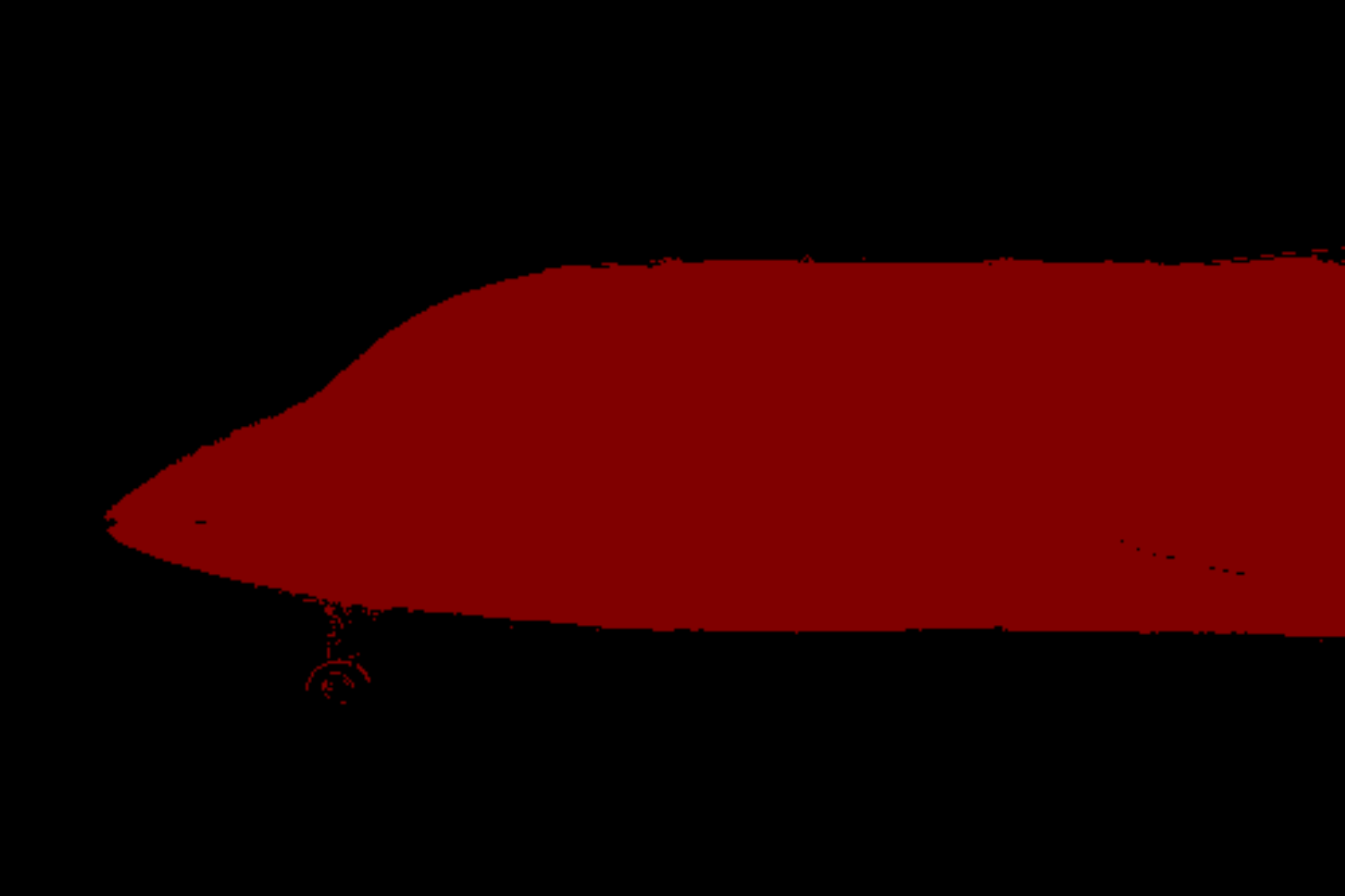}\\
        \vspace{0.02cm}
    \end{minipage}%
}%
\hspace{-4.5mm}
\subfigure{
    \begin{minipage}[t]{0.100\linewidth}
        \centering
        \includegraphics[width=0.5in,height=0.551in]{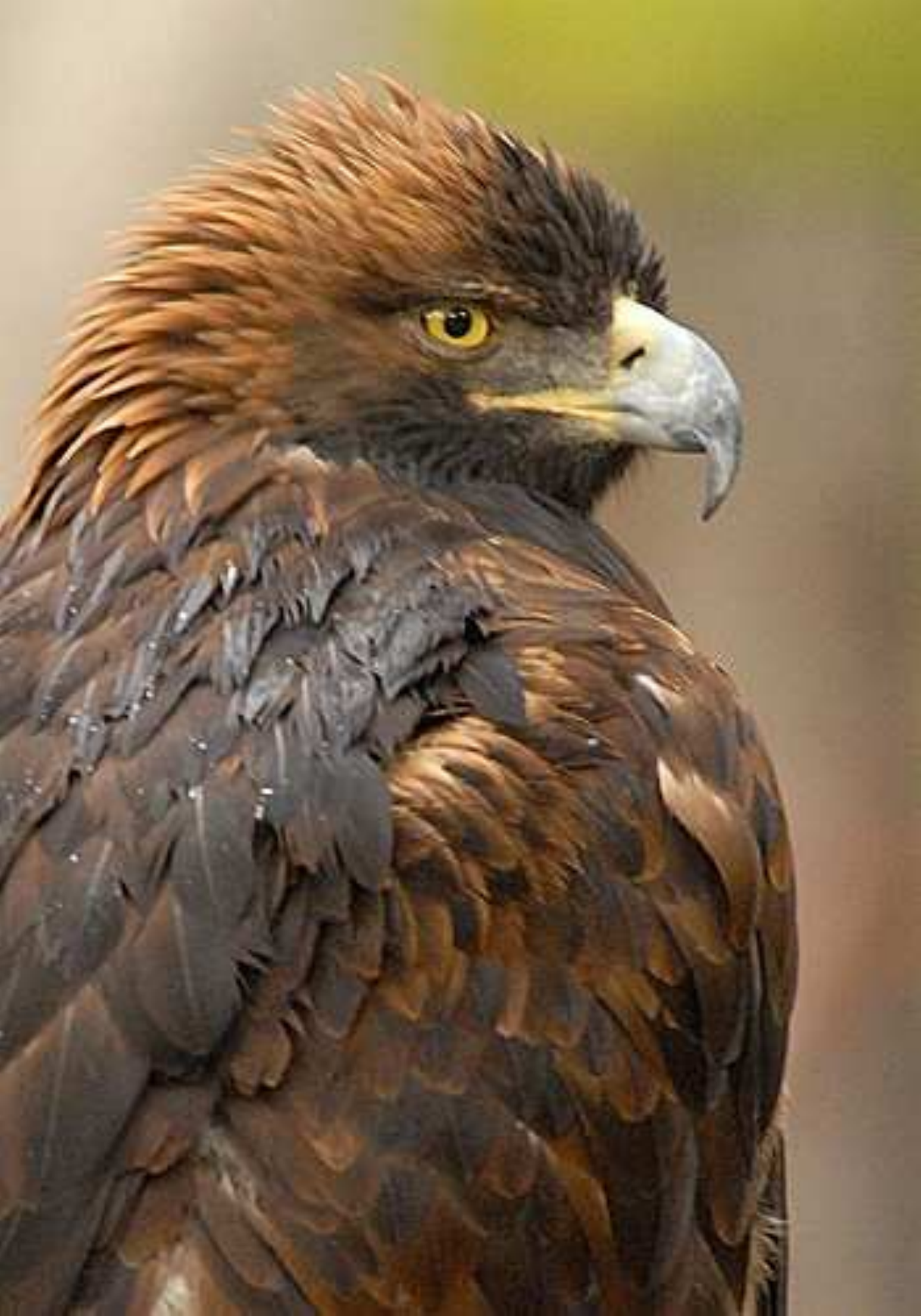}\\
        \vspace{0.02cm}
        \includegraphics[width=0.5in,height=0.551in]{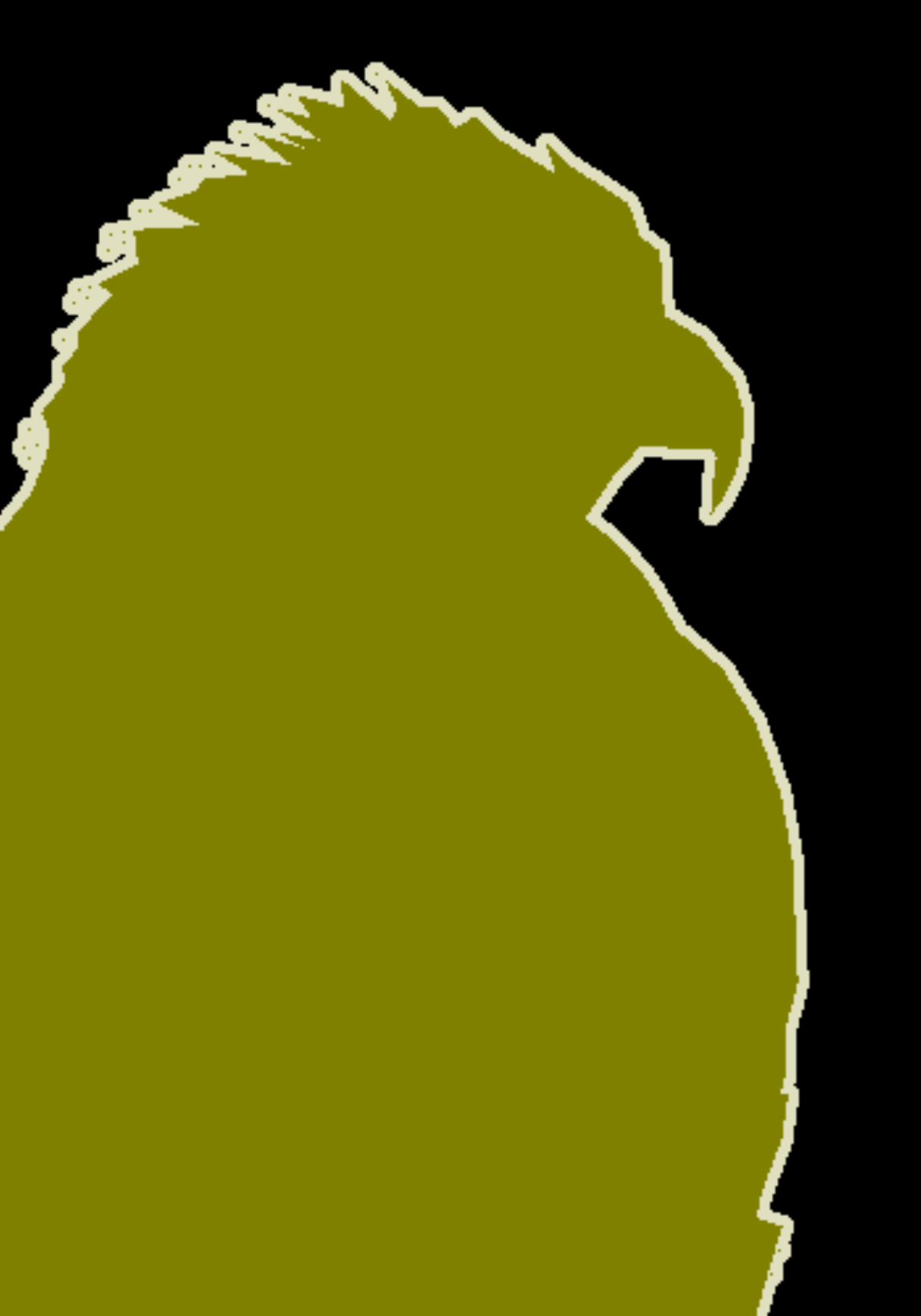}\\
        \vspace{0.02cm}
        \includegraphics[width=0.5in,height=0.551in]{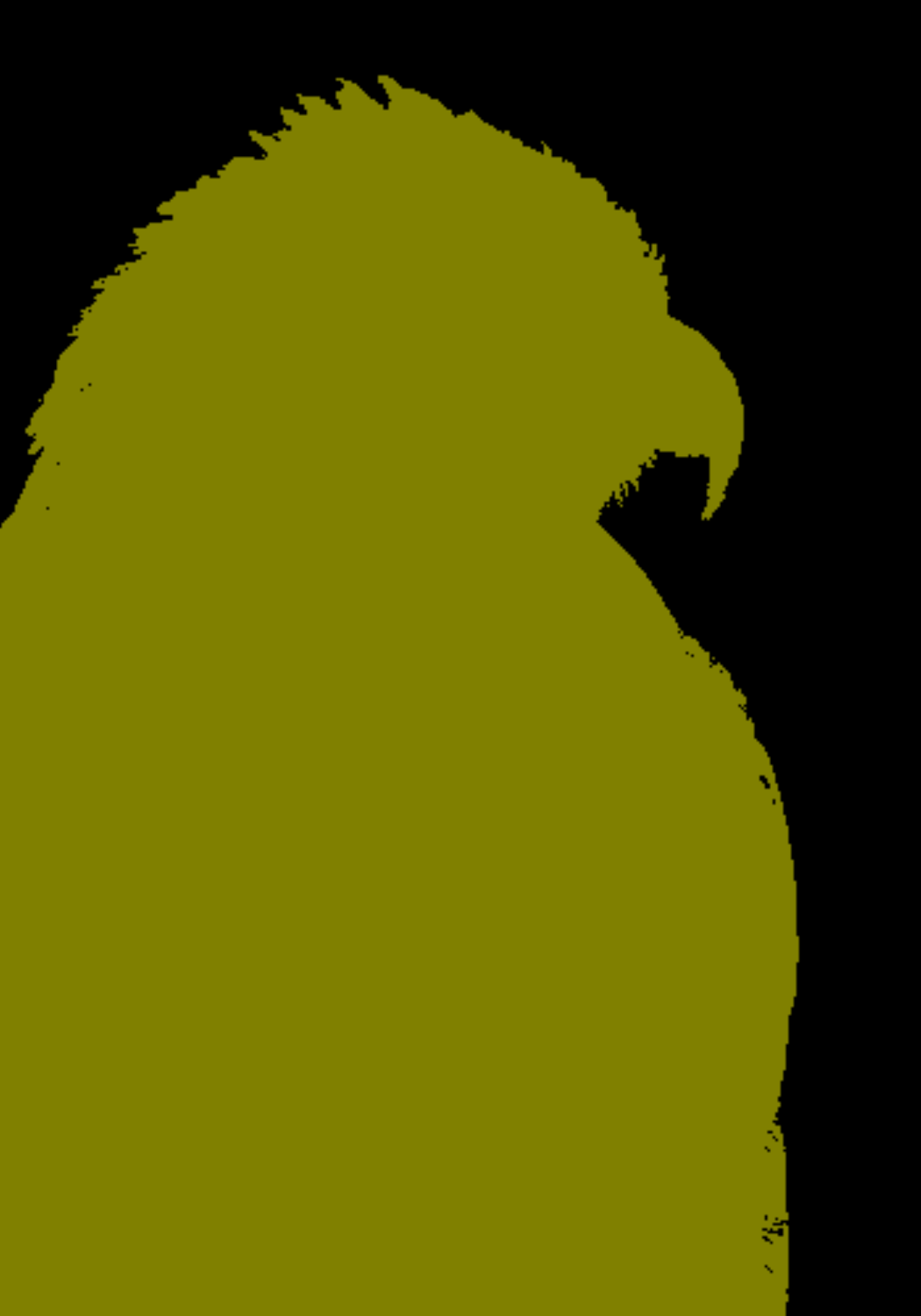}\\
        \vspace{0.02cm}
    \end{minipage}%
}%
\hspace{-4.5mm}
\subfigure{
    \begin{minipage}[t]{0.125\linewidth}
        \centering
        \includegraphics[width=0.8in,height=0.551in]{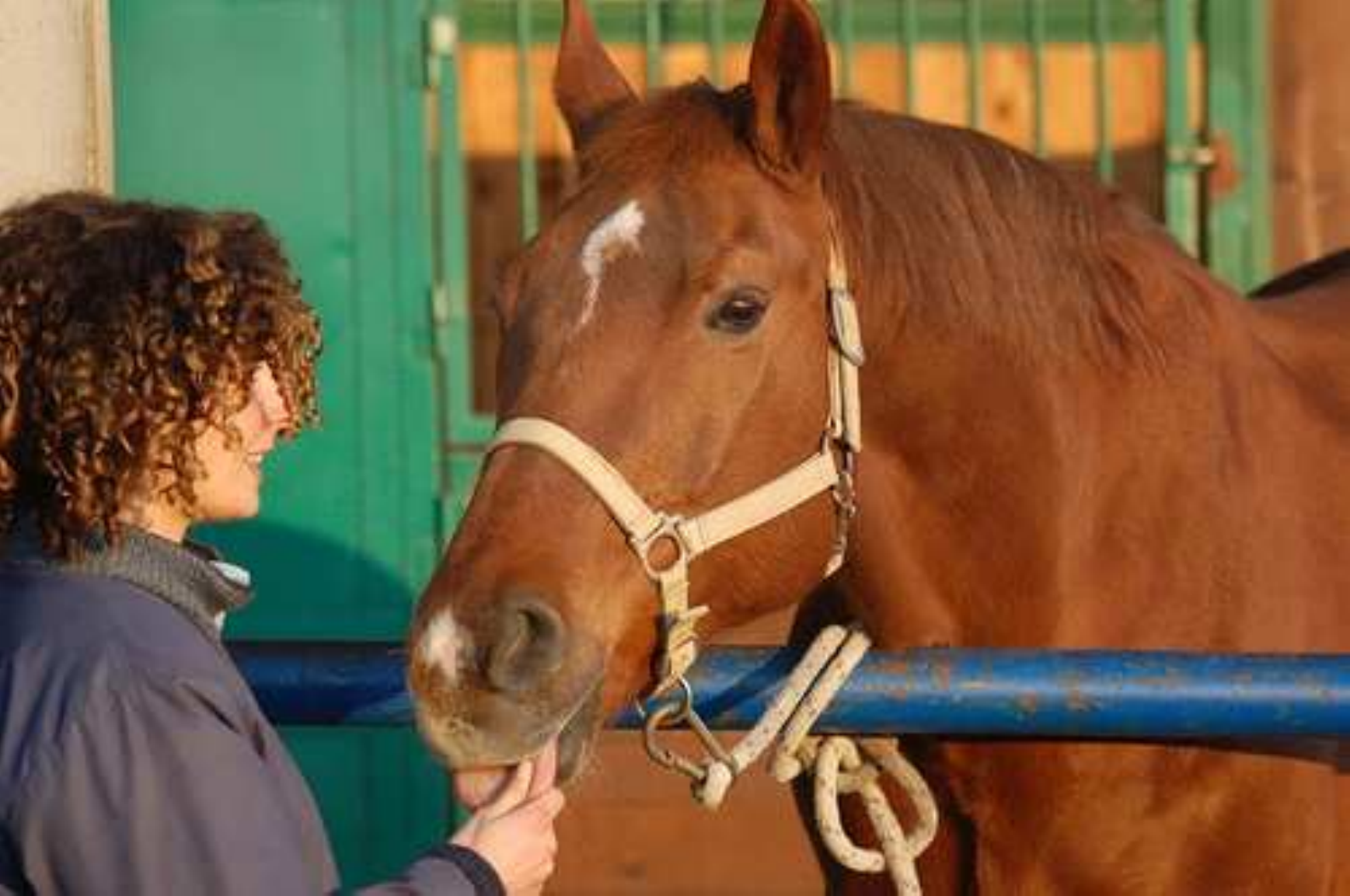}\\
        \vspace{0.02cm}
        \includegraphics[width=0.8in,height=0.551in]{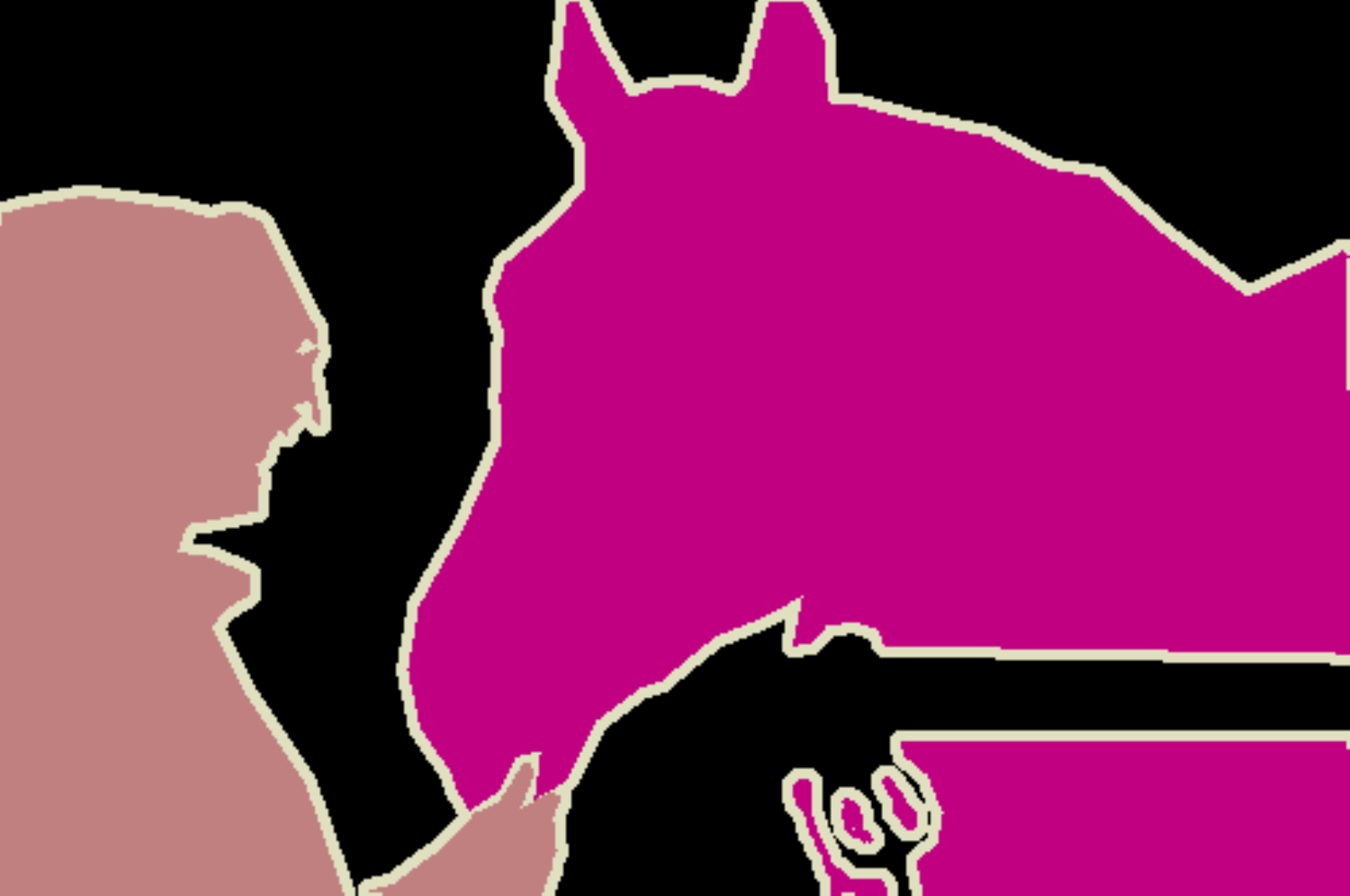}\\
        \vspace{0.02cm}
        \includegraphics[width=0.8in,height=0.551in]{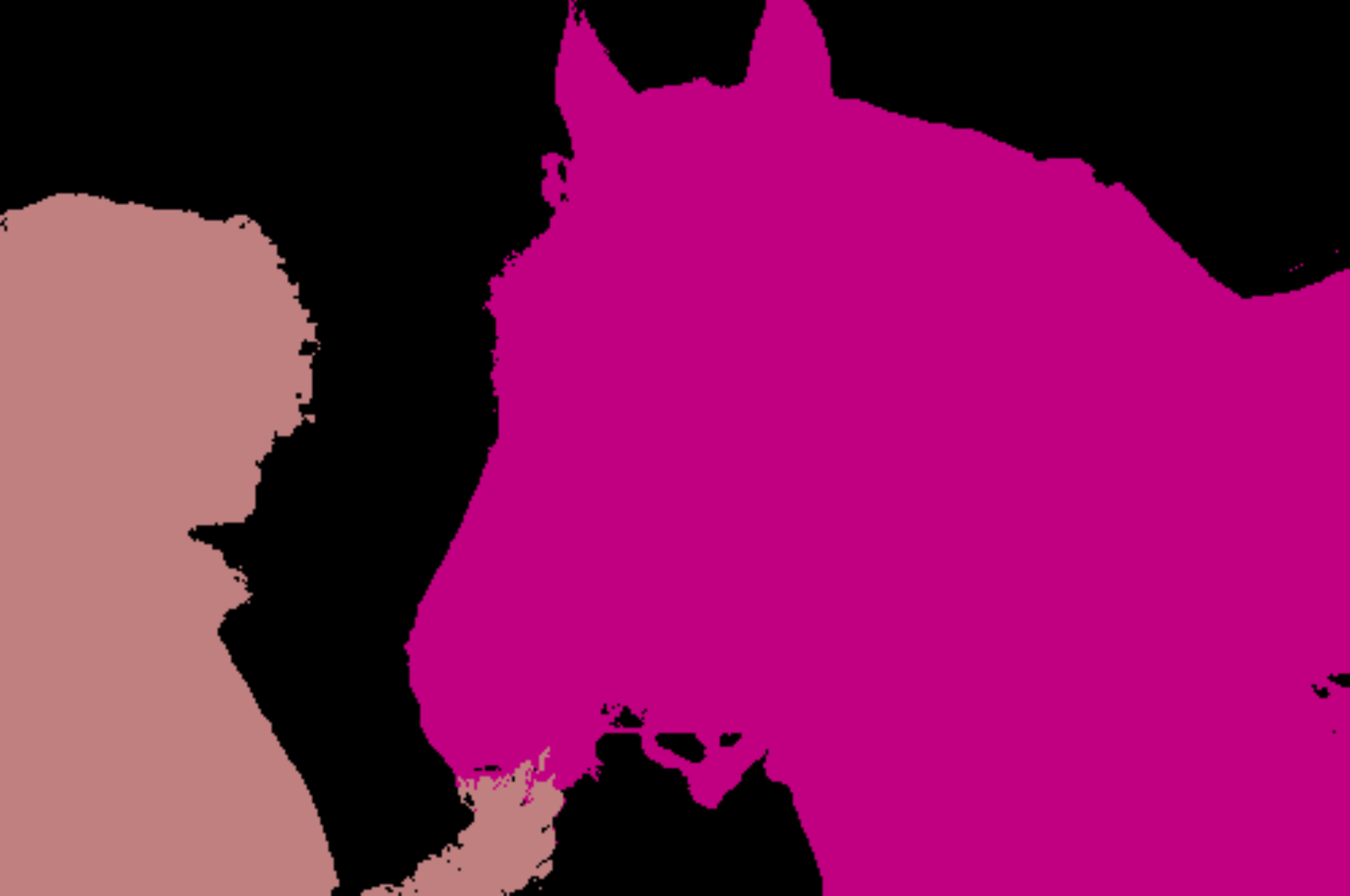}\\
        \vspace{0.02cm}
    \end{minipage}%
}%
\centering
\caption{Qualitative results on PASCAL VOC 2012 \textit{val} set. a) Input images. b) Ground-truth labels. c) Our segmentation results (w/ CRF).}
\label{fig:compare_sota}
\end{figure*}

\noindent\textbf{Hidden probability:}\; Recall that the hidden regions in one of the CP Pair are randomly selected with $p_h=0.5$ (Sec. \ref{sec_strategy}). Therefore, the number of the hidden patches in the CP Pair are expectedly equivalent. Here we aim to explore the relationship between the $p_h$ and our CPN. Due to the complementary attribution, we change $p_h$ from 0.1 to 0.5. Tab. \ref{tab:ablation on ph} shows the CPN with $p_h=0.5$ achieves the best performance (\textbf{57.43\%}) , and reaches the bottom (55.52\%) with $p_h=0.1$. The result also validates the effect of the extreme condition 1) on our model since $\boldsymbol{I}_h$ or $\boldsymbol{I}_{\overline h}$ is close to $\boldsymbol{I}$ as $p_h$ decreases.
\begin{table}[!htbp]\small
\centering\
\begin{tabular}{c|c|c|c|c|c}
\hline
$p_h$ & 0.1 & 0.2 & 0.3  & 0.4 & 0.5\\
\hline
mIoU(\%) & 55.52 & 56.87 & 56.29 & 57.05 &\textbf{57.43} \\
\hline
\end{tabular}
\caption{The performance of our CPN with different hidden probabilities $p_h$. }
\label{tab:ablation on ph}
\vspace{-2mm}
\end{table}

\subsection{Comparison with the state of the art}
To further improve our CAM, we use a common approach, namely Random Walk (RW)~\cite{affinity}, to improve the mIoU of our pseudo labels generated by the CPN up to 67.79\%. Following common practice, we then evaluate the quality of the final masks by using  DeepLab~\cite{v1} with ResNet38 backbone. Note that post CRF refinement is used for the output maps. Tab. \ref{tab:sota} gives a comparative overview concerning the previous methods. For all the approaches using ResNet38 backbone, our approach presents the state of the art performance on both PASCAL VOC 2012 \textit{val} and \textit{test} set, respectively scoring \textbf{67.8\%} and \textbf{68.5\%}. We also note that our results without applying CRF achieve better performance than MCIS~\cite{coatten}. In addition, our method achieves better performance on \textit{test} set than the ICD~\cite{ICD}, which uses extra supervision labels. Fig. \ref{fig:compare_sota} shows some samples of the final segmentation results, validating the effectiveness of our CPN.
\begin{table}[!htbp]\small
\centering
\begin{tabular}{c|c|c|c|c}
\hline
Methods & Pub. & Sup. & Val & Test \\
\hline
*MCOF~\cite{MCOF} & CVPR18& $\mathcal{I+S}$ & 60.3 & 61.2\\
*SeeNet~\cite{Seenet} & NIPS18 & $\mathcal{I+S}$ & 63.1 & 62.8 \\
*DSRG~\cite{dsrg} &CVPR18 & $\mathcal{I+S}$ & 61.4 & 63.2\\
$\dagger$AffinityNet~\cite{affinity}&CVPR18&$\mathcal{I}$&61.7&63.7\\
$\dagger$Single-Stage~\cite{1stage} &CVPR20&$\mathcal{I}$&62.7&64.3\\
*CIAN~\cite{CIAN}&AAAI20&$\mathcal{I}$&64.3&65.3\\
*FickleNet~\cite{fickle}&CVPR19&$\mathcal{I+S}$&64.9&65.3\\
$\dagger$SSDD~\cite{SSDD}&ICCV19&$\mathcal{I}$&64.9&65.5\\
$\dagger$SEAM~\cite{seam}&CVPR20&$\mathcal{I}$&64.5&65.7\\
*SubCat~\cite{subE} &CVPR20&$\mathcal{I}$&66.1&65.9\\
*RRM~\cite{RMM} &AAAI20&$\mathcal{I}$&66.3&66.5\\
*BES~\cite{BES} &ECCV20&$\mathcal{I}$&65.7&66.7\\
$\dagger$Conta~\cite{conta}&NIPS20&$\mathcal{I}$&66.1&66.7\\
*MCIS~\cite{coatten}&ECCV20&$\mathcal{I}$&66.2&66.9\\
*ICD~\cite{ICD}&CVPR20&$\mathcal{I+S}$&67.8&68.0\\
\hline
$\dagger$Ours (w/o CRF) &-&$\mathcal{I}$&66.8&67.6\\
$\dagger$Ours (w/ CRF) &-&$\mathcal{I}$&\textbf{67.8}&\textbf{68.5}\\
\hline
\end{tabular}
\caption{Comparison with SOTA on VOC 2012 \textit{val} and \textit{test} in terms of mIoU (\%). Methods marked by * use ResNet101 backbone, the others marked by $\dagger$ use ResNet38. The supervision (Sup.) contains image-level label ($\mathcal{I}$) and saliency maps ($\mathcal{S}$).}
\label{tab:sota}
\end{table}
\section{Conclusion}
In this paper, we have proposed a simple yet effective pipeline for weakly supervised semantic segmentation with only image-level labels provided. First, from the information theory perspective, we showed that the sum of CAMs generated by a pair of images with Complementary Patch regions (CP Pair) is able to mine out more foreground seeds.
Then, based on this observation, we presented a CP Network (CPN) with a bunch of regularization to achieve an improved CAM.
To further refine the results, we designed a Pixel-Region Correlation Module (PRCM) to bring more contextual information for the CAM. Extensive experiments on the PASCAL VOC 2012 dataset show that our proposed CPN achieves new state-of-the-art performance.

\noindent\textbf{Acknowledgements:}\; This research was supported in part by National Natural Science Foundation of China (61871325, 62001394), National Key Research and Development Program of China (2018AAA0102803, 2018YFB1703201, 2019YFB1704003, 2019YFB1706602), Shanghai Science and Technology Innovation Action Plan (19511105900), Chinese Ministry of Education Research Found on Intelligent Manufacturing (MCM20180703).

{\small
\bibliographystyle{ieee_fullname}
\bibliography{egpaper_final}
}

\end{document}